\documentclass[lettersize,journal]{IEEEtran}
\IEEEoverridecommandlockouts

\usepackage{amsmath,amsfonts}
\usepackage{algorithmic}
\usepackage{algorithm}
\usepackage{array}
\usepackage[caption=false,font=normalsize,labelfont=sf,textfont=sf]{subfig}
\usepackage{textcomp}
\usepackage{stfloats}
\usepackage{url}
\usepackage{verbatim}
\usepackage{graphicx}
\usepackage{cite}
\usepackage[utf8]{inputenc} 
\usepackage[T1]{fontenc}
\usepackage{hyperref} 
\usepackage{url}
\usepackage{booktabs} 
\usepackage{amsfonts} 
\usepackage{nicefrac} 
\usepackage{microtype}
\usepackage{graphicx}
\usepackage{placeins}

 \usepackage{multirow}
 \usepackage{pdfpages}
 \usepackage{tikz}
 \usepackage{amssymb}
\usepackage{pifont}
\newcommand{\cmark}{\text{\ding{51}}}
\newcommand{\xmark}{\text{\ding{55}}}
\usepackage{bbold}
\usepackage{caption}
\usepackage[normalem]{ulem}
\usepackage{xcolor}
\definecolor{azure}{rgb}{0.0, 0.5, 1.0}
\usepackage{algorithm}
\usepackage{algorithmic}

\usepackage{xspace}
\usepackage{amssymb}

\definecolor{rc1}{RGB}{4, 142, 255}
\definecolor{rc2}{RGB}{255, 104, 3}
\definecolor{azuremist}{rgb}{0.94, 1.0, 1.0}
\definecolor{beaublue}{rgb}{0.74, 0.83, 0.9}
\definecolor{blizzardblue}{rgb}{0.67, 0.9, 0.93}
\definecolor{lightgray}{rgb}{0.83, 0.83, 0.83}
\newcommand\rankone[1]{\colorbox{lightgray}{\textcolor{black}{\textbf{#1}}}}
\newcommand\ranktwo[1]{\colorbox{lightgray}{\textcolor{black}{\textbf{#1}}}}
\newcommand\rankthree[1]{\colorbox{lightgray}{\textcolor{black}{\textbf{#1}}}}
\newcommand{\RNum}[1]{\uppercase\expandafter{\romannumeral #1\relax}}

\usepackage{siunitx}
\newcolumntype{H}{>{\setbox0=\hbox\bgroup}c<{\egroup}@{}}
\title{Auditing and Generating Synthetic Data\\ with Controllable Trust Trade-offs}

\author{Brian Belgodere, Pierre Dognin, Adam Ivankay, Igor Melnyk, Youssef Mroueh, Aleksandra Mojsilovic, Jiri Navratil, Apoorva Nitsure, Inkit Padhi, Mattia Rigotti, Jerret Ross, Yair Schiff, Radhika Vedpathak, and Richard A. Young.\thanks{Authors are listed in alphabetical order. Correspondence to \texttt{mroueh@us.ibm.com} and \texttt{inkpad@ibm.com}}} 

\begin{document}
\maketitle
\begin{abstract}
Real-world data often exhibits bias, imbalance, and privacy risks. Synthetic datasets have emerged to address these issues by enabling a paradigm that relies on generative AI models to generate unbiased, privacy-preserving data while maintaining fidelity to the original data. However, assessing the trustworthiness of synthetic datasets and models is a critical challenge.
We introduce a holistic auditing framework that comprehensively evaluates synthetic datasets and AI models. It focuses on preventing bias and discrimination, ensuring fidelity to the source data, and assessing utility, robustness, and  privacy preservation.
We demonstrate our framework's effectiveness by auditing various generative models across diverse use cases like education, healthcare, banking, and human resources, spanning different data modalities such as tabular, time-series, vision, and natural language. This holistic assessment is essential for compliance with regulatory safeguards. We introduce a trustworthiness index to rank synthetic datasets based on their safeguards trade-offs.
Furthermore, we present a trustworthiness-driven model selection and cross-validation process during training, exemplified with ``TrustFormers'' across various data types. This approach allows for controllable trustworthiness trade-offs in synthetic data creation. Our auditing framework fosters collaboration among stakeholders, including data scientists, governance experts, internal reviewers, external certifiers, and regulators. This transparent reporting should become a standard practice to prevent bias, discrimination, and privacy violations, ensuring compliance with policies and providing accountability, safety, and performance guarantees.
\end{abstract}


\section*{Introduction}\label{sec:introduction}

Generative models have demonstrated impressive results in synthesizing high-quality data across multiple modalities from  tabular and time-series data \cite{qian2023synthcity,SDV,padhi2021tabular} to text \cite{brown2020language,christiano2017deep}, images \cite{ramesh2022hierarchical,ho2020denoising,sohl2015deep,brock2018large} and  chemistry \cite{molGPT}. We are entering a new era in training AI models, where synthetic data can be used to augment real data \cite{allassoniere}, or as a complete replacement, in the most extreme case \cite{gartner}. One of the main motivations behind controllable synthetic data usage in training AI models is its promise to synthesize privacy-preserving data that enables safe sharing without putting the privacy of real users and individuals at risk. This has the potential to circumvent cumbersome processes that are at the heart of many highly regulated fields such as financial services \cite{synthetic_finance} and healthcare, for example \cite{chen2021synthetic,bhanot2021problem,dahmen2019synsys,allassoniere}. Another motivation comes from controlling the generation process in order to balance the training data and reduce biases against protected groups and sensitive communities \cite{choi2020fair}. Finally, synthetic data also offers new opportunities in simulating non-existing scenarios, providing grounding for causal inference via the generation of counterfactuals that would help explain some observations in the absence of real data \cite{ebert2017causal}.\\

\noindent Synthetic data can take different forms, ranging from seedless approaches \cite{altman2021synthesizing,DBLP:conf/esws/Jimenez-RuizHEC20}, which rely on knowledge bases and rule-based generation but incur risks of biased grounding and linkage attacks, to data generated through AI models trained on real data, which may lead to memorization, privacy breaches, and bias amplification \cite{carlini2021extracting, hall2022systematic, choi2020fair}. These AI-generated datasets can also pose legal issues related to copyright and intellectual property \cite{vyas2023provable}. 
Both types of synthetic data need thorough auditing for safety, privacy, fairness, and utility alignment.\\

\noindent Amidst these technological advancements, the AI regulation landscape is rapidly evolving to institute safeguards and objectives for AI systems, aimed at mitigating societal risks and malicious use. For instance, the recent executive order on Safe, Secure, and Trustworthy Artificial Intelligence by the Biden administration highlights the urgency of these concerns. Other acts, like the U.S. Algorithmic Accountability Act and the EU AI Act (for a detailed comparison, see \cite{EU-US-AIacts}), are paving the way towards fostering trustworthy AI. The EU AI Act, in particular, enforces conformity assessments and post-market monitoring of AI models. Furthermore, quantitative auditing of predictive AI models has made significant progress in recent years, with various auditing systems emerging in domains such as algorithmic recruitment \cite{recruitement_audit} and healthcare \cite{medical_audit}. Multiple AI risk assessment frameworks have been proposed to tackle certain aspects of trust but have focused on individual trust pillars, such as fairness \cite{bellamy2018ai}, explainability \cite{arya2021ai}, or robustness \cite{nicolae2018adversarial}. With the advent of foundation models \cite{bommasani2021opportunities} and Large Language Models (LLMs), several techniques \cite{li2022large, psfairij} are being explored in an attempt to mitigate risks; moreover, multiple frameworks have suggested probing these models on specific trust aspects via red teaming \cite{perez2022discovering}, reconstruction of training data attacks \cite{carlini2021extracting}, or via holistic auditing, as proposed in HELM (Holistic Evaluation of Language Models) 
\cite{liang2022holistic} and the auditing framework of \cite{houssiau2022framework}. Finally, several governance mechanisms have been proposed to ensure transparency in communicating the risks of data and models via fact sheets \cite{FactSheets}, model cards \cite{mitchell2019model}, data sheets\cite{gebru2021datasheets}, and system and method cards \cite{adkins2022method}.\\

\noindent Existing frameworks, like Synthetic Data Vault (SDV) \cite{SDV} and Synthcity \cite{qian2023synthcity}, tend to focus on specific aspects of synthetic data auditing, often overlooking a comprehensive evaluation of all trust dimensions. The TAPAS framework  \cite{houssiau2022tapas}, on the other hand, is primarily dedicated to privacy \cite{houssiau2022tapas}, without fully addressing trade-offs with other essential dimensions. There are also efforts that concentrate on fidelity and utility auditing, such as \cite{alaa2022faithful}, and those examining privacy-preserving capabilities \cite{jagielski2020auditing,vanbreugel2023membership,chen2020gan}. Unfortunately, these initiatives in addition to not being holistic, they frequently  fail to account for the uncertainty introduced by data splits between training and testing sets. Addressing these gaps is crucial for achieving a comprehensive  audit of synthetic data in AI applications, as emphasized in a recent European Parliament report \cite{mokander2023auditing}.\\

\begin{figure*}[ht!]
\centering
\includegraphics[width=0.7\linewidth]{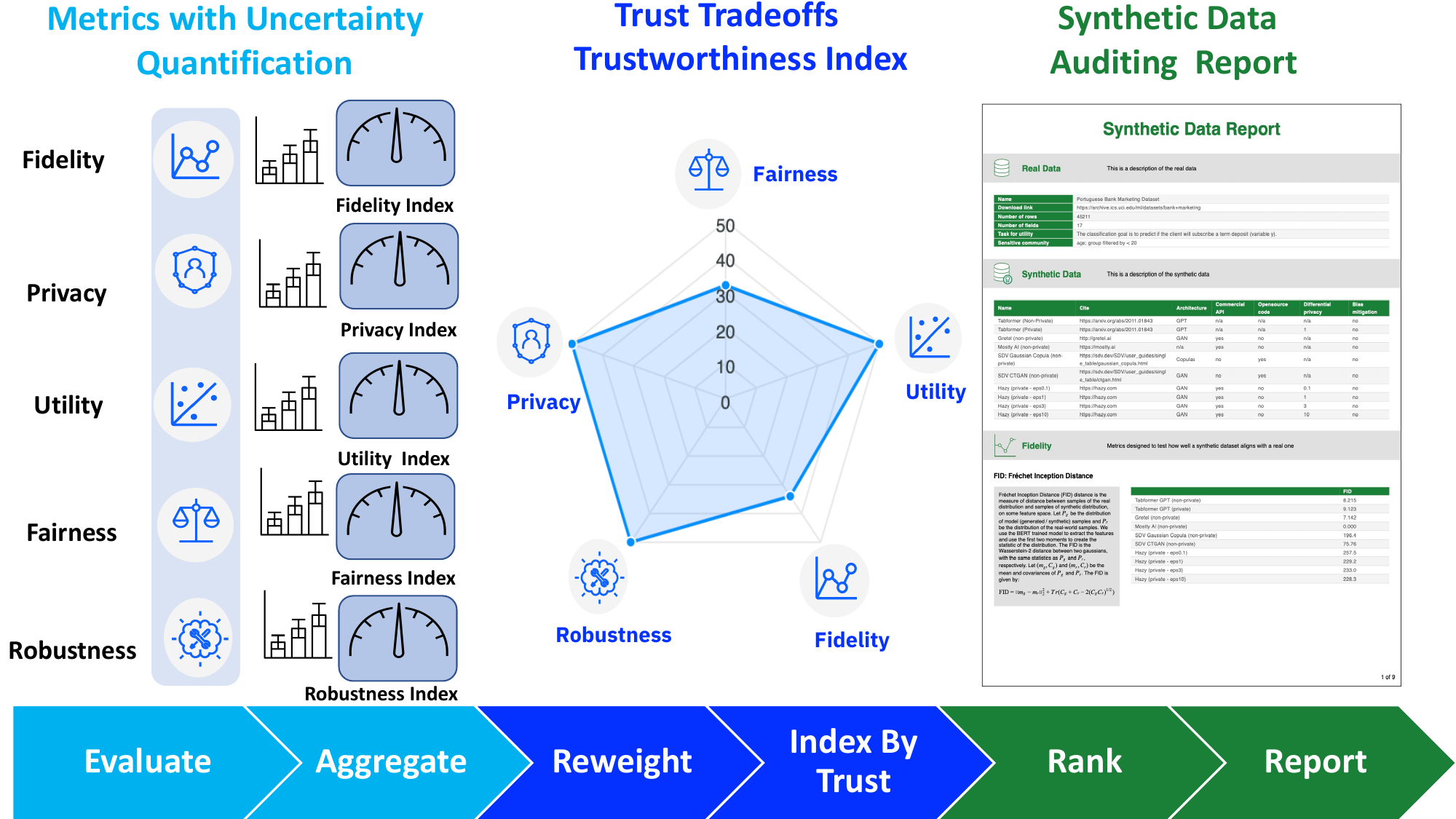} 
\caption{Summary diagram of our proposed holistic synthetic data auditing framework. For each trust dimension (fidelity, privacy, utility, fairness, and robustness), we evaluate multiple metrics on the synthetic data and quantify their uncertainty. Metrics are aggregated within each trust dimension, which results in trust dimension indices. These indices are re-weighted with desired trust trade-offs to produce the trustworthiness index. Different synthetic datasets are then ranked using this trustworthiness index, and a summary of the audit is written to an audit report. The ranking produced by our audit enables comparison of different synthetic data produced by various generative modeling techniques, and aids the model selection process for a given generation technique, allowing its alignment with prescribed safeguards. The model selection is performed via trustworthiness index  driven cross-validation, which results in controllable trust trade-offs by producing new ranks for different desired weighing trade-offs for a given application and use case.}
\label{Fig:flowchartM}
\end{figure*}

\noindent To address these challenges, we propose a framework for auditing the trustworthiness of synthetic data that is \emph{holistic}, \emph{transversal} across different modalities (tabular, time-series, computer vision and natural language) and assess the uncertainty in auditing a generative model (see Figure \ref{Fig:flowchartM} for a summary of our approach). \\

\noindent Our main contributions are as follows:  we introduce a holistic framework for auditing the trustworthiness of synthetic data, covering key trust dimensions like fidelity, utility, privacy, fairness, and robustness. We define a \emph{trustworthiness index} that evaluates synthetic data and their downstream tasks. We provide methods for controlling trust trade-offs in synthetic data during training, notably through model selection via the trustworthiness index. We instrument transformer models across multiple modalities with these control mechanisms and refer to them as ``TrustFormers''. 
By applying our framework, we demonstrate that downstream tasks using trustworthiness-index-driven cross-validation often outperform those trained on real data while meeting privacy and fairness requirements. Finally, our framework offers transparency templates for clear communication of the risks of synthetic data via an audit report.\\

\section*{Synthetic Data Auditing Framework }\label{sec:auditing_framework}

In this section, we present our auditing framework. For a given real dataset and several synthetic datasets, our framework evaluates a multitude of \emph{\textbf{trust dimensions}}, namely: fidelity, privacy, utility, fairness, and robustness.

\noindent\paragraph{Setup} Formally, given a real dataset $D_{r}$ and multiple synthetic datasets $D^j_{s},~j=1,\ldots,N$. The real dataset is split to training, development/validation, and testing sets as follows: 
\begin{equation}
D_r=\{D_{r,\text{train}},~D_{r,\text{val}},~D_{r,\text{test}}\}.
\end{equation}
Synthetic datasets come from various sampling schemes from different types of generative models. These models are trained on the real training set $D_{r,\text{train}}$ and validated on the real development set $D_{r,\text{val}}$. The utility of these synthetic data is measured via a predictive downstream task defined on the data space along protected and 
sensitive groups for whom we want to ensure a fair prediction. The downstream task is trained on the synthetic data $D_{s, train}^j$, validated on the real development set $D_{r,\text{val}}$, and evaluated on the real test set $D_{r,\text{test}}$ (note that $D_{r,\text{test}}$ could have a distribution shift w.r.t to $D_{r,\text{train}}$).
Without loss of generality, we assume for simplicity that all downstream tasks are classification tasks.

\begin{figure*}[htp!]
\centering
\includegraphics[scale=0.35]{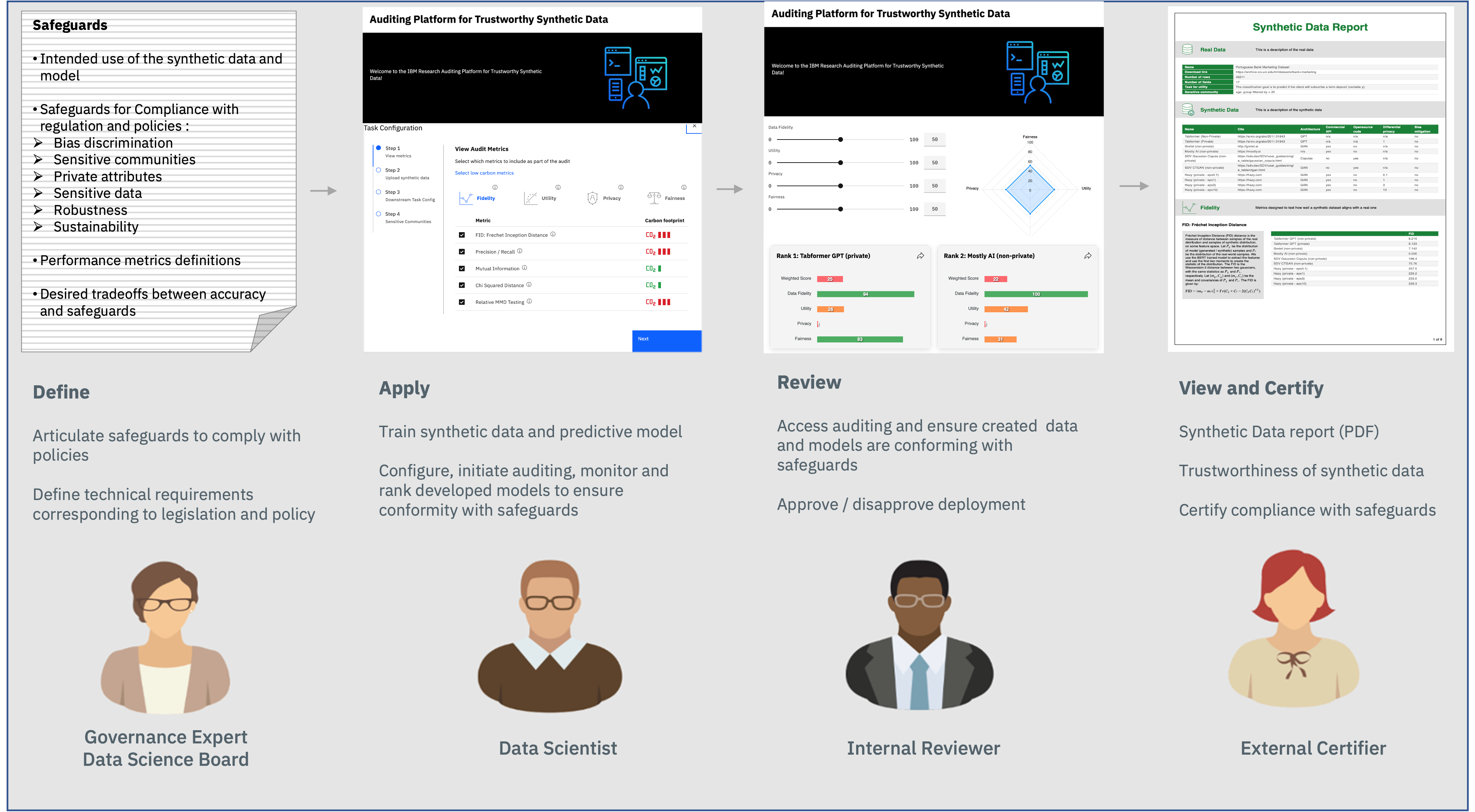} 
\caption{Auditing Platform and workflows connecting different stakeholders (e.g., data scientists, data governance experts, internal reviewers, external certifiers, and regulators) from model development to audit and certification via a synthetic data auditing report.}
\label{fig:workflowM}
\end{figure*}

\paragraph{Auditing Framework } Our synthetic data auditing framework, as depicted in Figure \ref{Fig:flowchartM}, comprises several key stages:  Evaluate, Aggregate, Reweight, Index by trustworthiness, Rank, and Report. It serves as a means to enhance communication among governance experts, data scientists, internal reviewers, and external regulators. The primary personas involved in our framework are outlined in  Figure \ref{fig:workflowM}. The key stages of our audit framework and the personas involved are explained below (more detailed explanations of trust dimension and our framework are given in the Methods Section and Supplementary Information): 
\begin{enumerate}
 \item\textbf{\textsc{Evaluate:}}  The governance expert and data science board collaborate to establish quantitative metrics for each trust dimension, as detailed in  Table \ref{tab:metricMM}, ensuring adherence to socio-technical safeguards. These metrics are then assessed by data scientists.

\item\textbf{\textsc{Aggregate:}} Metrics within each dimension are aggregated into a trust dimension index (denoted as $\pi_{T}$, where "T" corresponds to Fidelity, Privacy, Utility, Fairness, or Robustness), which ranges from 0 to 1, representing compliance probability with the requirements of the dimension. This aggregation method is explained in our Methods Section.
\item\textbf{\textsc{Re-weight:}} Stakeholders, including internal reviewers, governance experts, and the application owner, establish a trustworthiness profile through trade-off weights $\omega$ (examples are shown in Table  \ref{Table:weights}), indicating the relative importance of trust indices for fulfilling specific requirements.
\item \textbf{\textsc{Index by Trustworthiness:}}  Trust dimension indices are re-weighted by the trustworthiness profile weights  $\omega$ and combined via a geometric mean to produce the trustworthiness index, $\tau_{\text{trust}}(\omega)$, a context-specific measure based on the application's needs and trustworthiness profiles.
\item \textbf{\textsc{Rank:}} The trustworthiness index allows internal reviewers to rank synthetic datasets. Data scientists can use it for model selection within a given generation method and trustworthiness profile. When determining our trustworthiness index ranking, we can additionally consider the uncertainty in real data splits. For a trustworthiness profile $\omega$, our preference is  for a generative AI technique that exhibits the highest average trustworthiness index across splits ($\overline{\tau_{\text{Trust}}}(\omega)$) while minimizing volatility ($ \Delta_{\tau}(\omega)$). For $\alpha \geq 0$, this corresponds to choosing the model with highest  
\begin{equation}
R^{\alpha}_{\text{trust}}=\log (\overline{\tau_{\text{Trust}}}(\omega)) - \alpha\log(\Delta_{\tau} (\omega)).
\label{eq:selrulealpha}
\end{equation} \vskip -0.1 in

\item \textbf{\textsc{Report:}} To enhance transparency, our framework provides audit report templates (as seen in Supplementary Information Section  \ref{sec:AuditReport}) for communicating the audit results. These audit reports can be submitted to regulators or external third-party certifiers for validation.
\end{enumerate}

\paragraph{Controllable Trust Tradeoffs with TrustFormers} Trust constraints can be integrated in the training of Generative AI models. For example to ensure privacy of the  synthetic data we use differential private training \cite{dwork2014thealgorithmic} of the generative models with a privacy budget $\varepsilon$. Furthermore,  our trustworthiness index can be employed in an early stopping approach to select the model that best aligns with the desired trustworthiness profiles. Leveraging their adaptability in modeling various modalities, we integrate these trustworthy training and selection paradigms into generative transformer models, naming the resulting models TrustFormers. We denote the selected models as:
\[\text{TF}(\omega,\text{n-p})  \text{ for non-private training and }\]
\[\text{ TF}(\omega,\text{p-} \varepsilon)  \text{ for private training.}\]
If two trade-off weights $\omega_1$ and $\omega_2$ lead to the same checkpoints selection we use the following notation:
\[\text{TF}(\omega_1,\omega_2,\text{n-p})  \text{ for non-private training and }\] \[\text{ TF}(\omega_1,\omega_2,\text{p-} \varepsilon)  \text{ for private training.}\] \vskip -0.2in

\begin{table*}[htp!]
\centering
\begin{tabular}{@{}llccccccc@{}}
\toprule
Dimension & Metric & Polarity & Debiasing& Tabular & Time Series & NLP \\

 \midrule
\multirow{6}{*}{Fidelity} & \bf{\emph{Evaluated between $D_{r,train}$ and $D_{s}$:}} & & &&\\
& Maximum Mean Discrepancy (MMD) SNR \cite{kubler2022witness}&$ -1$&N/A& $ \cmark$ (\textbf{\textcolor{blue}{D}}/\textbf{\textcolor{blue}{E}}) & $\cmark$(\textbf{\textcolor{blue}{E}}) &$\cmark$(\textbf{\textcolor{blue}{E}}) \\
&MMD test$p$-value \cite{JMLR:v13:gretton12a} & $+1$ & N/A& $\cmark$ (\textbf{\textcolor{blue}{D}}/\textbf{\textcolor{blue}{E}}) & $\cmark$(\textbf{\textcolor{blue}{E}})&$\cmark$(\textbf{\textcolor{blue}{E}})\\
 & Fréchet Inception Distance (FID)\cite{heusel2017gans, semeniuta2019accurate} & $-1$ &N/A &$\cmark$ (\textbf{\textcolor{blue}{D}}/\textbf{\textcolor{blue}{E}}) & $\cmark$(\textbf{\textcolor{blue}{E}})&$\cmark$(\textbf{\textcolor{blue}{E}})\\
& Precision/ Recall \cite{kynkaanniemi2019improved} & $+1$ &N/A &$\cmark$ (\textbf{\textcolor{blue}{D}}/\textbf{\textcolor{blue}{E}}) & $\cmark$(\textbf{\textcolor{blue}{E}})&$\cmark$(\textbf{\textcolor{blue}{E}})\\
& Chi Squared& $-1$ & N/A &$\cmark$ (\textbf{\textcolor{blue}{D}}) &$\xmark$& $\xmark$ \\
 & $\ell_2$ mutual Information difference& $-1$ & N/A & $\cmark$ (\textbf{\textcolor{blue}{D}}) & $\xmark$ &$\xmark$ \\\midrule
\multirow{3}{*}{Privacy} &\bf{\emph{Evaluated between $D_{r,train}$ and $D_{s}$:}}& & &&\\
& Exact Replicas Count & $-1$ &N/A &$\cmark$ (\textbf{\textcolor{blue}{D}}) & $\xmark$ &$\xmark$\\
 & k-nearest neighbor median distance \cite{chen2020gan}& $+1$ &N/A & $\cmark$ (\textbf{\textcolor{blue}{D}}/\textbf{\textcolor{blue}{E}}) & $\cmark$(\textbf{\textcolor{blue}{E}})&$\cmark$(\textbf{\textcolor{blue}{E}})\\
 & k-nearest neighbor mean distance \cite{chen2020gan}& $+1$ &N/A & $\cmark$ (\textbf{\textcolor{blue}{D}}/\textbf{\textcolor{blue}{E}}) & $\cmark$(\textbf{\textcolor{blue}{E}})&$\cmark$(\textbf{\textcolor{blue}{E}})\\\midrule
\multirow{4}{*}{Utility } & \bf{\emph{Classifier trained on $D_{s}$} }& & &&\\
&\bf{\emph{Evaluated on $D_{r,val}$ @ validation and $D_{r,test}$@ test}}:& & &&\\
&Accuracy/ precision/ recall/ F1 score of:&& &&&\\
 & Linear Logistic Regression& $+1$ & $\xmark$ & $ \cmark$ (\textbf{\textcolor{blue}{D}}/\textbf{\textcolor{blue}{E}}) & $\cmark$ (\textbf{\textcolor{blue}{E}}) &$\cmark$ (\textbf{\textcolor{blue}{E}}) \\
 & Nearest Neighbor classification & $+1$ & $\xmark$ & $ \cmark$ (\textbf{\textcolor{blue}{D}}/\textbf{\textcolor{blue}{E}}) & $\cmark$ (\textbf{\textcolor{blue}{E}}) &$\cmark$ (\textbf{\textcolor{blue}{E}}) \\
 & MLP &$+1$ & $\xmark$ & $ \cmark$ (\textbf{\textcolor{blue}{D}}/\textbf{\textcolor{blue}{E}}) & $\cmark$ (\textbf{\textcolor{blue}{E}}) &$\cmark$ (\textbf{\textcolor{blue}{E}}) \\
 & MLP / Adversarial debiasing \cite{zhang2018mitigating} &$+1$ & $\cmark$ & $ \cmark$ (\textbf{\textcolor{blue}{D}}/\textbf{\textcolor{blue}{E}}) & $\cmark$ (\textbf{\textcolor{blue}{E}}) &$\cmark$ (\textbf{\textcolor{blue}{E}}) \\
 &MLP/ Fair Mixup \cite{chuang2021fair} & $+1$ & $\cmark$ & $ \cmark$ (\textbf{\textcolor{blue}{D}}/\textbf{\textcolor{blue}{E}}) & $\cmark$ (\textbf{\textcolor{blue}{E}}) &$\cmark$ (\textbf{\textcolor{blue}{E}}) \\ \midrule
\multirow{4}{*}{Fairness } & Applicable to all classifiers in utility:&& && \\
&\bf{\emph{Evaluated on $D_{r,val}$ @ validation and $D_{r,test}$@ test}}:& & &&\\
&Equal Opportunity Difference (absolute value)\cite{barocas-hardt-narayanan} & $-1$ & * & *& * & *\\
 & Average Odds Difference (absolute value) \cite{barocas-hardt-narayanan} & $-1$ & * & *& * & *\\
 & Equalized Odds Difference (absolute value) \cite{barocas-hardt-narayanan}& $-1$& * & *& * & *\\\midrule 
 \multirow{3}{*}{Robustness}& Applicable to all classifiers in utility:&& && \\
 &\bf{\emph{Evaluated on $D_{r,val}$ @ validation and $D_{r,test}$@ test}}:& & &&\\
& Adversarial Accuracy/ precision/ recall/ F1 score& $+1$ & * & *& * & * \\ 
 & Absolute Difference of Adversarial& $-1$ & * & *& * & * \\ 
 &and non adversarial utility metrics & &&&&\\
 \bottomrule
 
\end{tabular}%
\caption{Metrics and their associated polarities that are supported by our auditing framework under each trust dimension. Debiasing indicates if a utility classifier uses a debiasing technique. \textbf{\textcolor{blue}{D}} indicates that the metric is computed on the data space after quantization. \textbf{\textcolor{blue}{E}} indicates that the metric is computed in an embedding space. * refers to the same field values of the evaluated utility classifier. Note that our metrics are representative of each dimension and modality but are not exhaustive; other specialized metrics can be added and integrated seamlessly within our framework.}
\label{tab:metricMM}
\end{table*}

\begin{table*}
\begin{center}
\begin{tabular}{||c c c ||} 
 \hline
& $\omega= (\omega_f,\omega_P,\omega_U, \omega_F,\omega_{R})$&Interpretation \\ [0.5ex] 
 \hline\hline
 $\omega_{all}$ &$(100,100,100,100,100)/500$ & Equal Importance\\
 \hline
 $\omega_{e(PU)}$ &$(50,100,100,50,50)/350$ & Privacy/Utility Emphasis \\ 
 \hline
 $\omega_{e(PUF)}$ &$(50,100,100,100,50)/400$ & Privacy/Utility/Fairness Emphasis \\ 
 \hline
 $\omega_{U}$ &$(0,0,100,0,0)/100$ & Utility only \\ 
 \hline
$\omega_{PU}$ &$(0,100,100,0,0)/200$ & Privacy/Utility only \\
 \hline
 $\omega_{UF}$ &$(0,0,100,100,0)/200$ &Utility/Fairness only \\ 
 \hline
$\omega_{e(UF)r(R)}$ &$(50,50,100,100,0)/300$ &Utility/Fairness Emphasis No Robustness \\ 
 \hline
 $\omega_{UFR}$ &$(0,0,100,100,100)/300$ &Utility/Fairness/Robustness only \\ 
 \hline
 $\omega_{UR}$ &$(0,0,100,0,100)/200$ &Utility/Robustness only \\ 
 \hline
 $\omega_{PUR}$ &$(0,100,100,0,100)/300$ &Privacy/Utility/Robustness only \\ [1ex] 
 \hline
\end{tabular}
\captionof{table}{Examples of weights trade offs of trust dimensions reflecting priorities in auditing synthetic data. }
\label{Table:weights}
\end{center}
\end{table*}

\begin{table*}
\begin{center}
\resizebox{\textwidth}{!}{\begin{tabular}{lclc|clclclcl}
\toprule
Use Case & Dataset& Modality &Downstream Task &Safeguards & Policy Alignment Example\\
\midrule
\midrule
Banking &Bank Marketing\cite{bankmarketing} &Tabular & Campaign prediction & sensitive community (age); & Fair Lending Act \\
 & &&&user privacy; robustness & \\
 \midrule 
Recruitment & UK Recruitment\cite{recruitement} & Tabular & Employment prediction& sensitive community (ethnicity) & NYC Law 144 \\
 & &&&user privacy; robustness & \\
\midrule 
Education & Law School Admission& Tabular & Admission prediction & sensitive community (ethnicity);& Equal Educational Opportunity Act \\
 &Council Dataset \cite{lsac} & & & user privacy; robustness &\\
\midrule
\midrule
Financial Services & Credit Card \cite{altman2021synthesizing} & Tabular time-series & Fraud Detection & user privacy& Finance Regulation\\
Healthcare & MIMIC-III\cite{mimic3} & Tabular time-series & Mortality prediction & sensitive community (ethnicity);& Patient Protection 
and Affordable Care Act\\
 & &&& user privacy; robustness& \\
\midrule 
\midrule 
Healthcare & MIMIC-III Notes\cite{vanAken2021} & NLP & Mortality prediction & sensitive community (ethnicity)& Patient Protection 
and Affordable Care Act\\
 & &&& user privacy; robustness& \\
 \midrule 
\midrule 
Visual Recognition & Imagenet\cite{vanAken2021} & Vision &  classification & distribution shift& Robust Generalization\\
\hline
\end{tabular}}
\captionof{table}{Synthetic data use cases, safeguards and policy alignment.}
\label{Tab:usecases}
\end{center}
\end{table*}

\section*{Methods}


\paragraph{Trust Dimensions} We  start by giving precise definitions for the trust dimensions and their risks assessment that play a central role in our auditing framework:

\begin{itemize}
\item \emph{\textbf{Fidelity.}} Fidelity measures the quality of the synthetic data in terms of its closeness in distribution to the real data and its diversity in covering the multiple modes of the real data distribution \cite{alaa2022faithful,kubler2022witness,JMLR:v13:gretton12a,heusel2017gans,kynkaanniemi2019improved}.
\item \emph{\textbf{Privacy.}} Privacy assesses memorization and real data leakage to synthetic data. Membership inference attacks, such as nearest neighbor attacks, are instrumented to identify if an actual data point can be identified in the vicinity of a synthetic data point, thereby unveiling that the corresponding individual was a member of the real data training set \cite{vanbreugel2023membership,jagielski2020auditing,houssiau2022tapas}.
\item \emph{\textbf{Utility.}} Utility measures the accuracy and performance of a predictive downstream task, where predictive models are trained on the synthetic data and evaluated in terms of their predictive performance on real test data.\item\emph{\textbf{Fairness}.} Fairness has two aspects: the first is related to bias in the synthetic data \cite{calmon2017optimized}, and the second is the fairness of the predictions with respect to sensitive and protected communities evaluated on real test data points \cite{barocas-hardt-narayanan}.
\item \emph{\textbf{Robustness.}} Robustness refers to the accuracy of a predictive model trained on synthetic data and evaluated on real test points in the presence of imperceptible, worst-case adversarial perturbations. We use black box, greedy attacks on utility classifiers for tabular and time-series, as in \cite{ballet2019imperceptible,cartella2021adversarial,yang2020greedy,yang2020greedy,lei2019discrete,mathov2022not,agarwal2021black} (see Supplementary Information \ref{sec:robustness}).
\end{itemize}

\paragraph{Auditing Framework}The key steps of our auditing framework (Figure  \ref{Fig:flowchartM}) are explained below:

\begin{enumerate}
\item\textbf{\textsc{Evaluate:}} Given the specific synthetic data application and relevant policies and regulations, the governance expert and data science board collaboratively establish multiple quantitative metrics for each trust dimension. These metrics serve to evaluate the synthetic data's adherence to the requirements necessary for meeting socio-technical safeguards within each dimension. In  Table \ref{tab:metricMM}, a comprehensive set of metrics is presented for each dimension, chosen to strike a balance between interpretability and risk assessment. It's important to note that while these metrics represent each dimension and modality, they are not exhaustive. Our framework can seamlessly accommodate additional specialized metrics as needed. These metrics are then assessed by the data scientist.\\

\item\textbf{\textsc{Aggregate:}} 
The interpretation and communication of these metrics within each dimension pose a significant challenge. Dealing with numerous metrics with different polarities and dynamic ranges can be overwhelming for internal reviewers and regulators. In social sciences, it's common to aggregate metrics into a single score or index \cite{greco2019methodological}. Indices serve as powerful tools for simplifying complex information into an accessible format that can be interpreted and  understood by a wide range of stakeholders. They facilitate straightforward communication, enabling easy comparisons and benchmarking.  We address the issues related to varying ranges and polarities and aggregate the metrics within each dimension into a   \textbf{\emph{trust dimension index}} that falls within the range of 0 to 1, where 0 signifies poor conformity with the dimension requirements, and 1 indicates high conformity. This trust dimension index can be interpreted as a measure of compliance probability. Our aggregation method relies on the copula technique \cite{copula-aggregation} which provides us with this intuitive probabilistic interpretation. In Supplementary Information \ref{app:descFramework} we explain the copula method that consists in normalizing metrics under trust dimension using global CDFs estimated across synthetic data, and followed by a geometric mean. \\

\item\textbf{\textsc{Re-weight:}} Each synthetic dataset is now represented by trust dimension indices, denoted as $\pi_{T}$, where "T" corresponds to Fidelity, Privacy, Utility, Fairness, or Robustness. Considering the specific application, associated policies, and desired trade-offs between trust dimensions, internal reviewers collaborate with governance experts and the application owner to establish the \textbf{\emph{trustworthiness profile}}. This profile is expressed through  \textbf{\emph{tradeoff weights}}, symbolized as $\omega_{T}$, which indicate the relative importance of the trust indices necessary to fulfill performance and policy requirements. For instance, when training downstream tasks on-site, privacy may not be a necessity for the synthetic data, but it becomes essential for the predictive model. However, when conducting training in a public cloud environment, ensuring the privacy of synthetic data is mandatory. These varying requirements lead to different trustworthiness profiles with distinct trade-offs between privacy and utility for the synthetic data. Detailed examples of other possible trustworthiness profiles can be found in  Table \ref{Table:weights}.\\

\item \textbf{\textsc{Index by Trustworthiness:}} The trust dimension indices  are subsequently re-weighted  
by the trade-off weights and combined to produce the final \textbf{\emph{trustworthiness index}} ($\tau_{\text{trust}}(\omega)$) of the synthetic data. It's important to emphasize that this index is context-specific, contingent upon the application's requirements and the specified safeguards and trustworthiness profiles. Recalling that the trust dimension indices can be interpreted as probabilities, we define the trustworthiness index as a weighted geometric mean of the  dimension indices:
\begin{equation}
 \tau_{\text{Trust}}(\omega)= \exp\left(\sum_{T } \omega_{T} \log(\pi_{T})\right), 
 \end{equation}
 The choice of a geometric mean is preferred because it embodies an "and" operation interpretation, unlike the arithmetic mean, which implies an "or" interpretation.\\

\item \textbf{\textsc{Rank:}} With the defined trustworthiness profile, the trustworthiness index can serve multiple purposes. The internal reviewer can utilize it to establish a \textbf{\emph{ranking}} for various synthetic datasets generated by different models, enabling certification and validation of their adherence to specific requirements. Simultaneously, data scientists can leverage the trustworthiness index for \textbf{\emph{model selection}} within a given generation method and trustworthiness profile. When determining our trustworthiness index ranking, we can additionally consider the uncertainty in real data splits. For a trustworthiness profile $\omega$, our preference is  for a generative AI technique that exhibits the highest average trustworthiness index across splits ($\overline{\tau_{\text{Trust}}}(\omega)$) while minimizing volatility ($ \Delta_{\tau}(\omega)$). For $\alpha \geq 0$, this corresponds to choosing the model with highest  
\begin{equation}
R^{\alpha}_{\text{trust}}=\log (\overline{\tau_{\text{Trust}}}(\omega)) - \alpha\log(\Delta_{\tau} (\omega)).
\tag{\ref{eq:selrulealpha}}
\end{equation}

\item \textbf{\textsc{Report:}} To promote transparency and accountability, our framework defines templates for communicating auditing results in the form of an \textbf{\emph{audit report}}. An example of the audit report is given in Supplementary Information \ref{sec:AuditReport}. The audit report can be submitted to a regulator or an external third-party certifier that probe the validity of the conclusions of the internal audit report.\\
\end{enumerate}

\paragraph{Auditing  Workflows for Transparency and Accountability: Insights and Limitations}
Our work aligns closely with the core principles of AI auditing, as underscored in both the FAccT (Fairness, Accountability, and Transparency) and STS (Science and Technology Studies) literature, which delve into issues related to race, gender, bias, and fairness \cite{AlgorithmicFairness}. For instance, the study by Buolamwini and Gebru \cite{pmlr-v81-buolamwini18a} highlighted the  need for auditing racial and gender biases in facial recognition. In our holistic auditing approach, we address multiple dimensions, mirroring the discussions on intersectional biases frequently explored in these literatures \cite{hancock_2007}. While our audits are centered on specific technical and societal aspects, it's important to note that both the FAccT and STS literature encompass a broader spectrum of topics, including governance, closing the  accountability gap \cite{AccountabilityGap} , ethics, and the far-reaching societal implications of emerging technologies  \cite{strathern2000audit,AuditSociety}. We  discuss here a real-time platform that operationalizes our synthetic data auditing framework and further embraces the audit culture advocated in the  the FAccT  and STS literature in terms of  accountability, transparency and governance workflows. Our auditing platform connects different stakeholders from governance experts, to data scientists, to internal reviewers, to external certifiers or regulators.\\

\noindent We envision workflows for interactions between these different personas via the auditing platform. Figure \ref{fig:workflowM} summarizes our vision: governance experts define the intended use of synthetic data, the safeguards for compliance with regulations and policies, and the acceptable trade-offs between these safeguards. Next, the data scientist develops models and configures auditing
tasks to rank developed models, perform model selection, and ensure compliance with the safeguards. Internal reviewers also have access to the platform, verify the compliance of models and created data with prescribed policies and safeguards and approve / reject models' deployment and synthetic data usage. Finally, a portable audit report is generated on the fly within the platform, which can be submitted to external third-party certifiers that probe the validity of the conclusions of the internal audit report. \footnote{Snippets of such auditing workflows can be found in \cite{workflows}} \\

\noindent We believe that transparent reporting should become a \emph{de facto} part of any AI application, model, or data (real or synthetic). We demonstrated how transparencies could be created within our framework both for internal testing and validation and for external auditing or certification. While our framework helps connect various key players, there is a need for additional organization measures, playbooks, and governance practices to harmonize and orchestrate such workflows.
Another challenge in algorithmic auditing is the interpretable communication of how the technical metrics we compute relate to policy and legislation. To address this challenge, we adopt messages and warnings for detecting biases and harms to communicate auditing findings to policy experts. We envision a future auditing workflow that uses policy packs, which for a given application and set of policies, define templates for parameters, thresholds, technical metrics, and explanations.

\begin{table*}
\begin{center}
\resizebox{\textwidth}{!}{\begin{tabular}{ l|l }
\toprule
Generative AI Method & Trust Constraint\\
\midrule
\textbf{Non-Private \& Private TrustFormers}& \\
(Conditional) TrustFormer GPT (non-private):  \textbf{TF($\omega$,n-p)} & Fidelity/Utility \\
(Conditional) TrustFormer GPT Trained with Differential Private SGD : \textbf{TF($\omega$,p-$\varepsilon$)}&Fidelity/Privacy Preserving synthetic Data/Utility\\
Differential Private Sampling From non-private (Conditional)TrustFormer \cite{majmudar2022differentially,NEURIPS2021_f2b5e92f} & Fidelity/ Privacy Preserving synthetic Data/utility \\

\midrule
\textbf{Non-private Baselines} & \\ Gaussian Copula \cite{SDV} \textbf{Gaussian Copula (n-p)}& Fidelity \\
 Conditional Tabular GAN\cite{SDV}\textbf{CTGAN(np)}& Fidelity/utility\\
\midrule
 \textbf{Private Baselines} & \\
Differential Private Probabilistic Graphical Model \cite{mckenna2021winning} \textbf{DP-PGM(p-$\varepsilon$)}& Fidelity/Privacy Preserving synthetic Data\\ 
DP-PGM (targeted)\cite{mckenna2021winning} \textbf{DP-PGM(target,p-$\varepsilon$)}& Fidelity/Privacy Preserving synthetic Data/Utility\\ 
(Conditional) Differential Private-GAN \cite{xie2018differentially,qian2023synthcity} \textbf{DP-GAN(p-$\varepsilon$)} & Fidelity/ Privacy Preserving synthetic Data/Utility \\
(Conditional) PATE-GAN \cite{yoon2018pategan,qian2023synthcity}\textbf{PATE-GAN(p-$\varepsilon$)}& Fidelity/Privacy Preserving synthetic Data/Utility\\
\hline
\end{tabular}}
\captionof{table}{Generative Models and TrustFormer Models audited within our framework. For private models we consider privacy budgets $\varepsilon=1$ or $3$.}
\label{Table:baselines}
\end{center}
\end{table*}

In the following sections we present the training and auditing of various data generation techniques, including our TrustFormer Models, across a range of use cases encompassing tabular, time series, and natural language data (refer Table \ref{Tab:usecases}). We also conduct audits on synthetic data generated by state-of-the-art diffusion models, particularly in the computer vision domain, focusing on the ImageNet classification task and assessing their generalization under distribution shifts. The models under audit are summarized in  Table \ref{Table:baselines}, and we employ our trust dimension indices to assess each model's compliance with specific trust dimensions. For a given trustworthiness profile weights $\omega$ (refer to Table \ref{Table:weights} for examples and their interpretation), we rely on our trustworthiness index to conduct TrustFormer model selection and rank the models based on their alignment with predefined safeguards (given in the trustworthiness profile $\omega$).

\section*{Tabular use cases}
~\\
In this Section, we showcase our auditing framework and trustworthiness index driven model selection for TrustFormer on three tabular datasets: Bank Marketing Dataset\cite{bankmarketing}, Recruitment Dataset \cite{recruitement}, and the Law School Admission Council Dataset \cite{lsac}. Please refer to  Table \ref{Tab:usecases} for details of the datasets, downstream tasks , safeguards and TF models training details.

\paragraph{Setup} In our auditing setup, real data serves as the baseline, and we emphasize the contrast with synthetic generated data. The real data is split into training ($D_{r,\text{train}}$), validation ($D_{r,\text{val}}$), and test ($D_{r,\text{test}}$) sets through random sampling without replacement, repeated five times with different seeds, creating five real data folds ($\mathcal{D}_r$). Each generative AI method listed in Table \ref{Table:baselines} is trained five times independently on each real data fold. For TrustFormers, both non-private (TF(n-p)) and private (TF(p-$\varepsilon=1$) or TF(p-$\varepsilon=3$)) versions are trained. Trustworthiness Index Driven Model selection is performed independently within each fold, leading to the selection of a checkpoint $t^*$.  Synthetic datasets are generated by sampling from each generator trained on a specific fold. This results in $\mathcal{D}_s=\{D^{\ell}_s\}_{\ell=1}^5$. Audits  on downstream tasks, assess utility, fairness, and robustness evaluated on the corresponding real test data folds. A non-private tabular RoBERTa Embedding ($\textbf{E}$) \cite{padhi2021tabular}, trained on the real data $D_r$ (excluding task labels), is used for certain metrics  in  Table \ref{tab:metricMM}.

\paragraph{Real Data, Downstream Tasks, and Protected communities}
The \emph{Bank Marketing Dataset} \cite{bankmarketing} has 45211 samples. Each sample is a row with 17 fields representing a client. The downstream task is to predict if the client will subscribe to a term deposit or not. Protected groups include individuals with value of field \texttt{age} less than 30.
The \emph{Recruitment Dataset} \cite{recruitement} contains 6000 samples consisting of rows with 14 demographic fields. The classification task associated to this dataset is the prediction of a candidate's employment. The sensitive variable is defined as the binary indicator \texttt{white}. Finally, the \emph{Law School School Admission Council Dataset} \cite{lsac} has 20461 samples of with 11 fields of demographic features. The downstream task is the prediction of whether a candidate  was admitted to law school or not. The sensitive variable is  the binary indicator \texttt{black}.

\begin{figure}[htp]
\centering
\includegraphics[width=\linewidth]{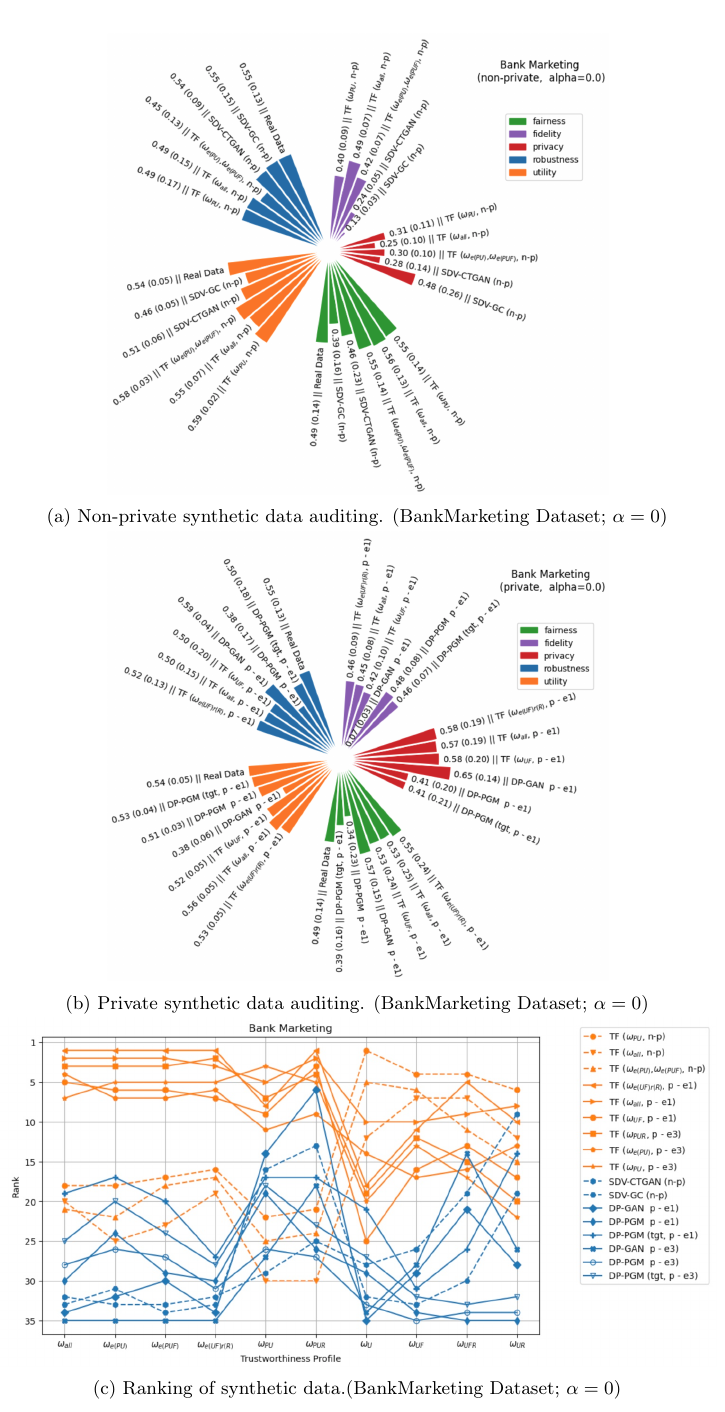} 
\caption{Summary of auditing and ranking results on the Bank Marketing dataset using the trustworthiness index given in \eqref{eq:selrulealpha}  for $\alpha=0$. (a) and (b) show trust dimension indices $\pi_{T}$( where "T" corresponds to Fidelity, Privacy, Utility, Fairness, or Robustness), and their "variance" ($\Delta_{T}$) on TrustFormer (TF) and baseline models. The format is $\pi_{T} (\Delta_{T}) ||$ Name of the synthetic data model. (c) shows the ranking of the models across different trustworthiness profiles $\omega$  given in Table \ref{Table:weights}.}
\label{Fig:BankMarketing_0}
\end{figure}

\begin{figure}[htp]
\centering
\includegraphics[width=\linewidth]{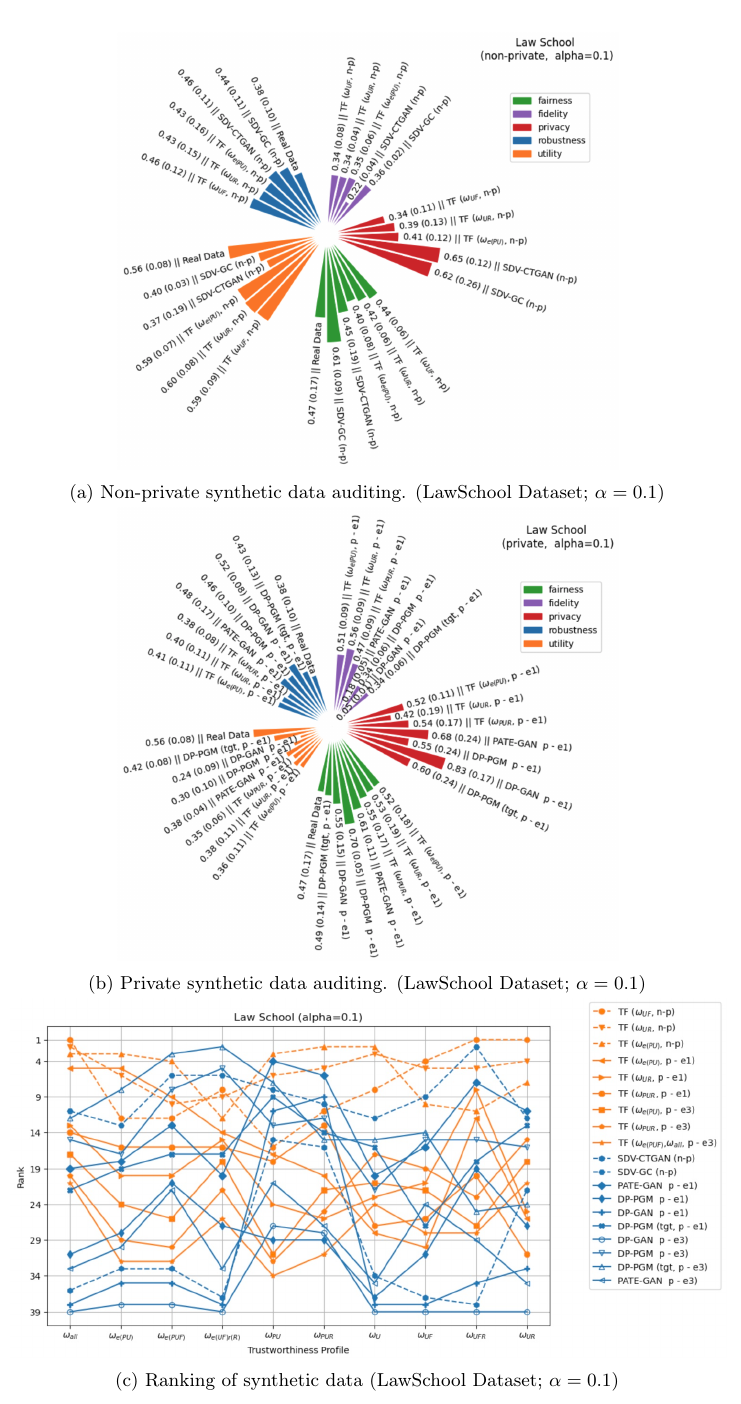} 
\caption{Summary of auditing and ranking results on the Bank Marketing dataset using the trustworthiness index given in \eqref{eq:selrulealpha}  for $\alpha=0.1$. This means that the uncertainty is taking into account in both model selection and in ranking the synthetic data using the trustworthiness index. (a) and (b) show trust dimension indices $\pi_{T}$ (where `T' corresponds to Fidelity, Privacy, Utility, Fairness, or Robustness), and their ``variance'' ($\Delta_{T}$) on TrustFormer (TF) and baseline models. The format is $\pi_{T} (\Delta_{T}) ||$ Name of the synthetic data model. (c) shows the ranking of the models across different trustworthiness profiles $\omega$  given in Table \ref{Table:weights}.\\
}
\label{Fig:LawSchool_alpha}
\end{figure}

\begin{figure}[htp!]
\centering
\includegraphics[width=\linewidth]{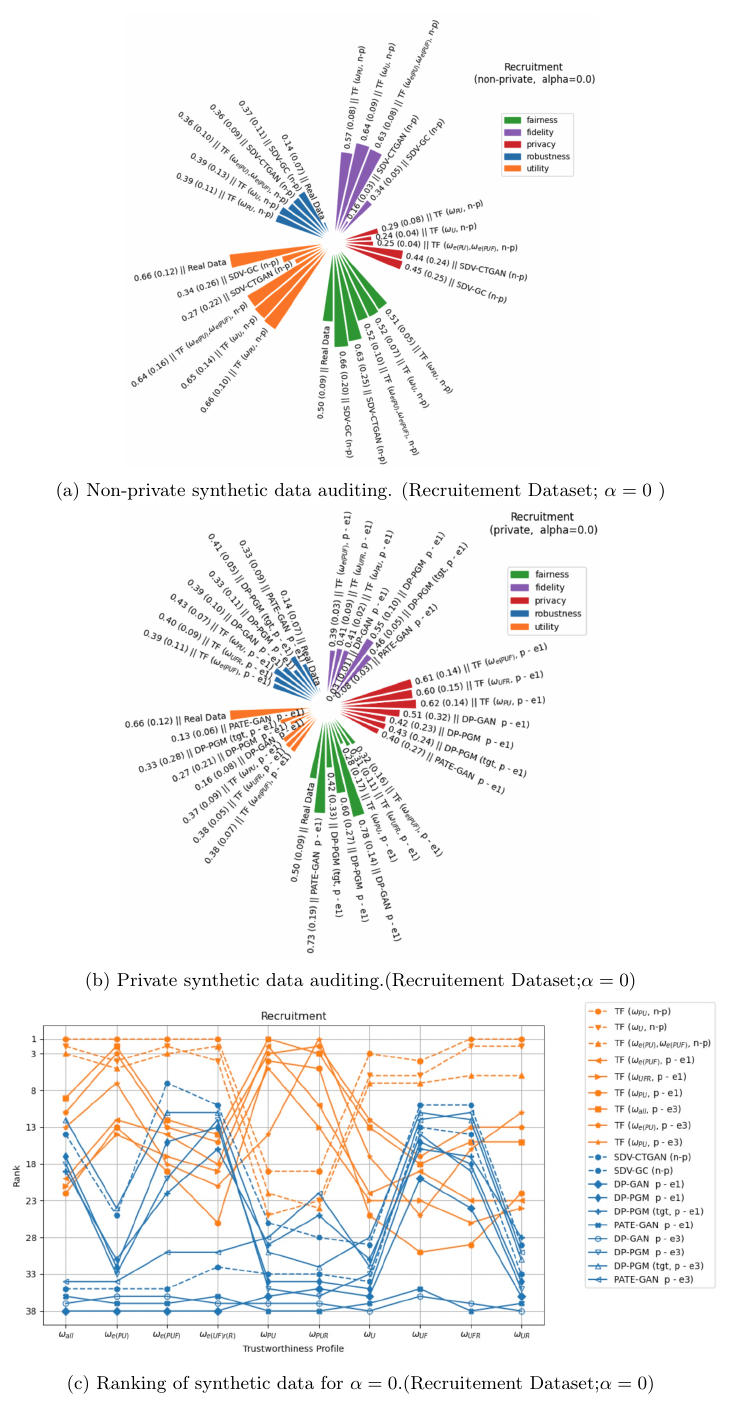} 
\caption{Summary of auditing and ranking results on the recruitment dataset using the trustworthiness index given in \eqref{eq:selrulealpha}  for $\alpha=0$. (a) and (b) show trust dimension indices $\pi_{T}$ (where "T" corresponds to Fidelity, Privacy, Utility, Fairness, or Robustness), and their "variance" ($\Delta_{T}$) on TrustFormer (TF) and baseline models. The format is $\pi_{T} (\Delta_{T}) ||$ Name of the synthetic data model. (c) shows the ranking of the models across different trustworthiness profiles $\omega$  given in Extended Table \ref{Table:weights}.}
\label{Fig:Recruitement_0}
\end{figure}

\paragraph{Audit Results} For the Bank Marketing Dataset, we present average trust dimension indices and their "variances" for non-private and private synthetic data generated by TrustFormers models (TF) in Figure \ref{Fig:BankMarketing_0} (a) and (b). In this case, TF models were selected across various trustworthiness profiles (as outlined in Table \ref{Table:weights}) using the trustworthiness index from Equation \eqref{eq:selrulealpha} for $\alpha=0$. We also include results for non-private and private baselines (refer to  Table \ref{Table:baselines}), shown in panels (a) and (b). Additionally, we provide information about the utility, fairness, and robustness of downstream tasks trained on real data as a reference for both panels (a) and (b). Panel (c) displays the ranking according to trustworthiness index  of synthetic data across different trustworthiness profiles based on the same $\alpha$ value. Similar plots for the Recruitment and Law School datasets can be found in  Figure \ref{Fig:Recruitement_0}  and in Supplementary Information \ref{app:additionalResults} Figure \ref{Fig:LawSchool_0}, respectively, for $\alpha=0$. To account for data split uncertainties, we've included audit results for $\alpha=0.1$, depicted in Figure \ref{Fig:LawSchool_alpha} for the Law School dataset, and in the Supplementary Information for the other datasets (see Figures \ref{Fig:Recruitement_alpha} and \ref{Fig:BankMarketing_alpha} in Supplementary Information \ref{app:additionalResults}). 

 \paragraph{Discussion of the Audit  Results: Trust Dimension Tradeoffs and Mitigations (Panels (a) and (b))} Analyzing these results we make the following  observations. First, when compared with real data, TF synthetic data demonstrates on par or superior performance in utility, fairness, and robustness indices across various datasets, while effectively balancing privacy, fidelity, and other trust dimensions. A careful trustworthiness index model selection of synthetic data, in conjunction with classical classifier model selection, facilitates alignment with prescribed safeguards without compromising performance. Second, the selection of synthetic data for TrustFormer models guided by the trustworthiness index consistently leads to competitive performance with all other baseline methods across trust dimension indices as  can be seen in panels (a) and (b) in all aforementioned figures. TrustFormers either achieve the highest index on each trust dimension or rank among the top-performing synthetic data, irrespective of the specific generative methods considered. Third, a distinction is observed between non-private and differentially private synthetic data, with fidelity and utility exhibiting higher indices in non-private data and privacy, fairness, and robustness showing higher indices in differentially private data. This aligns with the well-documented trade-offs in fairness, utility, privacy, and robustness in the literature \cite{sehwag2022robust,tradeoffs}.

 \paragraph{Discussion of the  Ranking  Results  and Controllable Tradeoffs (Panels (c))} For both Bank Marketing dataset and  Recruitment datasets we see that in both cases $\alpha=0$ and $0.1$, TF synthetic data outperforms other baselines across all trustworthiness profiles $\omega$ in terms of its trustworthiness index . We see in Panel (c) Figure \ref{Fig:BankMarketing_0} and  Figure \ref{Fig:Recruitement_0}, a clear phase transition between private and non private TF models depending on wether the trustworthiness profile highlights privacy as a requirement or not. This effect is achieved through our trustworthiness index model selection that offers control over trust trade-offs. On the other hand, for the law school dataset, TF models fall short  with respect to baselines when uncertainty is not considered ($\alpha=0$, Supplementary Figure \ref{Fig:LawSchool_0}),  but  lead to top performing synthetic data when uncertainty is considered ($\alpha=0.1$, Figure \ref{Fig:LawSchool_alpha}). This highlights the importance of assessing the uncertainty in auditing synthetic data.

\begin{table*}[htp!]
\resizebox{1\textwidth}{!}{
\begin{tabular}{lcccHHcHHcHH}
\toprule
 Model & Fidelity &Privacy &Utility &Utility & Utility & Fairness & Fairness & Fairness & Robustness & Robustness & Robustness\\

\midrule
\textbf{Non-private TrustFormer}&&& & & && &&& \\
TF ($\omega_{UR}$, n-p, mn) & 0.70  & 0.11  & 0.57  & 0.60  & 0.51  & 0.40  & 0.32  & 0.65  & 0.55  & 0.57  & 0.52  \\
 TF ($\omega_{e(PUF)}$,$\omega_{U}$,$\omega_{UF}$,$\omega_{e(UF)r(R)}$,$\omega_{all}$,$\omega_{UFR}$, n-p, mn) & 0.37  & 0.36  &\textcolor{blue}{\textbf{ 0.67}}  & 0.66  & 0.69  & 0.42  & 0.50  & 0.30  & 0.50  & 0.49  & 0.53  \\
 TF ($\omega_{e(PU)}$,$\omega_{PUR}$, n-p, mn) & 0.53  & 0.21  & 0.51  & 0.51  & 0.50  & 0.29  & 0.25  & 0.38  & 0.49  & 0.59  & 0.33  \\
 TF ($\omega_{PU}$, n-p, mn) & 0.81  & 0.42  & 0.56  & 0.53  & 0.61  & 0.25  & 0.29  & 0.18  & 0.39  & 0.37  & 0.46  \\
 TF ($\omega_{UR}$, n-p, topk) & 0.70  & 0.21  & 0.52  & 0.54  & 0.48  & 0.43  & 0.42  & 0.44  & 0.60  & 0.68  & 0.47  \\
 TF ($\omega_{UF}$,$\omega_{UFR}$, n-p, topk) & \textbf{\textcolor{blue}{0.86}}  & 0.49  & 0.61  & 0.55  & 0.74  & \textbf{0.44}  & 0.38  & 0.60  & 0.52  & 0.70  & 0.28  \\
 TF ($\omega_{U}$, n-p, topk) & 0.62  & 0.24  & 0.62  & 0.64  & 0.58  & 0.32  & 0.33  & 0.29  & \textcolor{blue}{\textbf{0.69}}  & 0.82  & 0.49  \\
TF ($\omega_{e(PU)}$,$\omega_{e(PUF)}$,$\omega_{PU}$,$\omega_{e(UF)r(R)}$,$\omega_{all}$,$\omega_{PUR}$, n-p, topk) & 0.77  & \textbf{0.54}  & 0.60  & 0.55  & 0.72  & 0.29  & 0.29  & 0.31  & 0.64  & 0.76  & 0.44  \\
 
\midrule 
\textbf{Private TrustFormer}&&& & & && &&& \\
 TF  ($\omega_{e(PU)}$,$\omega_{e(PUF)}$,$\omega_{PU}$,$\omega_{e(UF)r(R)}$,$\omega_{all}$, p - $\epsilon=$3, mn) & 0.34  & 0.93  & 0.30  & 0.32  & 0.26  & 0.58  & 0.59  & 0.56  & 0.41  & 0.29  & 0.81  \\
 TF  ($\omega_{U}$,$\omega_{UF}$,$\omega_{UFR}$,$\omega_{UR}$,$\omega_{PUR}$, p - $\epsilon=$3, mn) & 0.19  & 0.69  & \textbf{0.44}  & 0.45  & 0.41  & 0.55  & 0.46  & 0.80  & 0.47  & 0.49  & 0.44  \\
 TF  ($\omega_{PU}$, p - $\epsilon=$3, topk) & 0.15  &\textcolor{blue}{\textbf{ 1.00}}  & 0.22  & 0.22  & 0.22  & \textcolor{blue}{\textbf{0.83}}  & 0.81  & 0.88  & 0.46  & 0.37  & 0.70  \\
 TF  ($\omega_{PUR}$, p - $\epsilon=$3, topk) & 0.35  & 0.89  & 0.40  & 0.42  & 0.36  & 0.46  & 0.49  & 0.40  & \textbf{0.50}  & \textbf{0.50}  & 0.52  \\
 TF  ($\omega_{e(PU)}$,$\omega_{e(PUF)}$,$\omega_{UF}$,$\omega_{all}$, p - $\epsilon=$3, topk) & 0.40  & 0.81  & 0.33  & 0.34  & 0.30  & 0.57  & 0.49  & 0.74  & 0.29  & 0.23  & 0.45  \\
 TF  ($\omega_{e(UF)r(R)}$, p - $\epsilon=$3, topk) & 0.34  & 0.75  & 0.26  & 0.29  & 0.21  & 0.31  & 0.52  & 0.11  & 0.40  & 0.35  & 0.52  \\
 TF  ($\omega_{UFR}$,$\omega_{UR}$, p - $\epsilon=$3, topk) & 0.42  & 0.62  & 0.41  & 0.36  & 0.52  & 0.60  & 0.65  & 0.51  & 0.40  & 0.41  & 0.39  \\
 TF  ($\omega_{U}$, p - $\epsilon=$3, topk) & \textbf{0.64}  & 0.12  & 0.41  & 0.41  & 0.40  & 0.59  & 0.60  & 0.57  & 0.34  & 0.30  & 0.44  \\
\midrule 
\midrule
 Real Data & N/A & N/A & 0.44  & 0.38  & 0.57  & 0.07  & 0.02  & 0.57  & 0.67  & 0.69  & 0.62  \\

\bottomrule
\end{tabular}
}
\renewcommand{\tablename}{Table}
\captionof{table}{MIMIC-III/ In-Hospital Mortality downstream task evaluation: trust dimension indices of TrustFormer models. In bold highest index within each group of synthetic data. In blue highest value across all methods including real data.  }
\label{table:mimic_metrics}
\end{table*}

\begin{table*}[htp!]
\resizebox{1\textwidth}{!} {
\begin{tabular}{lrrrrrrrrrr}
\toprule
 Model &$\omega_{all}$ &$\omega_{e(PU)}$ &$\omega_{e(PUF)}$ &$\omega_{e(UF)r(R)}$ &$\omega_{PU}$ &$\omega_{PUR}$ &$\omega_{U}$ &$\omega_{UF}$ &$\omega_{UFR}$ &$\omega_{UR}$ \\
\midrule
\textbf{Non-Private TrustFormer}&&& & & && &&& \\
TF ($\omega_{UR}$, n-p, mn) & 13 & 15 & 15 & 14 & 15 & 15 & 5 & 6 & 5 & 6 \\
 TF ($\omega_{e(PUF)}$,$\omega_{U}$,$\omega_{UF}$,$\omega_{e(UF)r(R)}$,$\omega_{all}$,$\omega_{UFR}$, n-p, mn) & 8 & 7 & 9 & 5 & 8 & 5 &\rankone{1}&\rankone{1}&\rankone{1}&\rankthree{3}\\
 TF ($\omega_{e(PU)}$,$\omega_{PUR}$, n-p, mn) & 15 & 14 & 14 & 15 & 14 & 14 & 8 & 14 & 12 & 7 \\
 TF ($\omega_{PU}$, n-p, mn) & 9 & 9 & 10 & 10 & 9 & 10 & 6 & 15 & 14 & 8 \\
 TF ($\omega_{UR}$, n-p, topk) & 7 & 12 & 12 & 9 & 13 & 13 & 7 & 7 & 4 & 4 \\
 TF ($\omega_{UF}$,$\omega_{UFR}$, n-p, topk) &\rankone{1}&\rankone{1}&\rankone{1}&\rankone{1}& 4 &\rankthree{3}&\rankthree{3}&\ranktwo{2}&\ranktwo{2}& 5 \\
 TF ($\omega_{U}$, n-p, topk) & 6 & 10 & 11 & 11 & 12 & 7 &\ranktwo{2}& 8 &\rankthree{3}&\rankone{1}\\
TF ($\omega_{e(PU)}$,$\omega_{e(PUF)}$,$\omega_{PU}$,$\omega_{e(UF)r(R)}$,$\omega_{all}$,$\omega_{PUR}$, n-p, topk) &\ranktwo{2}&\ranktwo{2}&\rankthree{3}&\rankthree{3}&\ranktwo{2}&\rankone{1}& 4 & 12 & 7 &\ranktwo{2}\\

\midrule
\textbf{Private TrustFormer}&&& & & && &&& \\
 TF  ($\omega_{e(PU)}$,$\omega_{e(PUF)}$,$\omega_{PU}$,$\omega_{e(UF)r(R)}$,$\omega_{all}$, p - $\epsilon=$3, mn) & 5 & 5 & 5 & 7 & 5 & 6 & 14 & 13 & 13 & 13 \\
 TF  ($\omega_{U}$,$\omega_{UF}$,$\omega_{UFR}$,$\omega_{UR}$,$\omega_{PUR}$, p - $\epsilon=$3, mn) & 11 & 8 & 7 & 8 &\rankthree{3}& 4 & 9 & 4 & 6 & 9 \\
 TF  ($\omega_{PU}$, p - $\epsilon=$3, topk) & 12 & 11 & 8 & 12 & 10 & 9 & 16 & 10 & 10 & 15 \\
 TF  ($\omega_{PUR}$, p - $\epsilon=$3, topk) &\rankthree{3}&\rankthree{3}&\ranktwo{2}& 6 &\rankone{1}&\ranktwo{2}& 12 & 11 & 9 & 10 \\
 TF  ($\omega_{e(PU)}$,$\omega_{e(PUF)}$,$\omega_{UF}$,$\omega_{all}$, p - $\epsilon=$3, topk) & 10 & 6 & 6 & 4 & 6 & 11 & 13 & 9 & 15 & 16 \\
 TF  ($\omega_{e(UF)r(R)}$, p - $\epsilon=$3, topk) & 14 & 13 & 13 & 16 & 11 & 12 & 15 & 16 & 16 & 14 \\
 TF  ($\omega_{UFR}$,$\omega_{UR}$, p - $\epsilon=$3, topk) & 4 & 4 & 4 &\ranktwo{2}& 7 & 8 & 10 &\rankthree{3}& 8 & 11 \\
 TF  ($\omega_{U}$, p - $\epsilon=$3, topk) & 16 & 16 & 16 & 13 & 16 & 16 & 11 & 5 & 11 & 12 \\
 \bottomrule
\end{tabular}}
\renewcommand{\tablename}{Table}
\captionof{table}{MIMIC-III synthetic dataset ranking using the trustworthiness index corresponding to the trade-off weight $\omega$. We see that two models stand out across different trade-offs and they correspond to different decoding strategies. }
\label{table:mimic_ranking}
\end{table*}

\section*{Time-Series  Use Cases }

We  audit in this Section the use of time series synthetic data  in healthcare focusing on the utility/fairness tradeoffs. We also audit its use  in a financial application, fraud detection, with a focus on utility/privacy tradeoffs. 

\paragraph{Use Case I:  MIMIC-III Controllable Trust trade-offs on Healthcare Data} We explore  in this use case the promise of synthetic data in  the highly regulated healthcare domain, where patient  privacy and anti-discrimination regulations are enforced by law. This prohibits hospitals from sharing data in order to not expose the patients personal information. Moreover recent studies \cite{mimic-fairness} showed on the  MIMIC-III (Medical Information Mart for Intensive Care) time-series benchmark \cite{mimic3} that it has an inherent bias and discrimination \cite{mimic-fairness,mimic-IF,chen2018my}. We explore controllable trust trade-offs on synthetic times-series data obtained from learned TrustFormer models on this dataset.

\noindent \paragraph{MIMIC-III dataset.}  MIMIC-III (Medical Information Mart for Intensive Care) dataset\cite{mimic3} is a large database of about 40K patients with de-identified records collected during their stay in intensive care unit (ICU). The records contain high temporal resolution data including lab results, electronic documentation, and bedside monitor trends collected every hour. For each admission, we have an entry every hour of vitals measurement for a total of 48 entries capturing the dynamics in a patient state. Each hour, we have about 18 columns of vitals measurements augmented with a time-stamp, subject ID, and information of gender, ethnicity, and age. For complete in-depth description of the data, please refer to the MIMIC-III extensive documentation\cite{mimic3}.

\paragraph{Downstream Task} The In-Hospital Mortality (IHM) prediction task aims at predicting the mortality of patients in the ICU after a 48-hour stay. Given patients vitals evolution over the course of 48 hours the goal is to predict potential mortality of each patient. The data is therefore a time-series of measurements leading to a classification decision: did the patient expire or not. The training/val/test sets are composed of 14681/3222/3236 admissions respectively. Several studies on the MIMIC-III dataset, pointed the unfairness inherent to this dataset, disfavoring patients based on their ethnicity. 

\paragraph{Synthetic Data with Controllable Trust Trade-offs on MIMIC-III} In order to provide controllable trust trade-offs, we trained Tabular time-series GPT models\cite{padhi2021tabular} with regular and private differential training for a privacy budget $\varepsilon=3$ (See Supplementary Information for details on data preparation, model architecture and training hyper-parameters). For a trade-off weight $\omega$, we use our trustworthiness index  cross-validation to align the models with desired trust trade-offs. This results with the selected TrustFormers models: TF($\omega$, n-p) and TF($\omega$, p-$\varepsilon=3$). For inference from these models we used multinomal decoding (mn) or top-k decoding (sampling from top-k softmaxes for k=50), and refer to resulting trustformer models as: TF($\omega$, n-p, mn/top-k) and TF($\omega$, p-$\varepsilon=3$, mn/top-k). In order to audit this time-series dataset, we train an embedding $\textbf{E}$  that is a  TabRoBERTa model \cite{padhi2021tabular}. A masked language model is trained to predict masked fields from the patient vital records (masking 10\% of fields).  The vital records contains all the measurements over the 48-hour stay, we exclude patient IDs and labels from the TabRoBERTa training.

\paragraph{Trust Dimension Indices} Table \ref{table:mimic_metrics} summarizes the trust dimension indices of selected TrustFormers models, where the downstream tasks are evaluated on the real test set.  Interestingly similar to tabular data, we see that TrustFormer synthetic data outperforms real data on all trust dimensions. Interestingly the fairness index of real data is the lowest, and synthetic data  therefore improves the fairness/utility tradeoff . Similar observations on the relationship between trust constraints and trust trade-offs we made on tabular data hold for the time series case.

\noindent \paragraph{Analysis of Results} We compare the performance of the two decoding strategies considered (multinomial and top-k decoding) for synthetic data generation from TF models  and study how it impact trust trade-offs.   Note that we used herein fixed data splits from the literature and we don't report therefore the uncertainty of the audit.
Table \ref{table:mimic_ranking} gives the ranking of these synthetic datasets using the trustworthiness index for all trustworthiness profiles. We see  that for some trustworthiness profiles,  non-private TrustFormer  with multinomial decoding stands out, while top-k decoding outperforms it for other profiles. On private models we see that TF models with top-k decoding are  the best at balancing  the privacy/utility trade-offs.  This shows that our auditing framework highlights the effect of  hyper-parameters choices such as decoding strategies and their impacts on all trust dimensions

 \begin{table*}[htp!]
\centering
\begin{tabular}{l l *{5}{S[table-format=1.2]} S[table-format=1.2]}
\toprule
\multirow{4}{*}{\rotatebox{0}{\parbox{2cm}{\centering Training Data for Fraud Detector}}} & & \multicolumn{6}{c}{Training Regime for TabRoBERTa Feature Extractor } \\
\cmidrule{3-8}
 & & \multicolumn{5}{c}{Private} & \multicolumn{1}{c}{Non-Private} \\
\cmidrule(lr){3-7} \cmidrule(lr){8-8}
 && \multicolumn{1}{c}{$\varepsilon$=1} & \multicolumn{1}{c}{$\varepsilon$=3} & \multicolumn{1}{c}{$\varepsilon$=10} & \multicolumn{1}{c}{$\varepsilon$=30} & \multicolumn{1}{c}{$\varepsilon$=1000} \\
\midrule
\multirow{1}{*}{\rotatebox{0}{Real}} & & 0.72 & 0.77 & 0.73 & 0.83 & 0.80 & 0.88 \\
\cmidrule{2-8}
\multirow{8}{*}{Synthetic (TabGPT)} &Private ($\varepsilon$=0.1) & 0.48 & 0.49 & 0.47 & 0.45 & 0.48 & 0.50 \\
 &Private ($\varepsilon$=1) & 0.48 & 0.47 & 0.47 & 0.47 & 0.48 & 0.48 \\
 &Private ($\varepsilon$=5) & 0.49 & 0.48 & 0.48 & 0.47 & 0.48 & 0.48 \\
 &Private ($\varepsilon$=20)& 0.49 & 0.49 & 0.50 & 0.50 & 0.51 & 0.50 \\
 &Private ($\varepsilon$=50)& 0.51 & 0.54 & 0.60 & 0.56 & 0.59 & 0.70 \\
 &Private ($\varepsilon$=100) & 0.64 & 0.63 & 0.73 & 0.72 & 0.74 & 0.75 \\
 &Private ($\varepsilon$=200) & 0.66 & 0.66 & 0.72 & 0.71 & 0.78 & 0.77 \\
 &Nonprivate& 0.61 & 0.57 & 0.66 & 0.70 & 0.72 & 0.79 \\
\bottomrule
\end{tabular}
\captionof{table}{Performance (F1-macro) of the fraud classifier on the test set of credit card transactions for different training choices of classifier (rows) and TabRoBERTa features extractor (columns).}
\label{table:f1_macro_scores}
\end{table*}

\begin{table*}[ht!]
\centering
\begin{tabular}{lc||cccc||ccc}
\toprule
  &  Epoch & Fidelity & Utility  & Privacy & Fairness & $\omega_{all}$ & $\omega_{U}$ & $\omega_{UF}$  \\
 \midrule
\multirow{4}{*}{$\text{BioGPT}_{\text{Finetuned}}$} & 3   & 0.29 & 0.55 & 1.00  & 0.53 & \ranktwo{2}  & \ranktwo{2}  &  \ranktwo{2} \\
 & 5 & 0.44 & 0.38  & 0.75 & 0.60   & \rankthree{3} & 4  &  4 \\
 & 7 & 0.84 & 0.51 & 0.50 & 0.90   & \rankone{1} & \rankthree{3}  & \rankone{1} \\
 & 9 & 0.89 & 0.88 & 0.25  & 0.33& 4 & \rankone{1}  & \rankthree{3} \\
\end{tabular}
\caption{ Trust Indices of $\text{BioGPT}_{\text{Finetuned}}$ on MIMIC-III notes. Robustness dimension is not evaluated herein.  trustworthiness index Ranking of synthetic data sampled from different epoch during the finetuning of BioGPT model (Note that the robustness dimension was not considered) . The ranking corresponds to different trust tradeoffs $\omega$: for $w_{U}$ that is accuracy driven, we see that the last epoch is outperforming the other ones; When in addition  we consider the fairness of the prediction ($\omega_{UF}$) the last epoch (epoch 9) ranks third and the  epoch 7 presents better utility/ Fairness trade-offs. }
\label{tab:mimic_notes_results}
\end{table*}

\subsection*{Use Case II: Fraud Detection, Deep Dive on Utility and Privacy trade-offs in Synthetic Data }

\noindent In this use case, we investigate the use of  synthetic data in a financial application for  training fraud detectors. We focus here on the impact of the synthetic data on  the utility and privacy tradeoffs,    The goal  herein is therefore to highlight a use case of synthetic data in an end to end fashion without reporting aggregation level of metrics to have a more in depth analysis of the privacy/ utility trade offs. To conduct our study, we use the credit card transactions of\cite{altman2021synthesizing,padhi2021tabular} to train our TrustFormer models. These transactions were created using a rule-based generator, where values were generated through stochastic sampling techniques. The dataset contains 24 million transactions from 20,000 users, with each transaction (row) consisting of 12 fields (columns) that include both continuous and discrete nominal attributes.

\noindent \paragraph{Training RoBERTa-like Embedding}To train TabRoBERTa on our transaction dataset, we constructed samples as sliding windows of 10 transactions, using a stride of 5. We excluded the label column, "isFraud?", during training to prevent biasing the learned representation for the downstream fraud detection task. We masked $15\%$ of a sample's fields, replacing them with the [MASK] token, and predicted the original field token using cross-entropy loss. We used DP-SGD for transformer models \cite{li2022large} to train various RoBERTa-like models with differing degrees of privacy, ranging from highly private
($\varepsilon=1$) to non-private ($\varepsilon=1000$). Additionally, we trained a RoBERTa model without private training (see the columns labeled "Private" and "Non-Private" in Table~\ref{table:f1_macro_scores}).
\begin{table*}[ht!]
\centering
\begin{tabular}{@{}ccccccccc@{}}
\toprule
      & \multicolumn{4}{c}{ImageNet v1}                                            & \multicolumn{4}{c}{ImageNet v2}                       \\ \midrule
      & sig 0.0     & sig 0.1     & sig 0.3     & \multicolumn{1}{c|}{sig 0.5}     & sig 0.0     & sig 0.1     & sig 0.3     & sig 0.5     \\ \midrule
real  & \textbf{74.9} / \textbf{92.3} & \textbf{72.6} / \textbf{91.2} & 62.7 / 84.7 & \multicolumn{1}{c|}{48.8 / 73.3} & \textbf{63.1} / \textbf{84.5} & \textbf{60.4} / \textbf{82.7} & 48.6 / 72.5 & 34.8 / 58.6 \\
synthetic & 54.8 / 76.2 & 50.3 / 71.7 & 39.7 / 60.6 & \multicolumn{1}{c|}{27.9 / 46.7} & 45.6 / 68.9 & 40.7 / 62.7 & 29.9 / 50.1 & 19.0 / 35.8 \\ 
real + 0.5syn & 74.5 / 92.0 & 72.5 / 90.9 & \textbf{63.1} / \textbf{85.0} & \multicolumn{1}{c|}{\textbf{49.8} / \textbf{74.1}} & 62.3 / 83.9 & 59.5 / 82.2 & \textbf{48.8} / \textbf{73.0} & \textbf{35.4} / \textbf{59.5} \\
real + 1.0syn & \textbf{74.9} / 91.9 & 72.3 / 90.9 & 62.4 / 84.5 & \multicolumn{1}{c|}{48.8 / 73.0} & 62.5 / 83.6 & 59.8 / 82.2 & 49.1 / 72.2 & 34.1 / 57.5 \\ \bottomrule
\end{tabular}
\caption{Performance of ResNet50 on ImageNet v1 and v2 Datasets under varied noisy conditions (additive random Gaussian noise to tested images) and different training regimes (synthetic data augmentation). The resulting models are evaluated in Acc@1 (first number) and Acc@5 (second number). We see that while classifiers trained purely on synthetic data lag behind those trained on real, augmenting real data with synthetic images makes models more resilient to noise. }
\label{tab:imagenet_results}
\end{table*}

\noindent \paragraph{Synthetic data generation} We generated several privacy-preserving synthetic datasets using our non-private pretrained TabGPT model. For model selection in this experiment we relied on a fidelity validation of the TabGPT model. To generate private synthetic data, we used a private sampling technique \cite{NEURIPS2021_f2b5e92f} , which involves adding Laplacian noise with controlled variance (dependent on the user-provided 
$\varepsilon$ value) to the probability distribution over the generated tokens from the non-private GPT model. This is a form of output-perturbation methods that guarantees differential privacy.We generated seven datasets with varying privacy levels, from highly private ($\varepsilon=0.1$) to non-private ($\varepsilon=200$), as shown in the rows labeled "Synthetic" in Table \ref{table:f1_macro_scores}. Additionally, we considered real card transaction data and synthetically generated data without private sampling.

\noindent \paragraph{Training the downstream Fraud Detection Model} Given the various transaction datasets, we constructed a simple multi-layer perceptron (MLP) classifier that was trained directly on the embeddings of the various RoBERTa feature extractors that we trained . Note that thanks to the additivity property of differential privacy the overall privacy of the fraud detector is the addition of the privacy budget of synthetic data and the privacy budget of the feature extractor. The RoBERTa feature extractor remained fixed during the fraud detector training. For each training scenario, we selected 800K transactions for training, 100K transactions for validation, and 100K transactions for testing. Note that the test transactions were always the same across different datasets and were chosen from real data. In contrast, the training and validation splits were determined according to the training regimes.

\noindent \paragraph{Results and Discussion}  The highest utility performance is achieved when using a RoBERTa feature extractor trained without differential privacy, and when training the fraud detector on real transaction data (first row in the table). Conversely, utilizing a highly private RoBERTa model in conjunction with highly private synthetically generated data yields  (unsurprisingly) significantly poorer F1-macro performance (upper left corner of the table). Furthermore, it can be observed that for a fixed row (dataset for training the fraud detector), moving from left to right across columns (corresponding to decreasing privacy levels of the RoBERTa feature extractor) results in improved utility performance for the fraud detector. Similarly, for a fixed column (pretrained RoBERTa feature extractor), moving down the rows (excluding the first row, which depicts performance on real data, and excluding the last row which depicts performance on non-private synthetic data) leads to better classifier performance. It is interesting to see when comparing to the last row, that private synthetic data with private embeddings introduces a regularization effect leading to better performance than the same setup with non-private synthetic data.

\section*{Natural Language Synthetic Data : Deep Dive on Utility and Fairness Trade-offs}

In this section, we delve into controlling trust trade-offs within the context of language modeling, using the BioGPT model \cite{bioGPT} as our testbed. We fine-tune a BioGPT-Large model on the MIMIC-III notes dataset \cite{vanAken2021}, comprising 423,015 patient notes, with an accompanying label denoting patient survival (expiration flag). MIMIC-III is known for its bias issues \cite{mimic-fairness,mimic-IF,chen2018my}. \\

 We augment the MIMIC III notes dataset with patient age, gender, and ethnicity, and  fine-tune the model on the resulting data. We then fine-tune the BioGPT\-Large model (non-private training) on this augmented notes data which is first prompted by the target label and then by the controls (ethnicity, age, gender). At inference time, from a fine-tuned BioGPT\-Large model at a given epoch, we generate a balanced synthetic dataset of the same size as the real training data via prompting the model with the same amount of positive and negative labels (expiration flag). In this setting, we use a multinomial decoding strategy for generation.  At the end we obtain a labeled synthetic dataset of synthetic doctor notes along with the controls on ethnicity, age and gender for each sample.\\

\noindent We audit the synthetic dataset sampled from different epochs (namely after 3, 5, 7 and 9) during the fine-tuning process. The downstream task we consider in this audit is  the in hospital mortality prediction task, where the protected community is the ethnicity "ASIAN" \cite{chen2018my}. As an embedding \textbf{E}, we use the original pre-trained BioGPT model \cite{bioGPT} to extract embeddings for the synthetic notes as it has the capability of representing the biomedical domain.
In Table \ref{tab:mimic_notes_results} we see that our trustworthiness index driven model selection allows a controllable trade-off between utility and fairness : epoch 7 has a better utility fairness trade-off than the model at the last epoch that has higher utility. Therefore it is favorable to select the model at epoch 7 at the price of a reduced utility but with an enhanced fairness, which is of paramount importance for this use-case.

\section*{ Synthetic Image  Data: Deep Dive on Utility and Robustness to Noise and Distribution Shift }

\noindent We consider here the use of  synthetic image data for training classifiers on one of the key computer vision datasets,  Imagenet \cite{DenDon09Imagenet}, with a focus on the utility and generalization propreties of these classifiers on the Imagenet test set and under  distribution shifts. Imagenet consists of $1.2$ Million images of size 256x256x3 and 1000 categories. Authors in \cite{recht2019imagenet} constructed Imagenetv2 test set, that consists of a distribution shift from the original imagenet distribution and reported a performance drop between $11\% - 14\%$.  Recent works \cite{azizi2023synthetic} and \cite{burg2023data} showed promising results of synthetic data from diffusion models (a family of generative models that uses diffusion techniques \cite{sohl2015deep}) in improving Imagenet classification.  Following these promising works, we synthesize $1.2$ M labeled images  from the Imagenet 256x256 pretrained guided diffusion models  from OpenAI \cite{dhariwal2021diffusion}.
Table \ref{tab:imagenet_results} presents ResNet50 \cite{he2016deep} performance on ImageNet-v1 and v2 under Gaussian noise, considering four training data scenarios: (1) real Imagenet; (2) synthetic data; (3) a hybrid with real and synthetic images (real to synthetic ratio $1/0.5$); and (4) an equal mix totaling 2.4 million images. Results show synthetic-only models lag, but integrating with real images enhances robustness, especially in noisier settings (sigma 0.3 and 0.5). Combining synthetic data with real images improves a model's resilience to noise and image corruption.
~~\\

\section*{Conclusion}\label{sec:Conclusion}

We introduced a holistic framework for auditing synthetic data along trust pillars. Towards this end, we defined a trustworthiness index that assesses the trade-offs between trust dimensions such as fidelity, privacy, utility, fairness, and robustness and quantifies their uncertainty. Moreover, we devised a trustworthiness index driven model selection and cross-validation via auditing in the training loop, that allows controllable trust trade-offs in the resulting synthetic data. Finally, we instrumented our auditing framework with  workflows connecting various stakeholders from model development to certification, and we defined  templates to communicate transparency about model audits via a Synthetic Data auditing report. \\

\noindent Our framework highlights the potential of synthetic data in various modalities, including tabular, time series, natural language, and vision. However, for critical applications where the trustworthiness of synthetic data is paramount for its integration into the AI lifecycle of sensitive downstream tasks, it's important to recognize that not all generative AI techniques and training approaches are equally reliable. Thus, conducting rigorous audits of synthetic data is imperative to guide the training process and obtain certifications for internal use, third-party entities, and regulatory compliance.

\begin{figure*}[ht!]
    \centering
    \includegraphics[scale=0.55]{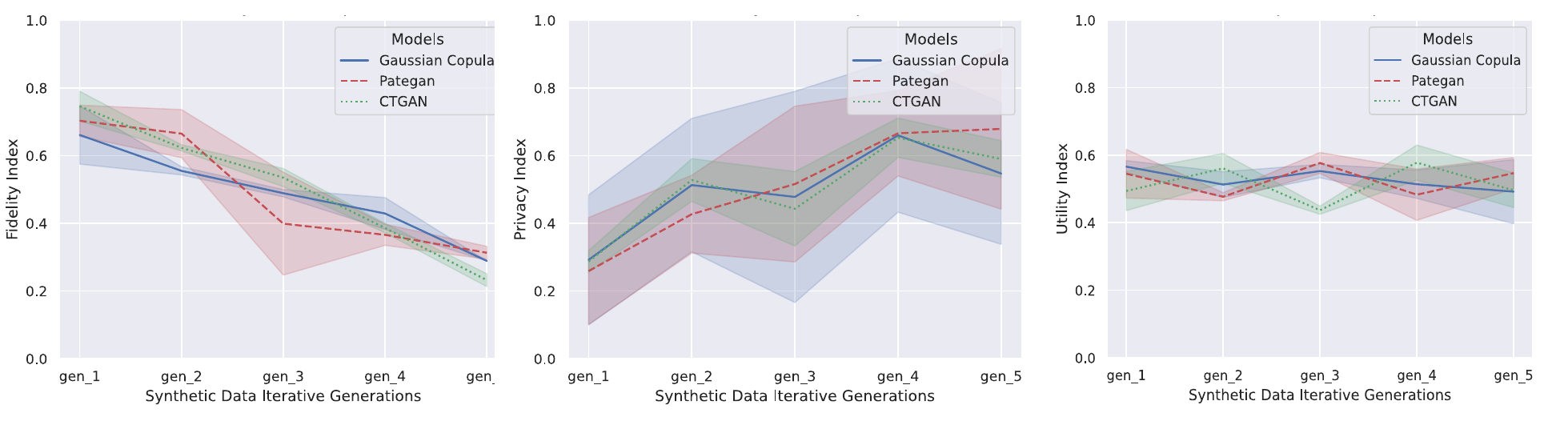}
    \caption{Iterative Use of Synthetic Data to Train Generative Models on the Bank Marketing Dataset.  The plot shows mean and variance of indices across folds for all generations. As we iteratively use synthetic data to train generative models, we see that while the fidelity index of the synthetic datasets collapses, the privacy index improves. The utility of the synthetic data on downstream tasks fluctuates but is essentially constant. Note that the utility of these models does not deteriorate, since the index of these synthetic data at first iteration is not high.  }
    \label{fig:collapse}
\end{figure*}

\section*{Ablation Studies}

\paragraph{Debiasing downstream tasks trained on Synthetic Data} We notice on tabular datasets that while private synthetic data with high performing robustness index are still competitive in terms of their utility, the ones with high fairness index are not competitive in terms of their utility. In order to further improve the fairness, privacy and utility tradeoffs, we explore including debiased downstream classifiers in Supplementary Information Tables \ref{table:bank_marketing_metrics_biased_debiased}, \ref{table:recruitment_metric_biased_debiased} and \ref{table:law_school_metrics_baised_debiased}.  Debiasing leads to an improvement fairness while maintaining utility of private synthetic data.

\paragraph{Independent Copula Versus Dependent Copula} In our aggregation we used an independent  copula as a mean to aggregate the metrics. We explore here using a Gaussian copula to capture the dependencies between metrics and see how it impacts the overall ranking of the synthetic datasets on all tabular datasets we consider. We explore  Gaussian copulas, since estimating dependent copulas from data is cursed in high dimension.  To estimate the Gaussian copula, we compute the mean and covariance matrix for the concatenation of all metrics under a trust dimension, and compute the multivariate Gaussian CDF. This comes with two caveats: 1) the dependency may not be Gaussian 2) given small number of samples the covariance matrix  is not well estimated and is  low rank and inverting it in the computation of the multivariate CDF leads to numerical instabilities that we circumvent with a regularization of the covariance matrix $\Sigma$ by $\Sigma + \lambda I$, for $\lambda=10^{-3}$. We first report the average time in seconds to compute the independent copula on tabular use cases on an Intel (R) Xeon (R) CPU E5-2667 v2 @ 3.30GHz  that is of  $11s$ versus $1700s$  for the Gaussian copula (See  Supplementary Information \ref{app:GCversusIC} Table
\ref{fig:timingIC_GC}). Second we report the overlap between the ranked list  of synthetic data returned by the  independent copula $L_{IC}$ and the one returned by the Gaussian copula $L_{GC}$:
\[\text{overlap@k} = \frac{| L_{IC}[1,k]\cap L_{GC}[1,k] |  }{| L_{IC}[1,k]\cup L_{GC}[1,k] |. }\]
From  Supplementary Information \ref{app:GCversusIC} Figure \ref{fig:overlap_IC_GC} we see that there is 60-70\%  overlap between the top-10 ranking synthetic datasets returned by independent and Gaussian copula. In light of these results, given the simplicity and the speed of the independent copula and the numerous issues of the Gaussian copula in terms of its slow computation and bad conditioning and unfavorable statistical guarantees, the independent copula  is preferable and offers a good computational tradeoff in estimating the copula. 

\begin{table*}[ht!]
\centering 
\begin{tabular}{|c|c|c|c|}
\hline
   Synthetic Data Model &  Gaussian Copula     &  CTGAN    & PateGAN \\    
\hline
 $1-\textrm{AUC}$ of Membership Attack \cite{van2023membership} & 0.4966 & 0.4882 & 0.4879 \\
\hline
Privacy Index & 0.2916 & 0.2866 & 0.2588 \\
\hline
\end{tabular}
\renewcommand{\tablename}{Table}

\caption{Membership Inference Attacks of \cite{van2023membership} versus our Privacy Index on the Bank Marketing dataset. In the second row we give $1-\textrm{AUC}$ of the membership attack classifier (where AUC is the Area Under the Curve). Higher values of $1-\textrm{AUC}$ means that the membership attack at identifying members of the training data is less successful, meaning higher privacy. We see that this membership attack lead to the same ranking of synthetic data  as our privacy index given in  the third row.  }
\label{tab:attack}
\end{table*}
 
\paragraph{Membership attacks} The membership attacks we used in our privacy auditing were based on nearest neighbor attacks that  assess memorization issues in synthetic data. Identity disclosure attacks assess privacy in synthetic data beyond memorization \cite{Taub2018DifferentialCA},\cite{el2020evaluating}. We explore here the impact of other targeted  membership attacks based on likelihood ratios between synthetic data and a reference to estimate the membership \cite{van2023membership}. This attack  is the state of the art membership inference method. We train this attack on the Bank Marketing dataset, evaluating synthetic data of three baselines: the Gaussian copula, CTGAN and PateGAN.  We give in 
in ] Table \ref{tab:attack},  $1-\textrm{AUC}$ of this attack, where the Area Under the Curve (AUC) is evaluated on the real data with members and non members of the training set.  Higher values of $1-\textrm{AUC}$ mean that the membership attack is less successful at identifying members of the real data, meaning higher privacy. We see in  Table \ref{tab:attack} that this membership attack lead to the same ranking of synthetic datasets  as our privacy index (computed without the inclusion of the membership attack). Those membership inference attacks can be seamlessly added to our privacy index computation.

\paragraph{Iterative Collapse Risk of Synthetic Data} As synthetic data becomes more and more prevalent, the risks of iteratively training generative models on synthetic data becomes more and more pressing. Authors in \cite{shumailov2023curse} referred to this issue as the iterative curse or iterative collapse in synthetic data.  They also showed that the fidelity of the synthetic data  deteriorates as we  iteratively train generative models on synthetic data. We show in  Figure \ref{fig:collapse} on the Bank Makerting dataset, the audits of synthetic data obtained from training models  successively on synthetic data generated from the previous iteration. As we iteratively use synthetic data to train generative models, we see in  Figure \ref{fig:collapse} that while the fidelity index of the synthetic datasets collapses, the privacy index improves as expected. The utility of the synthetic data on downstream tasks fluctuates but is essentially constant. Note that the utility of these models does not deteriorate, since the  utility of these synthetic data at first iteration is not high. 


\newpage
\section*{Data availability}
Data used in  the paper  is available online and open source. See Table \ref{Tab:usecases} for references. 
\section*{Code availability}
A code reproducing tables in the paper for the recruitment dataset is  available on \url{https://ibm.biz/synthetic-audit}.   Examples of full auditing reports on all use cases are provided in the Supplementary information. 

\section*{Acknowledgments}

We thank IBM Research for supporting this work. Y.M would like to thank Payel Das, Kush R. Varshney and Abdel Hamou-Lhadj for insightful discussions. 

\section*{Authors Contributions}
Y.M conceived the project and wrote the initial draft of the paper.  All authors contributed to developing the synthetic data auditing framework and designed experiments and contributed to their analysis.  All authors contributed to the writing of the paper. 

\section*{Ethics declarations}
 The authors declare no competing interests.

\newpage 


\begin{IEEEbiographynophoto}{Brian Belgodere}
 is a senior research software engineer in the Emerging Tech. Engineering Department at IBM Research. He currently works on hybrid cloud platforms and tooling for Machine Learning and AI. He holds B.S. degrees in economics and business administration from Carnegie Mellon University and a Juris Doctor from the University of Pittsburgh. \end{IEEEbiographynophoto}
\begin{IEEEbiographynophoto}{Pierre Dognin}
received his M.S. degree and Ph.D degree in Electrical Engineering from the University of Pittsburgh in 1999 and 2003.
He then joined IBM Research in the Human Language Technologies Department where he worked as a Research Staff Member on Automatic Speech Recognition, focusing
on acoustic modeling, robust speech recognition, and core algorithm technology.
Since 2015, he has been working on multimodal Machine Learning and is now in the Trusted AI Department where his research interests
include Deep Learning, Generative Modeling for Computer Vision and Natural Language.
\end{IEEEbiographynophoto}
\begin{IEEEbiographynophoto}{Igor Melnyk}
is currently a Research Staff Member in the Trusted AI Department at IBM Research. Before that he completed his PhD degree in Computer Science and Engineering in 2016 from the University of Minnesota. In 2010, he obtained his MS degree in Computer Science from the University of Colorado, Boulder. His research interests are in the areas of Machine Learning and AI with the focus on the problems of Natural Language Processing and Computer Vision.
\end{IEEEbiographynophoto}
\begin{IEEEbiographynophoto}{Youssef Mroueh}
is a research staff member in the Trusted AI department at IBM Research and a principal investigator in the MIT-IBM Watson AI lab. He received his PhD in computer science in February 2015 from MIT, CSAIL, where he was advised by Professor Tomaso Poggio. In 2011, he obtained his engineering diploma from Ecole Polytechnique Paris France, and a master of science in Applied Maths from Ecole des Mines de Paris. He is interested in Multimodal Deep Learning, Generative modeling, Computer Vision and Learning Theory.
\end{IEEEbiographynophoto}

\begin{IEEEbiographynophoto}{Aleksandra (Saška) Mojsilović}
 is a Serbian-American scientist. Her research interests are artificial intelligence, data science, and signal processing. She is known for innovative applications of machine learning to diverse societal and business problems. Her current research focuses on issues of fairness, accountability, transparency, and ethics in AI. She is an IBM Fellow and IEEE Fellow.
\end{IEEEbiographynophoto}

\begin{IEEEbiographynophoto}{Jiri Navratil}
received M.Sc. (’95) in EE \& PhD (’99) in CS from Technische Universitaet Ilmenau, Germany. He joined IBM Research in 1999, where he is currently a Principal Research Staff Member.
His research interests include Deep Learning, Natural Language Processing and Industrial AI Applications.
\end{IEEEbiographynophoto}

\begin{IEEEbiographynophoto}{Apoorva Nitsure}
is a Research Software Engineer at IBM Research. Her interests include Deep Learning, Natural Language Processing and Interpretability in Machine Learning to help users make informed choices while using AI. Prior to joining IBM Research, she completed her Masters degree from Heinz College, Carnegie Mellon University and B.Tech in IT from CoEP India. 
\end{IEEEbiographynophoto}

\begin{IEEEbiographynophoto}{Inkit Padhi}
 received Masters's degree in Computer Science from Viterbi School of Engineering, USC/ISI in May 2016. Since December of 2018, he is working as a research engineer in the Trusted AI department at IBM Research. He is interested in Machine Learning and Natural Language Processing. His pronouns are he, him, and his.
\end{IEEEbiographynophoto}
\begin{IEEEbiographynophoto}{Mattia Rigotti}
is a Research Staff Member in the AI Automation Department at IBM Research, which he joined in 2014. He holds an M.Sc.\ degree in Theoretical Physics from ETH Zurich and a Ph.D.\ in Computational Neuroscience from Columbia University.
His research interests include Deep Learning, Neuromorphic Engineering and Computational Neuroscience.
\end{IEEEbiographynophoto}
\begin{IEEEbiographynophoto}{Jarret Ross}
received an M.S. in Computer Science from Wichita State University. He has been with IBM research since 2016 and he is currently a research engineer in the Trusted AI department. His interests are in deep learning, distributed computing and generative modeling.  
\end{IEEEbiographynophoto}
\begin{IEEEbiographynophoto}{Yair Schiff}
 is a PhD student in the Computer Science department at Cornell University. Yair was a software developer working on IBM Watson Studio. Prior to joining IBM, he completed an M.S. degree in Computer Science at the Courant Institute of Mathematical Sciences at New York University. Yair collaborated with the Trusted AI department at IBM Research. .\end{IEEEbiographynophoto}
\begin{IEEEbiographynophoto}{Richard A. Young} received his B.S. Hons degree in computer science from The University of the Witwatersrand (South Africa) in 2010. He worked as a software developer in the banking and health insurance industries from 2011 to 2018. He joined IBM Research in South Africa as a Research Engineer in 2018 where he specialises in building web and cloud based applications.
\end{IEEEbiographynophoto}
\begin{IEEEbiographynophoto}{Radhika Vedpathak} is a designer with an experience in developing design for web applications. 
\end{IEEEbiographynophoto}

\begin{IEEEbiographynophoto}{Richard A. Young} received his B.S. Hons degree in computer science from The University of the Witwatersrand (South Africa) in 2010. He worked as a software developer in the banking and health insurance industries from 2011 to 2018. He joined IBM Research in South Africa as a Research Engineer in 2018 where he specialises in building web and cloud based applications.
\end{IEEEbiographynophoto}

\bibliographystyle{plain}
\bibliography{confs}





\begin{center}
\onecolumn
\appendix

\large{\textbf{Supplementary Material}}
\end{center}

\subsection{Auditing Framework } \label{app:descFramework}

\paragraph{Quantitative Assessment Metrics and Monotonicity Alignment (\textbf{\textsc{Evaluate}})} We select a set of metrics $\mathcal{M}_{T}$ that assess quantitatively the risks of each trust dimension $T\in \{\text{Fidelity, Privacy, Utility, Fairness, Robustness}\}$. A metric $m\in \mathcal{M}_{T}$is an interpretable measurement performed on synthetic and real datasets $D_s$ and $D_r$ and an associated downstream task configuration $\text{\ttfamily{cfg}}$: 
$$m: (D_r,D_s,\text{\ttfamily{cfg}}) \to \mathbb{R},$$
where the downstream task and auditing configurations \text{\ttfamily{cfg}} are defined as follows: $\text{\ttfamily{cfg}}=\{$\text{classification task}, \text{sensitive communities}, \text{protected attributes}, \text{privacy budget}, \text{radius of robustness}$\}.$
 
\noindent The numeric values of these metrics indicate risks for each trust dimension. Metrics may have varying polarities,  where lower values can suggest high risk for some and low risk for others. To address this, we normalize metric polarity to be in the following direction: higher metric values correspond to lower risks.  We achieve this normalization by introducing for each metric $m$ a polarity $p \in \{\pm 1\}$. If the metric monotonicity conforms to our convention it has a positive polarity ($+1$), otherwise it has a negative one ($-1$). For example for Fidelity, a metric measuring the distributional distance between synthetic data and real data has a negative polarity, as low distances mean high fidelity. Another example is the accuracy metric of a predictive model for Utility that has a positive polarity, as high accuracy values correspond to high utility. Hence for a trust dimension T, we preprocess the metrics $m_{T,i}, i=1\dots M_{T}$by multiplying them by their corresponding polarities $p_{T,i}$ so that they have aligned monotonicity, which yields the following set of aligned metrics:
$$\mathcal{A}(\mathcal{M}_{T})=\{ \tilde{m}_{T,i}=p_{T,i} \quad m_{T,i}, i=1\dots M_{T} \}, \text{ for } T \in \{\text{Fidelity, Privacy, Utility, Fairness, Robustness}\}.$$

\noindent In  Table \ref{tab:metricMM}, we provide the set of metrics corresponding to each trust dimension and their associated polarities. Note that all Fidelity and Privacy metrics are evaluated between the synthetic data $D_{s}$ and the real training set $D_{r,\text{train}}$. Utility, Fairness, and Robustness metrics refer to evaluations of classifiers trained on the synthetic data $D_{s}$, validated on the real development set $D_{r,\text{val}}$, with final measurements reported on the real test set $D_{r,\text{test}}$.
A glossary of all metrics and their definitions is provided in Supplementary Information \ref{sec:metricdetails}. 

\paragraph{Aggregation of Metrics within a Trust Dimension via the Copula Method (\textbf{\textsc{Aggregate}})}  For a trust dimension $T$, we would like to aggregate the aligned metrics which are evaluated on a synthetic dataset $D_s$ against a real dataset $D_r$, i.e, $\tilde{m}_{T,i}(D_r,D_s,\text{\ttfamily{cfg}}) \text{ for } i=1\dots M_{T}$. Such an aggregation would define an index for the trust dimension $T$.\\

\begin{figure}[htp]
\centering
\includegraphics[width=0.6\linewidth]{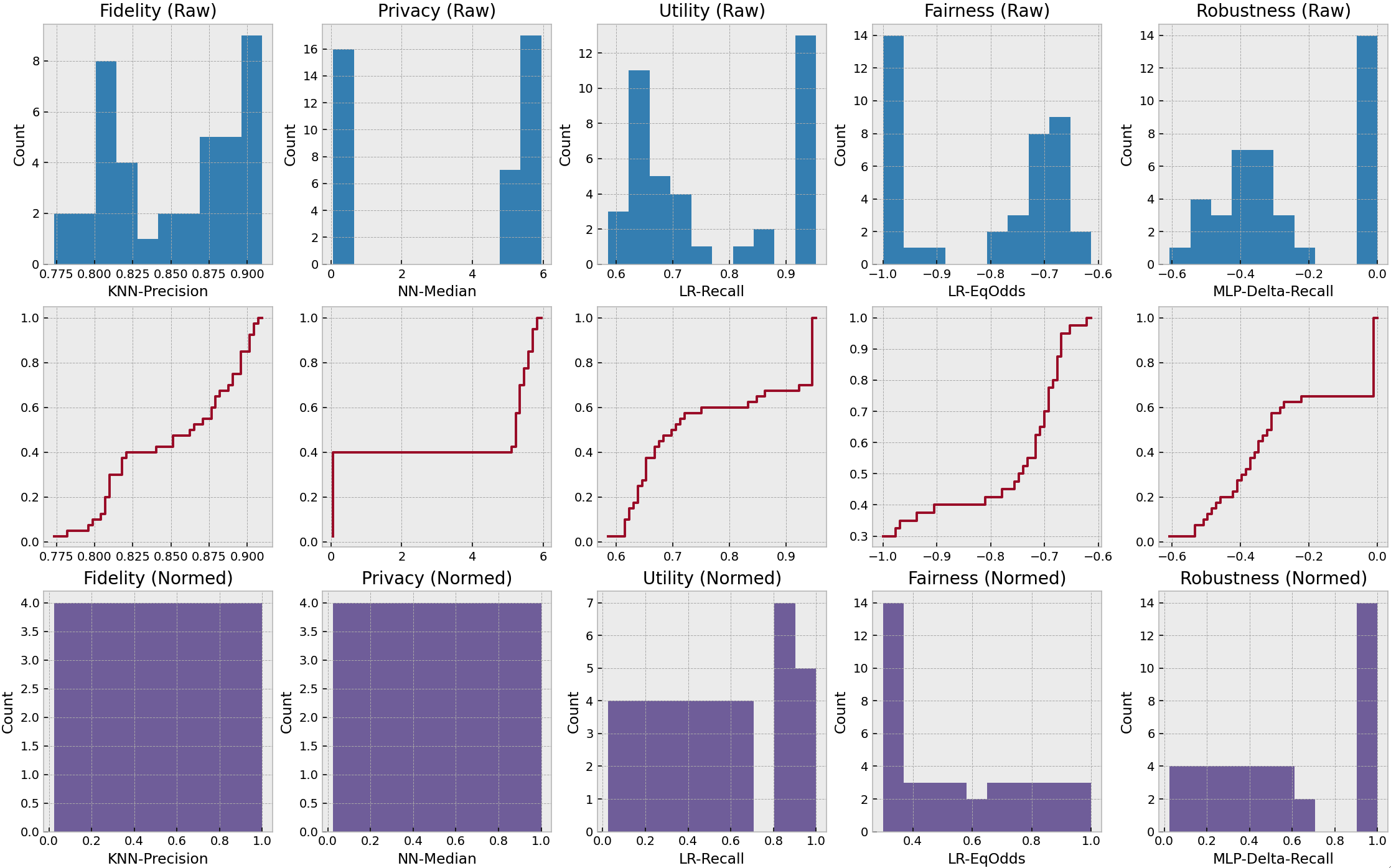} 
\caption{ Metric normalization via the empirical CDF transform.}
\label{Fig:metrics_histograms_ls}
\end{figure}

\noindent The main challenge here is that these metrics may have different dynamic ranges which makes a simple mean aggregation incorrect, as it favors metrics that have large dynamic range. To circumvent this challenge we resort to a popular method in software quality assessment, the so called copula aggregation method \cite{copula-aggregation} that proceeds as follows:
\begin{enumerate}
\item \textbf{Empirical Cumulative Distribution Function (ECDF) evaluation:} For each aligned metric $\tilde{m}_{i,T},$ we compute the Empirical CDF, given the observations across multiple synthetic datasets $D^j_{s},~j=1,\ldots,N$: 
 $$\text{ECDF}_{\tilde{m}_{i,T}} (x)= \frac{1}{N} \sum_{j=1}^N \mathbb{1}_{\tilde{m}_{T,i}(D_r,D^j_s,\text{\ttfamily{cfg}}) \leq x}, \text{ for } x\in \mathbb{R}.$$
\item \textbf{Score Normalization via ECDF mapping} We use empirical CDFs to map the distribution of each aligned metric $\tilde{m}_{T,i}$ to an (approximately) uniform distribution via $ECDF(\tilde{m}_{T,i})$. For a particular synthetic data $D_s,$ we transform all aligned metrics using their corresponding empirical CDF and obtain a normalized score:
\begin{equation}
u_{T,i} =\text{ECDF}_{\tilde{m}_{i,T}} (\tilde{m}_{T,i}(D_r, D_s,\text{\ttfamily{cfg}})),~i=1,\dots, M_{T}. 
 \label{eq:CDFTransf}
 \end{equation}
 $u_{T,i}$ is now very informative about the quality of synthetic data $D^j_s$with respect to the metric $\tilde{m}_{i,T}$, while being in $[0,1]$ range for all metrics. To see that, recall:
$u_{T,i} \approx \mathbb{P}(\tilde{m}_{i,T} \leq \tilde{m}_{T,i}(D_r, D^j_s,\text{\ttfamily{cfg}})).$
Hence for an aligned metric $\tilde{m}_{i,T}$, high scoring synthetic datasets will have a normalized score $u_{T,i}$ close to 1 and low scoring ones will have a normalized score close to 0. 
Figure \ref{Fig:metrics_histograms_ls} 
shows an example how the Empirical CDF transformation leads to an almost uniform distribution, which allows aggregation via the so called copula method using a simple geometric mean of normalized scores of metrics given in Equation \eqref{eq:CDFTransf}.

\item \textbf{Compute the Copula / Trust Dimension Index:} The conformity of synthetic data to a trust dimension can be formalized via the copula method as an aggregation of normalized scores of all metrics given in Equation \eqref{eq:CDFTransf} as follows:
$\pi_{T}(D_r, D_s,\text{\ttfamily{cfg}}) =C (u_{T,1}, \dots u_{T,M_{T}}),$
where $C$ is the copula corresponding to the density function of all metrics, considered jointly. In general, it is difficult to estimate such a copula, and we resort to simple geometric mean copula. The index of a trust dimension $T$ of a synthetic dataset $D_s$ can be therefore defined using the geometric mean copula aggregation:
\begin{equation}
\centering
\pi_{T}(D_r, D_s,\text{\ttfamily{cfg}})=\exp \left(\frac{1}{M_T} \sum_{i=1}^{M_T} \log(u_{i,T}) \right).
\label{eq:copula}
\end{equation}
If there is a known priority of certain metrics relative to other ones, this can be integrated via weights $\beta_{i,T}\in [0,1]$ and $\sum_{i=1}^T \beta_{i,T} =1$, that reflect the relative importance of each metric. The copula can then be evaluated as follows:
\begin{equation}
\centering
\pi_{T}(D_r, D_s,\text{\ttfamily{cfg}})=\exp \left(\sum_{i=1}^{M_T} \beta_{i,T} \log(u_{i,T}) \right).
\label{eq:copulaweighted}
\end{equation}

\end{enumerate}

\noindent The details of the implementation of the copula aggregation method are given in Supplementary Information \ref{sec:aggregationdetails}. Although effective and widely accepted as an aggregation method, the main limitation of the geometric mean copula is the inherent assumption of independence between metrics. This shortcoming can be alleviated if a dependency structure is known and can be incorporated via hierarchical or vine copulas \cite{copula-hier}.

\paragraph{Trustworthiness index: Trade-offs and Aggregation Across Trust Dimensions (\textbf{\textsc{Reweight}} \& \textbf{\textsc{Index by Trust})}} We are now ready for the last steps in our auditing framework \textbf{\textsc{Reweight}} \& \textbf{\textsc{Index by Trust}}. Trust dimensions can often be in conflict with each other, hence, for each application and use case and depending on the policy and regulation landscape, a trade off between these dimensions can be defined by introducing a weighing scheme $\omega$ for each trust dimension. This weighting scheme reflects the order of priorities among trust dimensions to ensure compliance with safeguards and regulations. The weights
$\omega=(\omega_{T}, T \in \mathcal{T}=\{$ \text{Fidelity, Privacy, Utility, Fairness, Robustness}$\}),$
define a probability distribution on trust dimensions i.e $\omega_{T}\geq 0, \sum_{T\in \mathcal{T}}\omega_{T}=1.$ 
Armed with these trade off weights and the trust dimension indices, we define the trustworthiness index of a synthetic dataset $D_s$ as the weighted geometric mean of all trust dimension indices:
\begin{equation}
\tau_{\text{Trust}}(D_r, D_s,\text{\ttfamily{cfg}},\omega)= \exp\left(\sum_{T \in \mathcal{T}} \omega_{T} \log(\pi_{T}(D_r, D_s,\text{\ttfamily{cfg}}))\right), 
 \label{eq:indextrust}
\end{equation}
where $\pi_{T}$ are trust dimension indices defined in Equations \eqref{eq:copula} or \eqref{eq:copulaweighted}. From now on , we use the notation $\omega= (\omega_f,\omega_P,\omega_U, \omega_F,\omega_{R})$ to refer to trust dimensions trade-offs corresponding to Fidelity, Privacy, Utility, Fairness, and Robustness, respectively.
Table \ref{Table:weights} provides examples of weighting schemes corresponding to different priorities in auditing synthetic data.

\paragraph{Ranking Synthetic Data via  the Trustworthiness Tndex (\textbf{\textsc{Rank}}) } Finally, for trade-off weights $\omega$, the trustworthiness index $\tau_{\text{Trust}}(D_r, D^j_s, \omega)$ defined in Equation \eqref{eq:indextrust}, allows us to rank the synthetic datasets $D^s_j,j=1\dots N$ from the highest complying (larger trustworthiness index) with the safeguards priorities reflected by $\omega$ to the least complying (smaller trustworthiness index). 

\paragraph{Uncertainty Quantification of the Trustworthiness Index \& Ranking under Uncertainty (\textbf{\textsc{Rank}})} 
Note that the trust dimension indices and the trustworthiness index are functions of the splits of real data between train, validation, and test sets. Hence, when we audit a generative modeling technique (for example GAN, private GAN, etc.), it is important to see how the synthetic data and the resulting auditing vary as we changes these splits. Towards this end, given $S$ splits of the the real data $ \mathcal{D}_r=\{D^1_r\dots D^S_r\}$,we train the same generative method on the training portion of each split and obtain $S$ different synthetic datasets $\mathcal{D}_s=\{D^1_s\dots D^S_s\}$. For all $T\in \mathcal{T}$, we report geometric mean and deviation around it of the trust dimension indices, evaluated as follows:
\begin{align}
~&\overline{\pi}_{T}(\mathcal{D}_r,\mathcal{D}_s,\text{\ttfamily{cfg}}) =\exp\left(\frac{1}{S}\sum_{\ell=1}^S \log (\pi_{T}(D^{\ell}_r, D^{\ell}_s,\text{\ttfamily{cfg}})) \right), \\
&\Delta_{T}(\mathcal{D}_r,\mathcal{D}_s,\text{\ttfamily{cfg}}) = \frac{1}{S} \sum_{\ell=1}^S \left(\pi_{T}(D^{\ell}_r, D^{\ell}_s,\text{\ttfamily{cfg}})-\overline{\pi}_{T}(\mathcal{D}_r,\mathcal{D}_s,\text{\ttfamily{cfg}})\right)^2.
\label{eq:geomTrustDimDev}
\end{align}
Note that this not the standard notion of variance, since we measure the expectation of squared $\ell_2$ distance around the geometric mean and not the arithmetic one. 

\noindent Similarly, we report geometric mean and deviation around the trustworthiness index corresponding to tradeoff weights $\omega$ across splits as follows:
\begin{align}
&\overline{\tau_{\text{Trust}}}(\mathcal{D}_r, \mathcal{D}_s,\text{\ttfamily{cfg}}, \omega)=\exp\left(\frac{1}{S}\sum_{\ell=1}^S \log (\tau_{\text{Trust}}(D^{\ell}_r, D^{\ell}_s,\text{\ttfamily{cfg}}, \omega)) \right)\nonumber\\
& \Delta_{\tau} (\mathcal{D}_r, \mathcal{D}_s,\text{\ttfamily{cfg}},\omega) = \frac{1}{S} \sum_{\ell=1}^S \left( \tau_{\text{Trust}}(D^{\ell}_r, D^{\ell}_s,\text{\ttfamily{cfg}},\omega) - \overline{\tau_{\text{Trust}}}(\mathcal{D}_r, \mathcal{D}_s,\text{\ttfamily{cfg}},\omega) \right)^2.
\label{eq:mean_var_index}
\end{align}

\noindent Now, to have a robust ranking between $N$ generation techniques for synthesizing data, given the real data splits $\mathcal{D}_r$, we train a generative model using each data set and obtain $S$ synthetic datasets which we collect as $\mathcal{D}^j_s$. For a trade off weighting $\omega$, we can rank these generation techniques using the mean trustworthiness index $\overline{\tau_{\text{Trust}}}(\mathcal{D}_r, \mathcal{D}^j_s,\text{\ttfamily{cfg}},\omega), j=1\dots N$ by sorting these values from the least complying to the most complying with the safeguards. Given $\alpha>0$, we also explore ranking under uncertainty by using:
\begin{equation}
~R^{\alpha}_{\text{trust}}(\mathcal{D}_r, \mathcal{D}^j_s,\text{\ttfamily{cfg}},\omega)=\log (\overline{\tau_{\text{Trust}}}(\mathcal{D}_r, \mathcal{D}^j_s,\text{\ttfamily{cfg}} ,\omega)) - \alpha\log(\Delta_{\tau} (\mathcal{D}_r, \mathcal{D}^j_s,\text{\ttfamily{cfg}}, \omega)),~~ j=1\dots N.
\label{eq:mean_var}
\end{equation}

Note than when ranking with $R^{\alpha}_{\text{trust}}$, we favor synthetic data with a high mean trustworthiness index and low ``variance'' across splits.

\paragraph{Auditing Report (\textbf{\textsc{Report}})} Our auditing framework sets templates for communicating the audit results for all synthetic data in the form of an auditing report. The auditing report starts by giving transparency about the real data used to train the generative models, and the training and inference configurations used for training and sampling synthetic data from each generative model. Then, for a given trade off weighting of trust dimensions $\omega$, a ranked list of models is produced using the trustworthiness index $\tau_{trust}$ defined in Equation~\eqref{eq:indextrust} or using ranking under uncertainty of trust indices given in Equations~\eqref{eq:mean_var_index} or \eqref{eq:mean_var}. Synthetic data cards in the ranked list contain the overall trustworthiness index $\tau_{\text{trust}}$, as well as the indices for each trust dimension along their variances. In addition, each card is paired with an interpretable message and warnings about potential harms and trustworthiness violations detected in the synthetic data for each trust dimension. Finally, a breakdown of all metrics evaluated for each dimension is presented, comparing all synthetic datasets to provide further insights on the aggregated trustworthiness index. An example of the audit report is given in Supplementary Information \ref{sec:AuditReport}.

\subsection*{Controllable Trust Trade-offs of the Synthetic Data via Auditing in the Training Loop }

In this section, we show how to infuse trust aspects into the training of generative models in order to meet socio-technical requirements of trust and performance. An important aspect that is often overlooked is the model and hyper-parameters selection of generative models and how this impacts the resulting downstream task of synthetic data generation. We propose a new paradigm for model selection that is not only accuracy driven, as in classical approaches \cite{hastie01statisticallearning}, but also anchored in all other trust dimensions. We achieve this by using our auditing framework as a means of cross validation of generative models and their model selection in the training loop, leading to controllable trade-offs of trust in the synthetic data. 

\paragraph{Trust Constraints in Learning Generative Models}
As we train generative models, incorporating trust dimension constraints in the training loop can guarantee better alignment between the models and the desiderata for safeguards concerning the trustworthiness of the synthetic data. We review in this Section different methodologies to infuse trust in the training of generative models.\\

\noindent The fidelity of generative models to real data is often the main objective in training synthetic data generators. This can be done for explicit likelihood models via cross-entropy training or auto-regressive training for sequential models, such as GPT models \cite{brown2020language} for text and TabGPT for tabular and tabular time series introduced in \cite{padhi2021tabular}. The main difference between tabular data and tabular time series is that for each field in the tabular data, we have a local vocabulary on which the cross-entropy training is performed. Other paradigms exist for implicit models that are defined via generators trained via variational auto-encoders\cite{kingma2013auto} and variational inference or via matching real and synthetic distribution in a min-max game between the generator and an adversary discriminator leading to the so-called Generative Adversarial Networks (GANs)\cite{goodfellow2020generative} .
Score-based training is yet another paradigm for training diffusion generative models \cite{song2019generative}.\\

\noindent Privacy preservation in learning generative models can be enforced via differential privacy \cite{dwork2014thealgorithmic}. An algorithm $A$ is $\varepsilon-$ differential private if its outputs are indistinguishable for any neighboring data-sets $D,D'$ that differ at a single element, i.e., if for all outputs subset $\mathcal{O}$: 
$$\log \left(\frac{\mathbb{P}(A(D) \in \mathcal{O})}{\mathbb{P}(A(D') \in \mathcal{O})}\right)\leq \varepsilon , \forall~ \text{neighboring } D,D'.$$
In order to use this concept in training generative models, Differential Private Stochastic Gradient (DP-SGD), introduced in \cite{dp-sgd-orig} and later specialized for learning deep learning model in \cite{abadi2016deep}, is a popular technique to induce differential privacy in synthetic data sampled from models trained with DP-SGD, thanks to the post processing property of differential privacy\cite{dwork2014thealgorithmic}. 
GAN models trained with DP-SGD result in the so called DP-GAN \cite{xie2018differentially}.
Scaling efficiently differential private SGD for learning large generative models, such as transformers and TabFormers, is challenging and this has been addressed recently in \cite{li2022large,li2022when}. We rely on this framework for scaling our training of differential private TabGPT for tabular and time series GPT models. Another framework for private training is Private Aggregation of Teacher Ensembles (PATE) \cite{papernot2018scalable} that has been adapted for learning the discriminator in GAN training in a private manner, leading to the PATE-GAN \cite{yoon2018pategan} model. Finally, an output perturbation approach can be employed to sample privately from non-private generative models via noising. This has been explored in \cite{majmudar2022differentially,NEURIPS2021_f2b5e92f}.An other important class of privacy preserving generative models for tabular data that won the NIST challenge is Differential Private Probabilistic Graphical Models (DP-PGM) \cite{mckenna2021winning} that learns to match noisy data marginals or 2-way marginals or more, if the target is included in the 2-way marginals we refer to it as DP-PGM (target). \\

\noindent In order to guarantee the utility of the generative model on a downstream task, conditional generative models are a key ingredient. At inference time, the model is conditioned or prompted with the target to ensure generation of labeled synthetic dataset that can be used later on to train a downstream classifier. Fairness in synthetic data can also be improved via de-biasing techniques\cite{nicolae2018adversarial,chuang2021fair}, reprogramming \cite{zhang2022fairness}, pre-processing \cite{calmon2017optimized}, or causal training \cite{van2021decaf}. Finally robustness of downstream classifiers trained on synthetic data transfers to real data and can be improved via adversarial training \cite{sehwag2022robust} .

\paragraph{Controllable Trust Trade-offs of the Synthetic Data via Trustworthiness Index Driven Model Selection} While integrating trust constraints in the training of the generative models can help in meeting certain aspects of trust, it is not clear how to perform model selection and how this impacts the trustworthiness of the downstream task of the synthetic data. 
To remedy this ambiguity, as we train a generative model to match a real dataset $D_r$, we sample synthetic datasets from checkpoints of this model $D^t_s$, where $t$ refers to the training time $t \in \{1,\dots, I_{\max}\},$ with $I_{max}$ being the maximum number of iterations. 
For example, $t$ can be the end of each epoch. We propose to track the trust dimension indices within the training loop:
 \begin{align}
 \pi_{\text{Fidelity}}(D_r, D^t_s,\text{\ttfamily{cfg}}), ~\pi_{\text{Privacy}}(D_r, D^t_s,\text{\ttfamily{cfg}}),
 ~\pi^{\text{val}}_{\text{Utility }}(D_r, D^t_s,\text{\ttfamily{cfg}}), ~ \pi^{\text{val}}_{\text{Fairness }}(D_r, D^t_s,\text{\ttfamily{cfg}}),
~\pi^{\text{val}}_{\text{Robustness }}(D_r, D^t_s,\text{\ttfamily{cfg}}),
 \label{eq:val_trust_indices}
  \end{align}

where utility, fairness and robustness indices are evaluated on the \textbf{validation set} of the real dataset. 
For a specific trade-off weighing $\omega$ between trust dimensions, we can therefore track the overall ``validation'' trustworthiness index:
\begin{align}
\tau_{\text{trust}}^{val}(D_r, D^t_s,\text{\ttfamily{cfg}},\omega)&=\omega_{f}~\pi_{\text{Fidelity}}(D_r, D^t_s,\text{\ttfamily{cfg}})\nonumber\\
&+ \omega_{P}~\pi_{\text{Privacy}}(D_r, D^t_s,\text{\ttfamily{cfg}}) +\omega_{U}~\pi^{\text{val}}_{\text{utility }}(D_r, D^t_s,\text{\ttfamily{cfg}})\nonumber\\
&+\omega_{F}~\pi^{\text{val}}_{\text{Fairness }}(D_r, D^t_s,\text{\ttfamily{cfg}}) +\omega_{R}~\pi^{\text{val}}_{\text{Robustness }}(D_r, D^t_s,\text{\ttfamily{cfg}})
\label{eq:val_trust_index}
\end{align}

\noindent With the validation trustworthiness index in hand, we can select the training time that produces the model and synthetic data that score best on validation trustworthiness index:
\begin{equation}
t^*= \arg\max_{t\in \{1,\ldots, I_{\max}\}} \tau_{\text{trust}}^{val}(D_r, D^t_s,\text{\ttfamily{cfg}},\omega),
\label{eq:selected_ckp}
\end{equation}
and finally return the audit of $D^{t^*}_{s}$, with utility, fairness, and robustness evaluated on \textbf{the real test set}. This process can be repeated within each split, if real data splits are available. 

\begin{figure}[htp]
\begin{minipage}{\textwidth}
\centering
\includegraphics[width=0.9\linewidth]{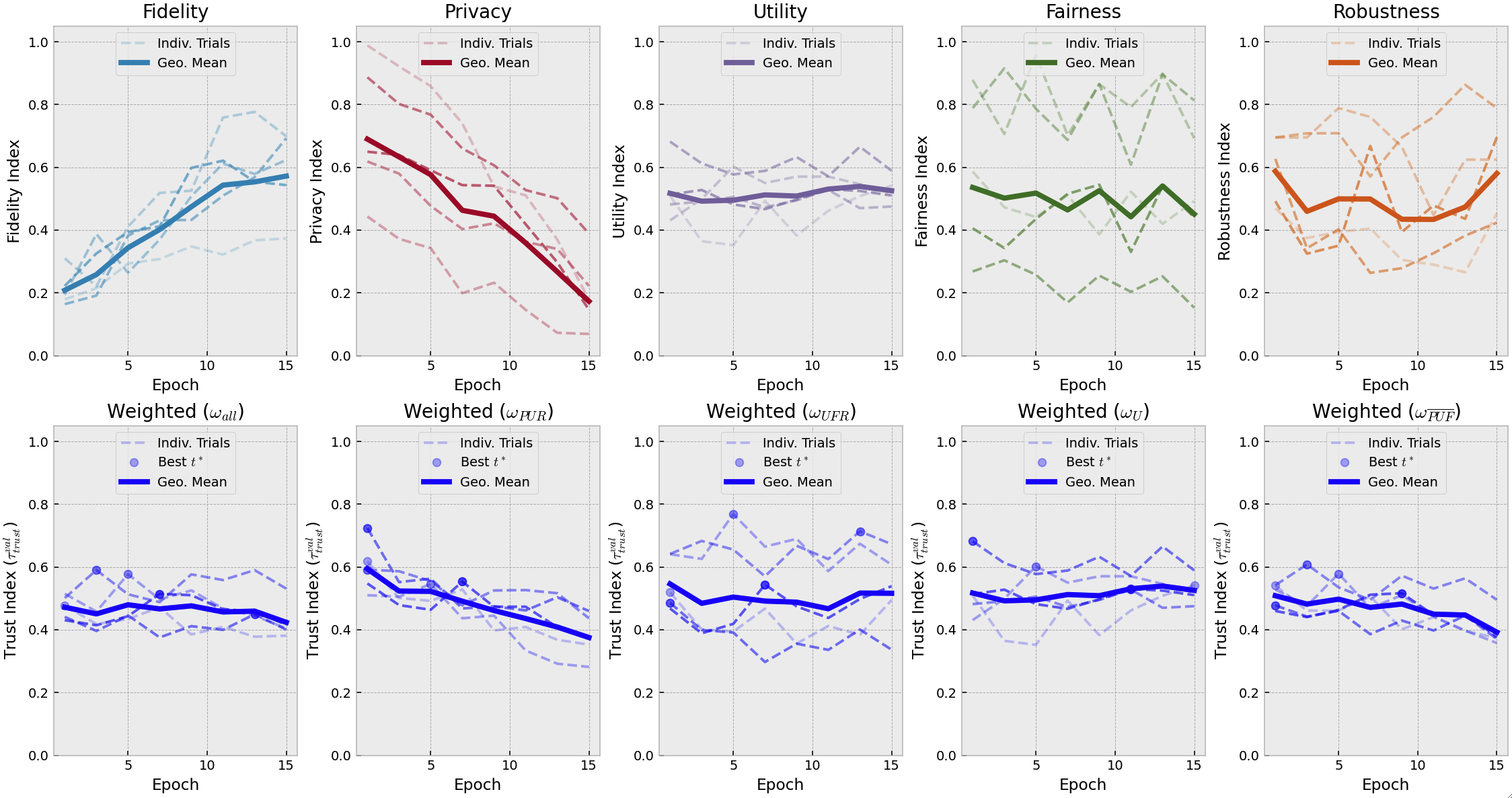} 
\caption{Controllable trust trade-offs via trustworthiness index driven cross-validation: Private TrustFormer TF(p,$\varepsilon=1$) (Bank Marketing dataset.)}
\label{Fig:checkpoints_private_bm}
\end{minipage}

\begin{minipage}{\textwidth}
\centering
\includegraphics[width=0.9\linewidth]{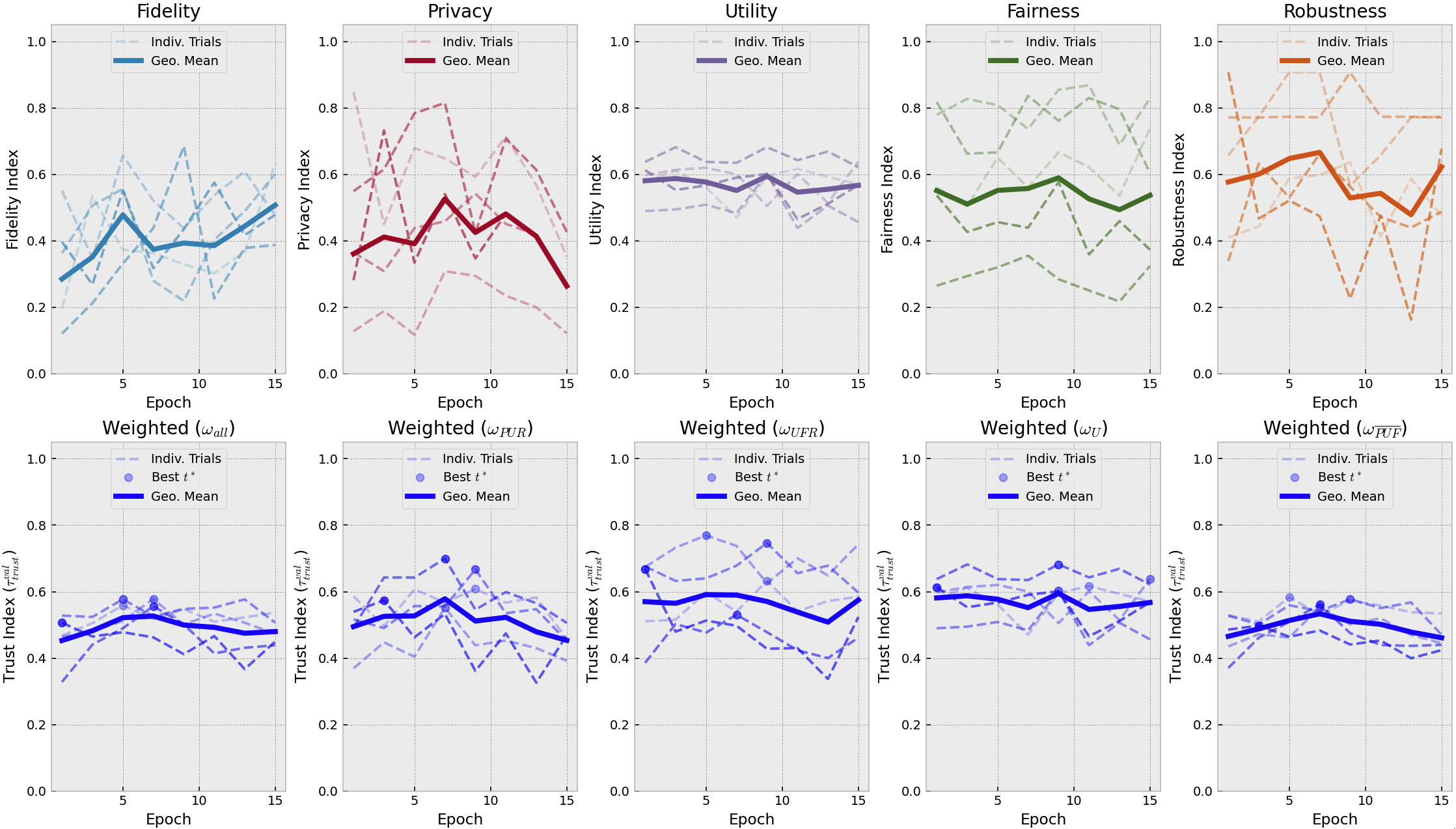}     
\caption{Controllable trust trade-offs via trustworthiness index driven cross-validation: Non-private TrustFormer TF(n-p) (Bank Marketing dataset.) }
\label{Fig:checkpoints_nonprivate_bm}
\end{minipage}
\caption{ \textbf{Controllable trust trade-offs.} In the first row of Figures \ref{Fig:checkpoints_private_bm} (Private TrustFormer) and \ref{Fig:checkpoints_nonprivate_bm} (Non-Private TrustFormer), we depict  the evolution of the \textbf{validation} trust dimensions indices (given in Equation \ref{eq:val_trust_indices})  as function of the training time (epoch) within the training of TF on 5 different real data splits (80\% train, 10\% validation, 10\% test, referred to in the plot as a trial). For the private case, we see that the fidelity index on average improves as training proceeds, while privacy index deteriorates on average as the training progresses. For the non-private case we see that while fidelity improves as training progresses, privacy has a sweet spot in the middle of the training. For both private and non private training, validation utility, fairness and robustness indices have different dynamics fluctuating around their geometric mean and have different peaks of performance within each trial. In the second row, we show for a trust trade-off weight ($\omega$)  the validation trustworthiness index given in Equation \eqref{eq:val_trust_index}, for each TF trained on a real data split as function of the training time (epoch). We plot in each panel, the validation trustworthiness index for $5$ different trade-offs weights $\omega_{all},\omega_{PUR}$,$\omega_{UFR}$, $\omega_{U},\omega_{e(PUF)}$ that are given in Table \ref{Table:weights}. We see that the selected epoch or checkpoint $t^*$ (given in Equation \ref{eq:selected_ckp}) is different for each training split and for each trade-off weight. This model selection, using the trustworthiness index  allows the alignment of the generative model with the desired trust trade-off. }
\end{figure}

\paragraph{TrustFormer: Instrumenting Transformers with Trust} This trustworthiness index driven model selection is applicable to any generative modeling method, however for the sake of demonstrating a concrete and useful example, we focus on instrumenting transformers models with it. Owing to their versatility, transformer models and their variants provide state of the art generation quality across the different modalities we consider here, namely tabular, time series, and natural language. We refer to transformer models trained under trust constraints and selected via our trustworthiness index as \textbf{TrustFormer}. Figures~\ref{Fig:checkpoints_private_bm} and \ref{Fig:checkpoints_nonprivate_bm}  
showcases model selection of TabFormer GPT 
models \cite{padhi2021tabular
} ( private and non-private, TF(p,$\varepsilon=1$) and TF(n-p) respectively), based on the validation trustworthiness index corresponding to trade-off weights $\omega$. We see that indeed this trustworthiness index driven cross-validation leads to different model checkpoints being selected as $\omega$ varies, translating to an improved trustworthiness index when evaluated on real test data, thereby controlling the trust trade-offs of the resulting synthetic dataset. Note that this trustworthiness index selection is not only applicable for checkpoint selection, for example, given a particular checkpoint one can select the sampling strategy and other hyper-parameters of the inference that lead to the highest validation trustworthiness index.

\paragraph{Putting It All Together: From Training to Model \& Data Selection to Auditing } 
 We train our TrustFormer GPT models in unsupervised way using auto-regressive training. For tabular data and tabular time series, discrete valued fields are tokenized and continuous ones are quantized to obtain a discrete vocabulary for all fields \cite{padhi2021tabular}. Trust constraints are imposed in the training and the inference of TrustFormer GPT as discussed previously. For example, we use Differential Private training for privacy \cite{li2022large,li2022when} and balanced sampling for fairness. \\
 
 \noindent Given that some metrics in the audit rely on an embedding $\textbf{E}$, we also train a TrustFormer RoBERTa in an unsupervised way using a masked loss function. For NLP data, RoBERTa training is performed using the classical token level masking. For tabular and time series data the masking is applied at the field level and prediction is performed on field level vocabularies \cite{padhi2021tabular}. We distinguish between three types of embeddings below:
 \begin{enumerate}
 \item Non-private TrustFormer RoBERTa trained on real data with an unsupervised masking loss function (See Tabular use cases).
 \item Non-private TrustFormer RoBERTa trained on public data that is not overlapping with sensitive real data (See NLP use case). 
 \item Private TrustFormer RoBERTa trained on real data with Differential Private SGD \cite{li2022large} (see Fraud Detection use case). 
 \end{enumerate}

\noindent Note that all our embeddings are trained on real or public data and not on synthetic data. \\

\noindent We highlight here that within our auditing framework we perform two types of cross-validation and selection:
\begin{enumerate}
\item \textbf{\emph{Classifier Selection}.}
We train our utility classifiers on real or synthetic data using the embeddings discussed above. We perform early stopping in order to select the best performing classifier on the real validation set. 
\item \textbf{\emph{Trustworthiness index  driven Synthetic Data Selection}.} In order to obtain training data with controllable trust trade-offs for utility auditing, we use our trustworthiness index to select TrustFormer GPT checkpoints that lead to the highest trustworthiness index given the trade off weights. Note that within the utility audit  we use the classifier selection described above to select the best classifier for the downstream task. 
\end{enumerate}
For downstream tasks trained on real data, we rely on classifier selection alone. For tasks trained on synthetic data, we rely on both classifier selection and trustworthiness index  driven synthetic data selection. In the next Section, we will see that these controllable trust trade-offs via synthetic data selection allow the synthetic data to have improved utility, fairness, and robustness with respect to the real data. These findings are novel and highlight the importance of a holistic auditing framework and its promise in cross-validating generative models.

\subsection*{Assets and License}\label{app:assets}

\begin{center}
\resizebox{\textwidth}{!}{\begin{tabular}{ l|l|l }
\toprule
 Asset & License & Link \\
\midrule
AIF360 & Apache 2.0 License
& https://github.com/Trusted-AI/AIF360/blob/master/LICENSE \\
HuggingFace Datasets & Apache 2.0 License & https://github.com/huggingface/datasets/blob/master/LICENSE \\
 Nvidia APEX & BSD 3-Clause "New" or "Revised" License &https://github.com/NVIDIA/apex/blob/master/LICENSE \\ 
 Pytorch & BSD Style License & https://github.com/pytorch/pytorch/blob/master/LICENSE \\
 Pytorch Lightning & Apache 2.0 License & https://github.com/PyTorchLightning/pytorch-lightning/blob/master/LICENSE \\
 Faiss & MIT License & https://github.com/facebookresearch/faiss/blob/main/LICENSE \\
 synthcity & Apache 2.0 License & https://github.com/vanderschaarlab/synthcity/blob/main/LICENSE \\
 Synthetic Data Vault (SDV) & Business Source License 1.1 & https://github.com/sdv-dev/SDV/blob/master/LICENSE \\
 nist-synthetic-data-2021 (DPPGM) & Apache 2.0 License & https://github.com/ryan112358/nist-synthetic-data-2021/blob/main/LICENSE \\
 Private Transformers & Apache 2.0 License& https://github.com/lxuechen/private-transformers\\
scikit learn & BSD license & https://scikit-learn.org/stable/about.html\\
\hline
\end{tabular}}
\end{center}

\subsection{Implementation Details for Training TrustFormers}

\subsection{Feature Extractor}

\noindent Table~\ref{tab:arch_fea} provides for each dataset the list of hyperparameters used in training for RoBERTa models used as an embedding \textbf{E} in the auditing.

\begin{table}[ht!]
\centering
\begin{tabular}{lccccccccccc}
\toprule
 Dataset &  $N_e$ & $|V|$ & $\lambda_r$ & $b_s$ & $\mathrm{h_F}$ &  $\mathrm{AH_F}$  & $\mathrm{h}$ & $\mathrm{AH}$ & window & stride & $p_\text{mlm}$ \\
 \midrule
 \textbf{Tabular} & &  &  &  &  &  & & &\\
Bank Marketing    & 50 & 2,073  & $10^{-4}$  & 64  & 32 & 8 & 640 & 8 & 1 & 1 & 0.2 \\
Recruitment       & 50 & 248 & $10^{-4}$  & 64 & 65 & 13 & 845 & 13 & 1 & 1 & 0.2 \\
Law School        & 50 & 210 & $10^{-4}$  & 64 & 40 & 10 & 400 & 10 & 1 & 1 & 0.2 \\
 \midrule
 \textbf{Time-Series} &  &   &  &  &  &  &  &  &\\
 Credit Card                &50 &26047  & $10^{-5}$  & 128 & 64 & 8& 768& 12& 10& 10&0.2\\
 MIMIC-III IHM        &  40 &  4,697 & $10^{-5}$ & 128 & 112 & 8 & 2464 & 8 & 48 & 48 & 0.1\\
 \midrule
 \textbf{Text}     & & & & & & & & &\\
MIMIC-III Notes     & - & 57,716 & - & - & - & - & 1600 & 25 & - & - & -\\
\end{tabular}
\setcounter{table}{0}
\caption{Feature extractor RoBERTa models: columns are $N_e$ for number of epochs, $|V|$ for vocab size, $\lambda_r$ for learning rate, $b_s$ for batch size, $\mathrm{h_F}$ for field hidden size, $\mathrm{AH_F}$ for field number of attention heads, $h$ for hidden size, AH for number of attention heads, window size, stride, and $p_{\text{mlm}}$ for probability for masked language model. Please note that for MIMIC-III Notes dataset, our feature extractor is a pre-trained BioGPT-Large model. }
\label{tab:arch_fea}
\end{table}

\subsection{GPT}

\noindent Table~\ref{tab:arch_gen} provides for each dataset the list of hyperparameters used in training for GPT models.

\begin{table*}[ht!]
\centering
\begin{tabular}{lccccccccc}
\toprule
 Dataset & $|V|$ & $\lambda_r$ & $b_s$ & window & stride & $N_e$ & DP & DP-$\epsilon$ & $\max(||\Delta||)$ \\
 \midrule
 \textbf{Tabular} &  &  &  &  &  &  & &  &\\
\multirow{2}{*}{Bank Marketing} &  \multirow{2}{*}{2075}  & \multirow{2}{*}{$10^{-4}$}  & \multirow{2}{*}{64}   & \multirow{2}{*}{1}   & \multirow{2}{*}{1} & 25 & no & - & - \\
&  &  &  &  & & 20 & yes & [1, 3]  & 0.01 \\
\multirow{2}{*}{Recruitment}    & \multirow{2}{*}{250}  & \multirow{2}{*}{$10^{-4}$}  & \multirow{2}{*}{64}   & \multirow{2}{*}{1}   & \multirow{2}{*}{1} & 25  & no & -  & - \\
  &  &  &  &  & & 20 & yes & [1, 3]  & 0.1 \\
\multirow{2}{*}{Law School}    & \multirow{2}{*}{212} & \multirow{2}{*}{$10^{-4}$}  & \multirow{2}{*}{64}   & \multirow{2}{*}{1}   & \multirow{2}{*}{1} & 25  & no &  -  & - \\
  &  &  &  &  & & 20 & yes & [1, 3]  & 0.1 \\
 \midrule
 \textbf{Time-Series} &      &  &  & &  &  & \\
 Credit Card                 & 26132 & $10^{-5}$ &64  &10  &10& 25& no&--&-- \\
  \multirow{2}{*}{MIMIC-III IHM}         &  \multirow{2}{*}{4,699}  & \multirow{2}{*}{$10^{-5}$} & \multirow{2}{*}{16} & \multirow{2}{*}{48} & \multirow{2}{*}{48} & 25 & no & -- & -- \\
 &   &  &  &  &  & 25 & yes & [3] & 0.01 \\

 \midrule
 \textbf{Text}     &  & & & & & & \\
\multirow{1}{*}{MIMIC-III Notes$^{\dagger}$}     & \multirow{1}{*}{57,716}  &$10^{-5}$ &16 &-- &-- & 10&no& --&-- \\
 &  & & & & & & \\
\end{tabular}
\caption{Generator GPT models: columns are $|V|$ for vocab size, $\lambda_r$ for learning rate, $b_s$ for batch size, window size, stride, $N_e$ for number of epochs, DP enabled (yes/no), DP $\epsilon$ budget, and $\max(||\Delta||)$ for DP max gradient norm.$\dagger$ for distributed training on 8 GPUs. }
\label{tab:arch_gen}
\end{table*}

\subsection{Details For Compute Environment and Distributed Training} 
All experiments were performed on a GPU cluster where each node contains either $8$ NVIDIA Tesla V100 (32GB) or $8$ Ampere A100 (40GB) GPUs connected via NVLink. The V100 nodes are equipped with dual 28-core (Intel Xeon Gold 6258R) CPUs, the A100 nodes are equipped with dual 64-core (AMD EPYC 7742) CPUs, and all nodes are connected by 2 non-blocking EDR InfiniBand (100Gbps) network adapters as well as 2 100Gbps Ethernet adapters. All nodes are installed with RHEL 8.3, CUDA 10.2, and cuDNN 7.5. 

\noindent Training for tabular and time-series TrustFormers was done on single GPU. For BioGPTLarge finetuning and non private training of Tabular RoBERTa and GPT on the credit card  we relied on on the Distributed Data Parallel functions provided by Pytorch and Pytorch Lightning utilizing the NCCL backend using 8 GPUs. Due to known incompatibilities between the Differential Privacy library\cite{li2022large} implementation and Distributed Data Parallel training, all private training was limited on single GPUs. Fully Sharded Data Parallel training may be able to overcome this limit and is an avenue for future work.

\subsection{Timing of Training, Auditing and Aggregation }
\label{supp:timing}
Table~\ref{tab:timings} provides timings for training both feature extractor and generator models, auditing, and aggregation of results for all our datasets.
One can notice that Differential Privacy trainings comes at a computational cost making the training time per epoch longer.

\begin{table}[ht!]
\centering
\begin{tabular}{lccccc}
\toprule
Dataset & Fea Ext & GPT (non-private) & GPT (private) & Audit & Aggregation \\
 & (h/epoch)  & (h/epoch) & (h/epoch) & (h) & (h) \\ 
\midrule
Bank Marketing      &  0.140    & 0.056   & 0.063   &   2.51    &    0.003 \\
Recruitment         &  0.026    & 0.009   & 0.012   &   0.47    &  0.004    \\
Law School          &  0.042    & 0.015   & 0.020   &   0.98    &   0.003   \\
MIMIC-III IHM       & 0.096  & 0.71 & 1.26 & 0.75   &   0.002\\
\bottomrule
\end{tabular}
\caption{Timings for trainings of our feature extractor models (Fea Ext), GPT models for both non-private and private trainings, audit of synthetic data, and aggregation of results. All training timings are given as hours per epoch while audit timings and aggregation are given in hours.}
\label{tab:timings}
\end{table}

\subsection{Private Sampling from TabGPT}
\label{sec:private_sampling}
\begin{algorithm}
    \caption{Private Sampler}
    \label{alg:logit}
    \hspace*{\algorithmicindent} \textbf{Input:} logits $\in \mathbb{R}^{\text{batch} \times |V_G|},  \varepsilon>0, \text{attribute} j$\\
    \hspace*{\algorithmicindent} \textbf{Output:} Updated logits
    \begin{algorithmic}[1]
            \STATE $\mathbf{P} \gets \text{Softmax(logits)}$ \COMMENT{Convert to probabilities from logits space}
            \STATE $\mathbf{P}_{L} \gets \text{Select}(\mathbf{P}, j) \quad \mathbf{P}_{L} \in \mathbb{R}^{\text{batch} \times |V_L|}$ \COMMENT{Select probabilities for local vocabulary $L$ corresponding to attribute $j$}
            \STATE $\mathbf{N} \sim \text{Lap}(\frac{2T}{\varepsilon |V_L|})$ \COMMENT{Sample Laplace noise}
            \STATE $\mathbf{P}^{noisy}_{L} \gets \mathbf{P}_{L} + \mathbf{N}$ \COMMENT{Add noise}
            \STATE $\mathbf{P}^{\text{proj}}_{L} \gets \text{Project}(\mathbf{P}^{noisy}_{L})$ \COMMENT{Project probabilities to probability simplex}
            \STATE $\mathbf{P} \gets \text{Insert}(\mathbf{P}, \mathbf{P}^{\text{proj}}_L, j)$ \COMMENT{Insert updated probabilities}
            \STATE {$\text{logits} \gets \log(\mathbf{P})$} \COMMENT{Convert to logits from probability space}
            \RETURN logits
    \end{algorithmic}
\end{algorithm}

\noindent To generate synthetic data for our fraud classifier, we adapted a differentially private sampler from \cite{NEURIPS2021_f2b5e92f}. In our setting, each sample consists of 10 transactions, which represent the window size. Every transaction contains 12 attributes, each with its distinct vocabulary. The algorithm presented above demonstrates the sampling process at a specific time step $t$, where logits for each attribute are preprocessed separately.

\noindent Given the logits of size $\mathbb{R}^{\text{batch} \times |V_G|}$, where $|V_G|$ denotes the global vocabulary size encompassing all attributes, we initially extract the logits corresponding to attribute $j$ to obtain $\mathbf{P}_{L} \in \mathbb{R}^{\text{batch} \times |V_L|}$. Subsequently, we sample Laplace noise with variance determined by the local vocabulary size, privacy parameter $\varepsilon$, and total sequence length $T$. After adding the noise to the probabilities, we project them back onto the probability simplex.

\noindent Finally, we reinsert the updated probabilities into their original positions and convert them back into the logits space. These logits are then employed to sample attribute $j$ at time step $t$.
\subsection{Baselines Training Details}
Multiple synthetic data generation techniques were employed to generate a varied set of baseline synthetic datasets to compare against TrustFormer generated synthetic datasets for quality assessment. Non private as well as diferentially private baselines were generated as seen in Table \ref{Table:baselines}. 

\subsection{Non Private Baselines}
Non private baselines were generated using Synthetic Data Vault \cite{SDV} (SDV) a python library developed by MIT and now maintained by datacebo. Two datasets were generated, one using a statistical method called Gaussian Copula and the other using a Conditional GAN. We used the default hyperparameters for the Gaussian Copula training and synthetic data was sampled from the trained model using a batch size of 100. The conditional Tabular GAN model configuration was as follows:
\textbf{epochs: 500},
\textbf{batch size: 100}, 
\textbf{generator dimensions: (256, 256, 256)}, 
\textbf{discriminator dimensions: (256, 256, 256)}.

\subsection{Private Baselines}
Differentially private, baseline synthetic datasets, were generated using privacy preserving GANs available in synthcity \cite{qian2023synthcity} and a NIST challenge winning probabilistic graphical method that uses adaptive grid. Using methods described below datasets were generated to preserve privacy with epsilon 1 and 3.\\
\noindent Synthcity is also a python library for generating and evaluating synthetic tabular data consisting of several generative models. We chose DP-GAN (Differentially Private GAN) and PATEG-AN (Private Aggregation of Teacher Ensembles GAN) as the two methods to generate synthetic data. For both of the models we specify the target i.e label in the dataset allowing the model to be conditioned by it. 
Differentially Private Probabilistic Graphical Model (DP-PGM) involved two steps. First we created a data representation (quantization) that could be inputed to the model using their schema generator and preprocessor which was followed by probability estimation and generation of synthetic data. Multiple baselines were generated with and without special consideration of the label(target) leading to higher importance in marginal preservation.

\subsection{Implementation Details For Metrics }\label{sec:metricdetails}

\begin{figure}[ht!]
\centering
\includegraphics[scale=0.5]{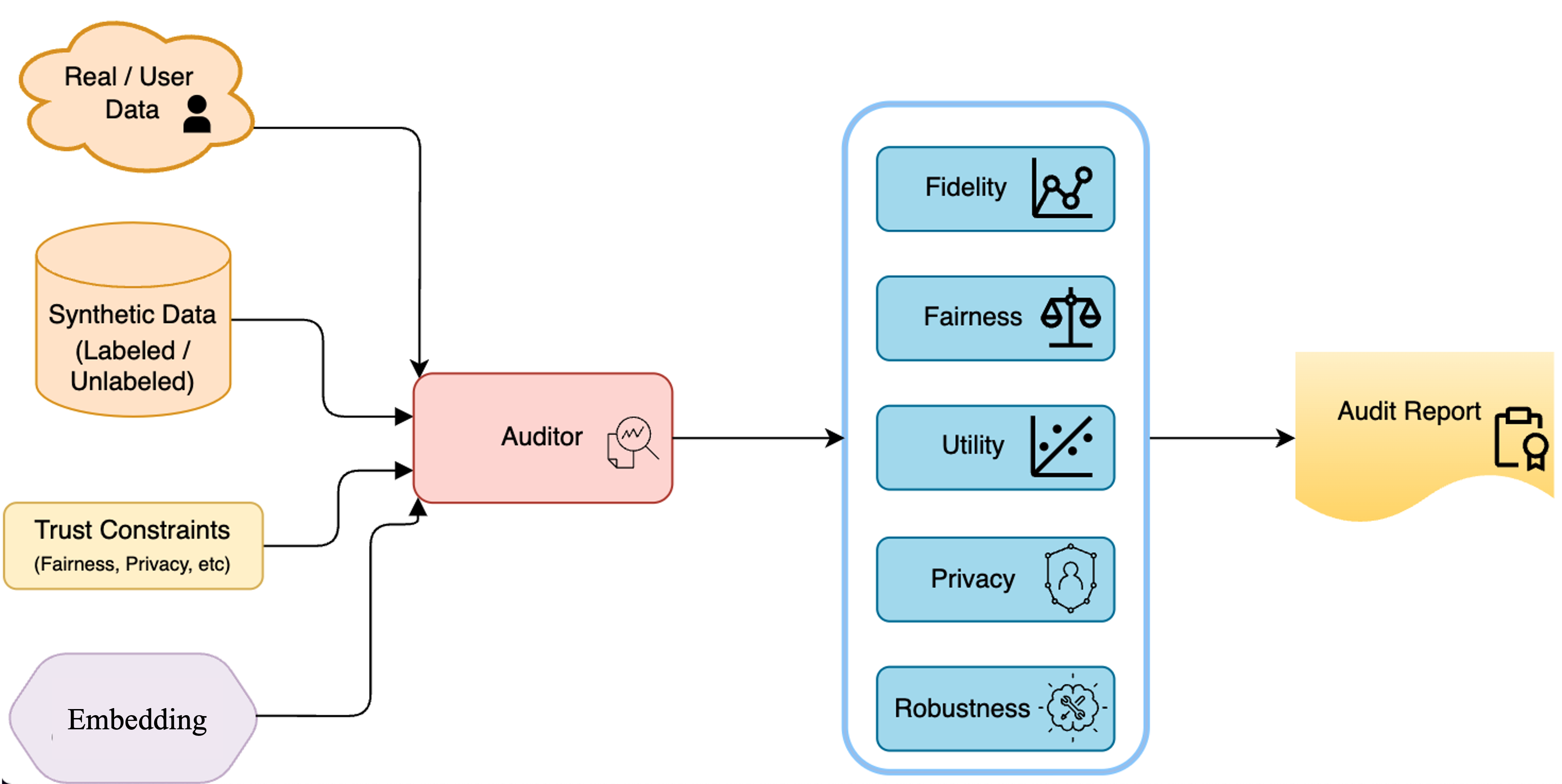} 
\setcounter{figure}{0}
\caption{Auditing the trustworthiness synthetic data form metric evaluations on trust dimensions to the audit report.}
\end{figure}

\subsection{Fidelity}
The fidelity of synthetic data quantifies how closely the synthetic data resembles the real data.
Fidelity metrics are therefore naturally expressed in terms of differences between the real distribution $P_{r}$ and the synthetic distribution modeled by the generator $P_{s}$ using dataset $D_{r}$ and $D_{s}$ sampled from these distributions, respectively.
Most of the fidelity metrics are evaluated in the feature space corresponding to embedding the data samples with a trained transformer encoder like RoBERTa.
We denote the resulting feature vectors of the real and synthetic samples by $\phi_r$ and $\phi_s$, respectively, and
the corresponding sets of feature vectors by $\Phi_r$ and $\Phi_s$.
With this notation defined, these are the metrics that we used to quantify fidelity:
\begin{itemize}
    \item {\bf Chi-squared.}
    This metrics operates directly on the real data $D_{r}$ and synthetic data $D_{s}$ \emph{before} it is embedded into the feature vectors $\Phi_r$ and $\Phi_s$.
    Concretely, for each field of our tabular datasets we compute the histogram for real and synthetic data and compute the Chi-squared distance between the two.
    Chi-squared distance is computed as follows:
    $\chi^2(c_r, c_s) =\frac{1}{2}\sum_{i=1}^{n}\frac{(c_r(i) - c_s(i))^2}{(c_r(i) + c_s(i))},$ where $c_r$ and $c_s$ are the histograms from the real and synthetic data for a particular field.
    \item {\bf $\ell_2$ Mutual Information Difference.}
    Mutual information (MI) is another metric that operates directly on the tabular data by measuring dependencies between different columns.
    For both real and synthetic data, we compute MI pairwise between all columns as follows.
    Given two columns $i$ and $j$ we discretize the table jointly according to the values of $i$ and $j$ and obtain the joint histogram $c(i,j)$.
    MI is then computed as
    $I(i;j) = \sum_{i,j} c(i,j) \log\left(\frac{c(i,j)}{c(i)c(j)}\right)$, where $c(i)=\sum_j c(i,j)$ and $c(j)=\sum_i c(i,j)$ are the marginalized histograms.
    For both real and synthetic datasets we then build an MI matrix running the above calculation for all pairs of columns $i,j$ and then compute the $\ell_2$ distance between the resulting MI matrices for the real and synthetic datasets.
    \item {\bf Maximum Mean Discrepancy (MMD) SNR.}
    MMD is a state-of-the-art nonparametric test for tackling the two-sample problem.
    Its statistic is computed by computing the difference in expectation of a witness function defined as the mean of kernel evaluations on a set of basis points. Following \cite{kubler2022witness} we optimize the witness function on data obtained by splitting training sets out of $D_{r}$ and $D_{s}$.
    For efficiency and simplicity, as witness function we use a random Fourier embedding followed by a linear function, which corresponds to a Gaussian kernel \cite{rahimi2007random}.
    In practice, learning the witness function boils downs to performing Linear Discriminant Analysis in the random Fourier embedding space by trying to maximize a signal-to-noise-ratio (SNR) of a discriminator between the featurized training samples $\Phi_{r}$ and those from $\Phi_{s}$.
    The same SNR can be computed using the learned witness function on the test splits from $\Phi_{r}$ and $\Phi_{s}$. This gives us a cross-validated measure of difference between the datasets $\Phi_{r}$ and $\Phi_{s}$; the higher it is, the more different the corresponding real $D_{r}$ and the synthetic datasets $D_{s}$ are declared to be.
    \item {\bf MMD test $p$-value.}
    The statistical significance of the SNR computed with the MMD two-sample test above can be quantified with a permutation test which provides a $p$-value.
    In practice, we re-compute the SNR ratio between the test splits from  $\Phi_{r}$ and $\Phi_{s}$ using the fixed learned witness function multiple times by randomly re-assigning samples between the two test splits.
    This allows for the construction of a distribution of SNR values across realization of the permutation, which in turn allows for the derivation of a $p$-value by counting the fraction of permutations for which the SNR is lower than the originally computed value.
    \item {\bf Fréchet Inception Distance (FID).}
    This is a metric introduced in the GANs literature in \cite{heusel2017gans} which essentially computes the Fréchet distance \cite{frechet1957distance} between the \emph{featurized} datasets $\Phi_r$ and $\Phi_s$ under the approximation that they are sampled from multi-variate Gaussians.
    This gives the closed-form expression:
    $\mathrm{d_F}(D_r, D_s) = ||\mu_r - \mu_s||^2_2 + \mathrm{tr}\left(\Sigma_r + \Sigma_s - 2 \left(\Sigma^{\frac{1}{2}}_r \cdot \Sigma_s \cdot \Sigma^{\frac{1}{2}}_r\right)^{\frac{1}{2}}\right)$,
    where $\mu_r$ and $\Sigma_r$ are the sample mean and covariance of the featurized samples in $\Phi_r$, and $\mu_s$ and $\Sigma_s$ are the sample mean and covariance of the featurized samples in $\Phi_s$.
    \item {\bf Precision/Recall.}
    The generative models literature recently noted the necessity of quantifying the quality of generated samples along two distinct dimensions: the quality of the individual generated samples in terms \emph{faithfulness} with respect to the real distribution, and the \emph{diversity} of the distribution of generated samples whose variation should match that of the real distribution.
    The paper \cite{sajjadi2018assessing} proposed that these two aspects could be quantified in terms of \emph{precision} and \emph{recall}, which intuitively correspond to the average sample quality and the coverage of the generated distribution, respectively.
    We follow the improved implementation of precision and recall by \cite{kynkaanniemi2019improved} which has been developed to emphasize the trade-off between sample quality and diversity.
    In particular, we use the expressions 
    \begin{align*}
        \mathrm{precision}(\Phi_r, \Phi_s)=\frac{1}{|\Phi_s|}\sum_{\phi_s\in\Phi_s}f(\phi_s,\Phi_r) \qquad\qquad
        \mathrm{recall}(\Phi_r, \Phi_s)=\frac{1}{|\Phi_r|}\sum_{\phi_r\in\Phi_r}f(\phi_r,\Phi_s),
    \end{align*}
    with 
    \begin{equation*}
        f(\phi,\Phi)=
        \begin{cases}
          1, \text{ if } ||\phi-\phi'||_2\le ||\phi'-\mathrm{NN}_k(\phi',\Phi)||_2 \text{ for at least one }\phi'\in\Phi\\    
          0, \text{ otherwise,}
        \end{cases}
    \end{equation*}
    where $\mathrm{NN}_k(\phi',\Phi)$ returns the $k$-th nearest vector $\phi'$ in the set $\Phi$.
    The intuition here is that $f(\phi_s,\Phi_r)$ quantifies whether $\phi_s$ is close to the manifold from which $\Phi_r$ was sampled, meaning that \emph{precision} quantifies how realistic the generated samples $\Phi_s$ are on average.
    Other hand, $f(\phi_r,\Phi_s)$ says whether a vector $\phi_r$ corresponding to a real sample could be produced by the generator, meaning that \emph{recall} quantifies the diversity of the generated data in terms of how well it covers the real dataset.
\end{itemize}

\subsection{Privacy}
These metrics help evaluate and quantify information leakage from the real data to synthetic data. The two metrics we have identified to measure privacy are based on L2 distance between data points in the real and synthetic datasets. Consider the real dataset with $N$ datapoints and k features $D_{real} = \{\mathbf{x}_{(i)}\}_{i=1}^N$
where $\mathbf{x_{(i)}} = {x_{i0}, x_{i1}, \dots, x_{ik}}$ and synthetic dataset $D_{synth} = \{\mathbf{y}_{(i)}\}_{i=1}^N$ where $\mathbf{y_{(i)}} = {y_{i0}, y_{i1}, \dots, y_{ik}}$ .
The L2 distance between a point in the real data and synthetic data can be represented as $ dist(x_{(i)},y_{(j)}) = ||\mathbf{x_{(i)}} - \mathbf{y_{(j)}}||_2 = \sqrt{(x_{i0} - y_{j0})^2 + (x_{i1} - y_{j1})^2 + \ldots + (x_{ik} - y_{jk})^2}$ .
\begin{itemize}
    \item {\bf{Replicated Rows}.} This metric identifies whether there is an exact copy of rows from $D_{real}$ in $D_{synth}$. For every row in the raw feature space of $D_{synth}$, L2 distance $dist(x_{(i)},y_{(j)})$ between all rows of $D_{real}$ in raw feature space is calculated. A count of rows from $D_{synth}$ which have exact matches with rows in $D_{real}$ i.e. $dist(x_{(i)},y_{(j)}) = 0$ is reported.
    \item {\bf{Nearest Neighbor Distance}.} This metric identifies nearest neighbors [1,3,5] in the real data $D_{real}$ for every row in the synthetic data $D_{synth}$. This is done both in the raw feature space as well as RoBERTa embedding space. For these neighbors, we provide aggregate statistics such as mean, median, mode and standard deviation of the distances for each set [1,3,5] of nearest neighbors. For 3 and 5 neighbors, we use the median distance among the neighbors for computing the aggregate statistics. Even if rows in $D_{synth}$ aren't replicas of rows in $D_{real}$, they may be very close, which still shows leakage and this metric helps detect such a problem. For the nearest neighbor computation we have used Faiss which helps in an efficient search in large feature spaces as well.
\end{itemize}

\subsection{Utility} 
The utility metric that we consider are meant to quantify both the quality of the synthetic generated data and of the RoBERTa encoder in terms of the performance that they enable in downstream classification tasks.
Analogously to most of the fidelity metrics above, utility metrics are therefore evaluated on the sets of feature vectors $\Phi_r$ and $\Phi_s$ obtained by embedding the real data $D_r$ and synthetic data $D_s$, respectively.
For the metrics performance on the classification task was assessed in terms of accuracy, precision, recall, and F1 score.
In order to quantify the variability due to different initializations of the utility architectures we run each training loop with 5 times with 5 different random seeds and report aggregates across these 5 realizations.
\begin{itemize}
    \item {\bf Linear Logistic Regression.}
    This utility metric corresponds to training a simple Logistic Regression model on top of the feature vectors.
    We used \textsc{scikit-learn} \cite{pedregosa2011scikit} to implement the Logistic Regression model.
    We first apply the standardize the individual features.
    Unless otherwise stated, we also reduce the dimensionality of the problem by selecting the top 100 features according to their ANOVA F-value (using the \texttt{f\_classif} feature selection function of \textsc{scikit-learn}).
    We also considered non-linear feature selection methods (e.g., \cite{mroueh2019sobolev}), but settled for this simpler one.
    \item {\bf Nearest Neighbor classification.}
    We also implemented Nearest Neighbor Classification as a classifier.
    In practice, a test feature vector is assigned a labeled corresponding to the label of the closest training sample (in L2-norm).
    For scalability, we implemented nearest neighbor retrieval using the approximate nearest neighbor library FAISS \cite{johnson2019billion}.
    \item {\bf MLP classification.}
    Utility also includes an MLP classifier consisting in a 2-layer network with a hidden layer using Batch-Normalization \cite{ioffe2015batch} and ReLU non-linearity.
    The network is optimized using the Adam optimizer \cite{kingma2014adam} with learning rate generally set to $3.0\times 10^{-4}$ unless otherwise stated, and use early-stopping with patience of 3 on F1-score computed on the hold-out set.
    \item {\bf MLP classification / Adversarial debiasing.}
    Our MLP classifier optionally includes the option of using adversarial debiasing \cite{zhang2018mitigating} as a method to mitigate bias.
    This method essentially corresponds to introducing an adversary network that that is trained to infer the protected labels of the value of the protected variables of the input samples.
    In our implementation the adversary has the same architecture as the original network, but a hidden size that is 4 times smaller, and is optimized by alternating its update with the update of the original network.
    Adversarial debiasing then adds an adversarial loss term to the loss of the original MLP classifier that corresponds to the accuracy of the adversary (see \cite{zhang2018mitigating} for more details).
    \item {\bf MLP classification / Fair Mixup.}
    Another debiasing method that we implemented is Fair Mixup \cite{chuang2021fair}, which  uses data augmentation and adds a gradient regularizer to the original loss in order to  mitigate the bias with respect to protected variables (see cited paper for details).
\end{itemize}

\subsection{Fairness}
To evaluate fairness metrics, we require a dataset with a designation as to which items belong to a privileged group.
Formally, for a dataset $D = \{(\mathbf{x}^{(i)}, y^{(i)}, \rho^{(i)})\}_{i=1}^N$ with $N$ data points each having features $\mathbf{x} \in \mathcal{X},$ a binary classification label $y \in \{0, 1\},$ and an indicator $\rho \in \{0, 1\}$ as to whether the data point belongs to a privileged group or not, we split the dataset into $D_{privileged} = \{(\mathbf{x}^{(i)}, y^{(i)}, \rho^{(i)}) \mid \rho^{(i)} = 1\}$ and $D_{unprivileged} = \{(\mathbf{x}^{(i)}, y^{(i)}, p^{(i)}) \mid \rho^{(i)} = 0\}$.
We then use a classifier $f: \mathcal{X} \rightarrow \{0, 1\}$ to compute the true and false positive rates for both the privileged and unprivileged dataset splits.
We let the classifier $f$ come from one of three classes for functions: k-NN classifier ($f_{KNN}$), logistic regression classifier ($f_{LR}$), or multi-layer perceptron classifier ($f_{MLP}$).
For a given dataset $D'$ and classifier $f,$ we define the true ($\mathrm{TPR}$) and false positive rates ($\mathrm{FPR}$) as: 
\begin{align*}
    &\mathrm{TPR}_{D'}(f) = \frac{\sum_{i=1}^{|D'|}\mathbb{1}(f(\mathbf{x}^{(i)}) = 1)\mathbb{1}(y^{(i)} = 1)}{\sum_{i=1}^{|D'|}\mathbb{1}(y^{(i)} = 1)} \\
    &\mathrm{FPR}_{D'}(f) = \frac{\sum_{i=1}^{|D'|}\mathbb{1}(f(\mathbf{x}^{(i)}) = 1)\mathbb{1}(y^{(i)} = 0)}{\sum_{i=1}^{|D'|}\mathbb{1}(y^{(i)} = 0)}
\end{align*}
where $\mathbb{1}(\cdot)$ is the indicator function that equals 1 if the argument is true and 0 otherwise.
Letting
\begin{align*}
    \Delta_{\mathrm{TPR}}& = \mathrm{TPR}_{D_{privileged}} - \mathrm{TPR}_{D_{unprivileged}} && \Delta_{\mathrm{FPR}}\\
    &= \mathrm{FPR}_{D_{privileged}} - \mathrm{FPR}_{D_{unprivileged}}
\end{align*}
for each classifier $f \in \{f_{KNN}, f_{LR}, f_{MLP}\}$, we report three metrics based on these true / false positive rate differences:
\begin{align*}
    \text{Equal Opportunity Difference} &= \Delta_{\mathrm{TPR}} \\
    \text{Average Odds Difference} &= \frac{1}{2}(\Delta_{\mathrm{TPR}} + \Delta_{\mathrm{FPR}}) \\
    \text{Equal Odds} &= \max(\Delta_{\mathrm{TPR}}, \Delta_{\mathrm{FPR}})
\end{align*}

\subsection{Robustness}\label{sec:robustness}
We evaluate the worst-case adversarial robustness of the utility with a greedy heuristic estimator, inspired by the attack from the authors of \cite{yang2020greedy}. The tokens in the input sequence are iterated over in a random order, and for each, the closest $N = 5$ elements (measured by the \textit{cosine similarity} of their embedding vectors) in their local vocabularies are extracted as substitution candidate. The original tokens in the input then are substituted with the candidate that maximizes the cross-entropy loss of the classification on the input sample. The algorithm is aborted when at most $\rho = 0.3$ ratio of substituted tokens is reached.\\

\noindent We evaluate the drop in classification accuracy, F1-score, Precision and Recall of adversarially perturbed input samples compared to the clean, unperturbed inputs.

\subsection{Implementation Details for Aggregation and trustworthiness index Scoring}\label{sec:aggregationdetails}

The empirical CDF (ECDF) estimation for each metric described in our auditing framework was implemented using Python \texttt{statsmodels} package\footnote{www.statsmodels.org}. We used same set of ECDF mappings (created from same data pool) across all ranked items to ensure consistency. The $\beta$ weights in Eq.~\ref{eq:copulaweighted} were set to uniform values in our experiments, in absence of an a-priori preference over the individual metrics. Table \ref{table:number_metrics_aggregated} gives the total number ($M_{T}$) of aggregated metrics within each trust dimension.  

\noindent In order to determine optimum configuration for a particular $\omega$ combination of TF trustworthiness index we performed steps as follows: (1) Using the development data partition, in each fold we evaluated trustworthiness indexes across all available epochs by pooling ECDF statistics across all such 
folds and epochs of the individual model at hand. (2) For each $\omega$ listed in Table~\ref{Table:weights}, we stored the set of optimum epoch per fold. (3) We picked data from the {\em test} partition of the corresponding epoch of each fold. (3) A union set of test epochs from previous step was formed determining the final set of TF "models," some of which were optimum for multiple $\omega$ configurations (as can be seen in trustworthiness index tables, e.g., Table~\ref{table:recruitment_metrics}). Note that, after the final set of epochs and TF models was determined, the final rankings were based on ECDF estimated from pooled statistics of {\em all} TF configurations as well as baseline models. 
\begin{table}[t!]
\resizebox{1\textwidth}{!} {
\begin{tabular}{r|ccccc}
\toprule
  & Fidelity & Privacy & Utility & Fairness & Robustness \\
\midrule
\multirow{24}{*}{\rotatebox[origin=c]{90}{Metric Names}} & ChiSq\_a\_level & ReplicatedRows & CC\_LR\_accuracy\_<seed> & CC\_LR\_EOD\_<seed> & CC\_MLPFairMixup\_delta\_accuracy \\
 & ChiSq\_gcse & NNRawData\_Median\_<seed> & CC\_LR\_precision\_<seed> & CC\_LR\_Average Odds Difference\_<seed> & CC\_MLPFairMixup\_delta\_f1\_score \\
 & ChiSq\_honours & NNRawData\_Mean\_<seed> & CC\_LR\_recall\_<seed> & CC\_LR\_Equalized Odds\_<seed> & CC\_MLPFairMixup\_delta\_precision \\
 & ChiSq\_income & NNEmbeddings\_Median\_<seed> & CC\_LR\_f1\_score\_<seed> & CC\_MLP\_EOD\_<seed> & CC\_MLPFairMixup\_delta\_recall \\
 & ChiSq\_it\_skills & NNEmbeddings\_Mean\_<seed> & CC\_MLP\_accuracy\_<seed> & CC\_MLP\_Average Odds Difference\_<seed> & CC\_MLPFairMixup\_adv\_accuracy \\
 & ChiSq\_quality\_cv & & CC\_MLP\_precision\_<seed> & CC\_MLP\_Equalized Odds\_<seed> & CC\_MLPFairMixup\_adv\_f1\_score \\
 & ChiSq\_race\_white & & CC\_MLP\_recall\_<seed> & CC\_KNN\_EOD\_<seed> & CC\_MLPFairMixup\_adv\_precision \\
 & ChiSq\_referred & & CC\_MLP\_f1\_score\_<seed> & CC\_KNN\_Average Odds Difference\_<seed> & CC\_MLPFairMixup\_adv\_recall \\
 & ChiSq\_russell\_group & & CC\_KNN\_accuracy\_<seed> & CC\_KNN\_Equalized Odds\_<seed> & CC\_MLP\_delta\_accuracy \\
 & ChiSq\_sex\_male & & CC\_KNN\_precision\_<seed> & CC\_MLPFairMixup\_EOD\_<seed> & CC\_MLP\_delta\_f1\_score \\
 & ChiSq\_years\_experience & & CC\_KNN\_recall\_<seed> & CC\_MLPFairMixup\_Average Odds Difference\_<seed> & CC\_MLP\_delta\_precision \\
 & ChiSq\_years\_gaps & & CC\_KNN\_f1\_score\_<seed> & CC\_MLPFairMixup\_Equalized Odds\_<seed> & CC\_MLP\_delta\_recall \\
 & ChiSq\_years\_volunteer & & CC\_MLPFairMixup\_accuracy\_<seed> & CC\_MLPAdversarial\_EOD\_<seed> & CC\_MLP\_adv\_accuracy \\
 & MutualInformation & & CC\_MLPFairMixup\_precision\_<seed> & CC\_MLPAdversarial\_Average Odds Difference\_<seed> & CC\_MLP\_adv\_f1\_score \\
 & knnPrecisionRecallprecision & & CC\_MLPFairMixup\_recall\_<seed> & CC\_MLPAdversarial\_Equalized Odds\_<seed> & CC\_MLP\_adv\_precision \\
 & knnPrecisionRecallrecall & & CC\_MLPFairMixup\_f1\_score\_<seed> & & CC\_MLP\_adv\_recall \\
 & FID & & CC\_MLPAdversarial\_accuracy\_<seed> & & CC\_MLPAdversarial\_delta\_accuracy \\
 & MMD\_snr & & CC\_MLPAdversarial\_precision\_<seed> & & CC\_MLPAdversarial\_delta\_f1\_score \\
 & MMD\_p\_value & & CC\_MLPAdversarial\_recall\_<seed> & & CC\_MLPAdversarial\_delta\_precision \\
 & & & CC\_MLPAdversarial\_f1\_score\_<seed> & & CC\_MLPAdversarial\_delta\_recall \\
 & & & & & CC\_MLPAdversarial\_adv\_accuracy \\
 & & & & & CC\_MLPAdversarial\_adv\_f1\_score \\
 & & & & & CC\_MLPAdversarial\_adv\_precision \\
 & & & & & CC\_MLPAdversarial\_adv\_recall \\
 \midrule
 \rotatebox[origin=c]{90}{Total Count} & 19 & 13 & 100 & 75 & 24 \\
\bottomrule
\end{tabular}}
\captionof{table}{ Total metrics aggregated within each trust dimensions. <seed> refers to the 5 seeds considered in training MLP classifiers. }
\label{table:number_metrics_aggregated}
\end{table}
\noindent When ranking models with uncertainty (see Eq.~(\ref{eq:mean_var}), we chose the parameter $\alpha=0.1$ ad-hoc considering a value of 0.1 as being a good example of balance between the mean and deviation. The balance itself is application-dependent. Note that in order to keep our tables consistent in terms of models sets between mean-only and mean-variance ranking, we first fixed the final TF model set via epoch selection that used mean-only ranking, and only then applied mean-variance ranking. Optimally, however, same ranking method should be used in both the epoch selection and final comparison. 

\subsection{Independent Copula Versus Gaussian Copula}\label{app:GCversusIC}

\begin{figure}[htp]
\centering
\includegraphics[width=0.9\linewidth]{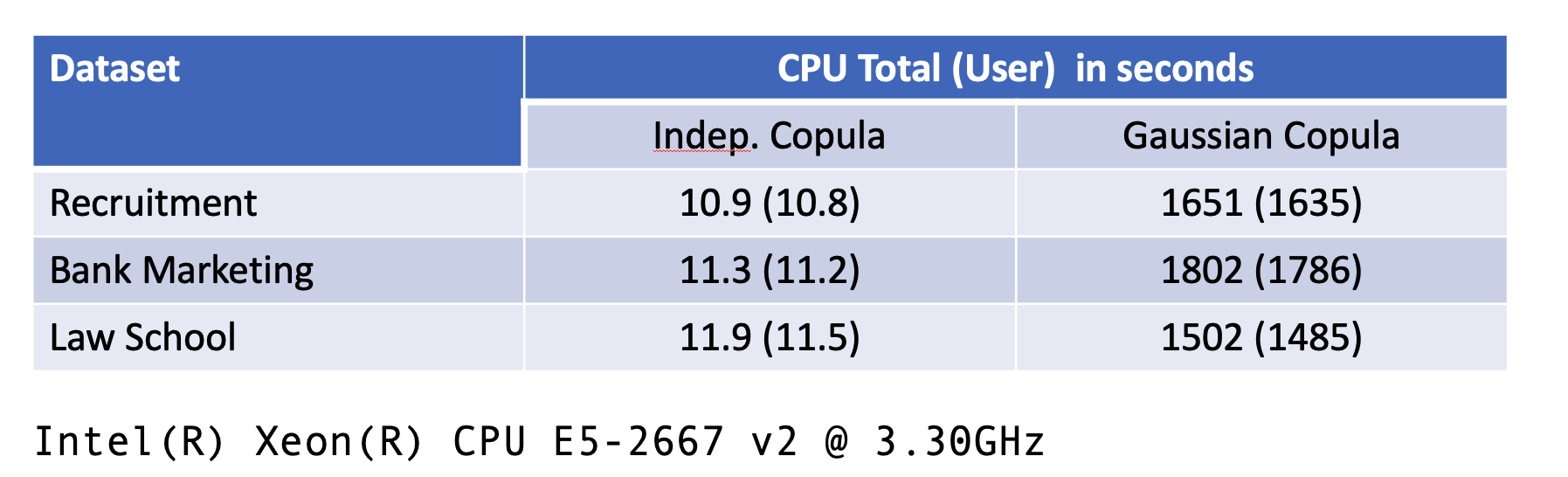} 
\caption{Timing of Independent Copula evaluation versus Gaussian Copula. The Independent copula is 100x faster and more stable numerically and enjoy favorable statistical propreties. }
\label{fig:timingIC_GC}
\end{figure}

\begin{figure}[htp]
\begin{minipage}{\textwidth}
\centering
\includegraphics[width=0.5\linewidth]{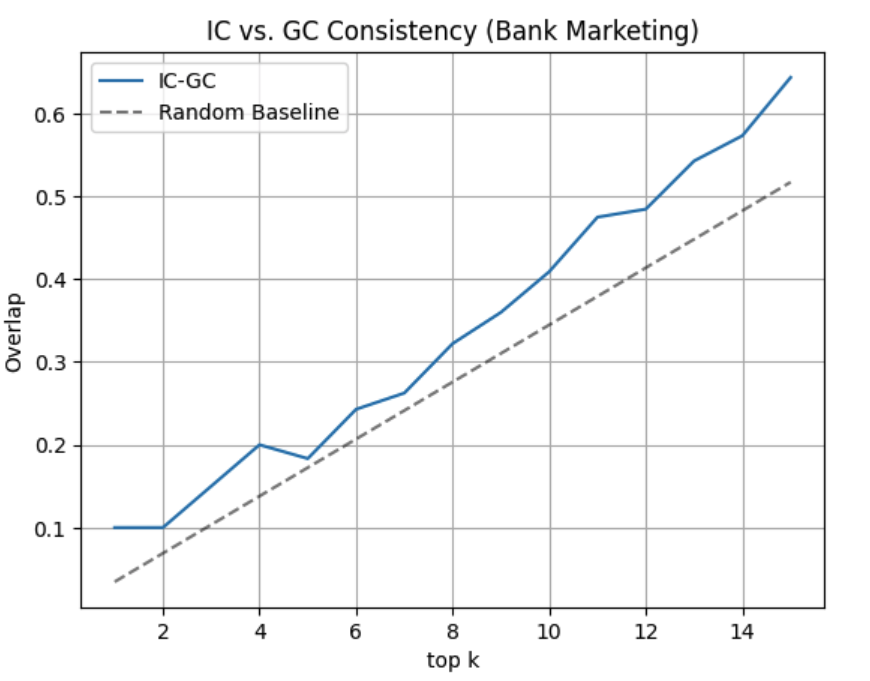} 
\end{minipage}

\begin{minipage}{\textwidth}
\centering
\includegraphics[width=0.5\linewidth]{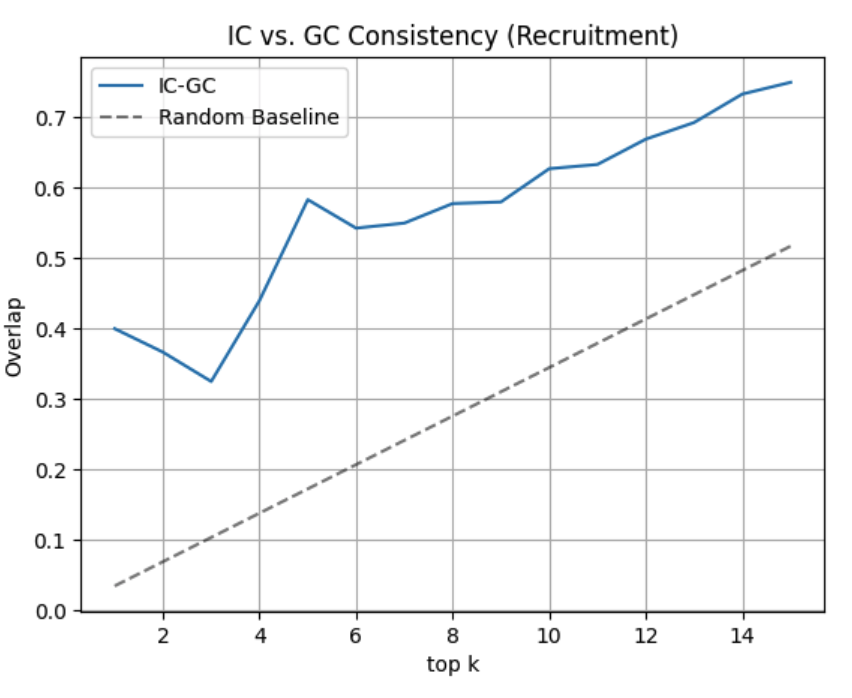}     

\end{minipage}
\begin{minipage}{\textwidth}
\centering
\includegraphics[width=0.5\linewidth]{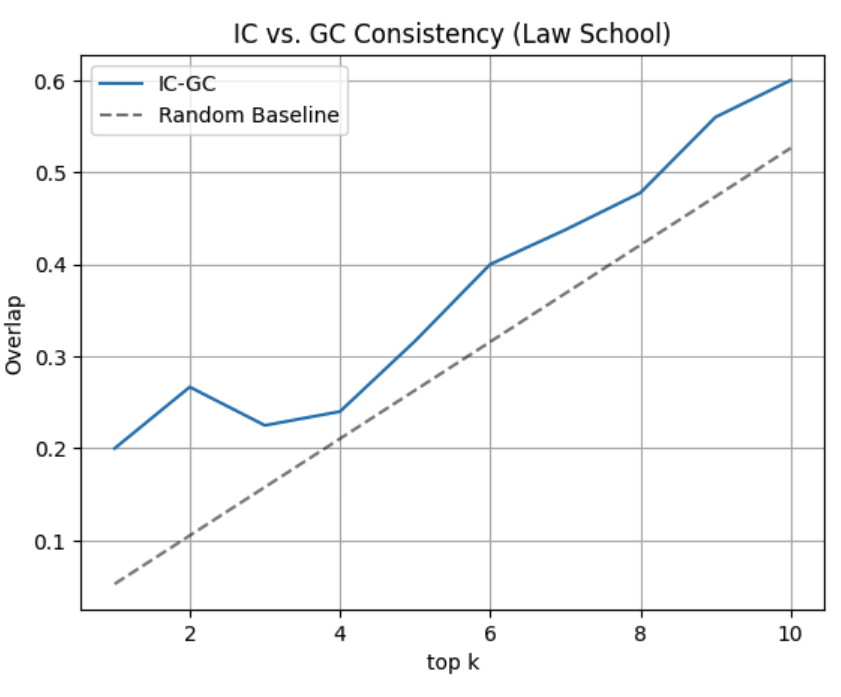}     
\end{minipage}
\caption{ The overlap between ranked list of synthetic data using Independent Copula and Gaussian copula. Dashed line represents random baseline. We see that the top 10 synthetic datasets have between 60-70\% overlap.  }
\label{fig:overlap_IC_GC}
\end{figure}

\subsection{Trustworthiness Index Driven Cross Validation}

\subsection{Example of Auditing Report} \label{sec:AuditReport}
We provide here an example of the audit report on the law school dataset. We provide here only the summary of the results, for full report please refer to the supplementary material.
\newpage


# Synthetic Data Report

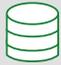
## Real Data

| | |
|---|---|
| **Dataset** | Law School Admission Council Dataset (long) |
| **Download link** | https://github.com/jjgold012/lab-project-fairness/blob/master/fairness_project/datasets/ucla-law-school-dataset/lsac.csv |
| **Number of rows** | 20461 |
| **Number of fields** | 11 |
| **Task for utility** | The goal is to predict wether a candidate passed the bar (variable pass_bar) |
| **Sensitive attribute** | race and gender |

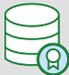
## Synthetic Data

| Name | Cite | Number of rows | Architecture | Commercial API | Opensource code | Differential privacy | Bias mitigation | Selection criteria |
|---|---|---|---|---|---|---|---|---|
| TrustFormer (only_UR, private - eps1) | https://github.com/IBM/TabFormer | 20000 | GPT w/DP-SGD | no | yes | yes, epsilon 1 | no | Utility and Robustness only |
| DP-PGM (private - eps1) | https://github.com/ryan112358/private-pgm | 20000 | Differential Private Probabilistic Graphical Models | no | yes | yes, epsilon 1 | no | n/a |
| SDV-Gaussian-Copula (non-private) | https://sdv.dev/SDV/user_guides/single_table/gaussian_copula.html | 20000 | SDV Gaussian Copula | no | yes | n/a | no | n/a |
| DP-PGM (private - eps3) | https://github.com/ryan112358/private-pgm | 20000 | Differential Private Probabilistic Graphical Models | no | yes | yes, epsilon 3 | no | n/a |
| DP-PGM (target, private - eps1) | https://github.com/ryan112358/private-pgm | 20000 | Differential Private Probabilistic Graphical Models (2-way marginals with target) | no | yes | yes, epsilon 1 | no | n/a |
| DP-PGM (target, private - eps3) | https://github.com/ryan112358/private-pgm | 20000 | Differential Private Probabilistic Graphical Models (2-way marginals with target) | no | yes | yes, epsilon 3 | no | n/a |

# Aggregated Summary Rankings

Synthetic data is evaluated on multiple trust dimensions: fidelity (its closeness to real data ), privacy preservation, utility (accuracy on downstream tasks tested on real data), bias and discrimination and fairness on the downstream task. For each trust dimension, several metrics are evaluated and then aggregated to a score ranging between 0 and 100 % where 100% means perfect compliance for this trust dimension . All trust dimensions are then aggregated to a final score using a geometric mean (for the purpose of this report we use uniform weighting). A final ranking of synthetic data is then obtained using these final scores. Interpretable explanations of potential issues, harms and biases are also given for each trust dimension.

### Rank 1: SDV-Gaussian-Copula (non-private)

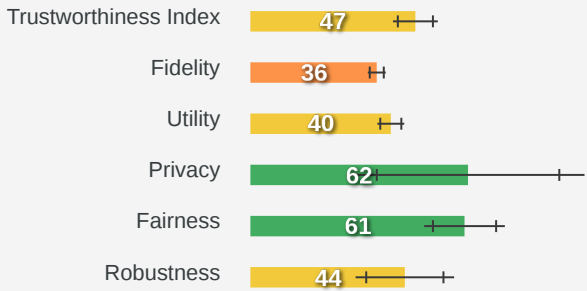

High Fidelity. High Diversity.

⚠ There is 0.82% real data leakage!

Downstream task performance has a F1 score of 0.765. The performance is 80.00% of the performance of a classifier trained on real data.

⚠ Bias Detected: Underprivileged community is disfavored by Equal Opportunity Difference of 0.80 w.r.t Privileged community.

Downstream classifier trained on the synthetic data has an adversarial F1 score of 0.581, with a delta difference from non-adversarial F1 score by -0.376

### Rank 2: TrustFormer (only_UR, private - eps1)

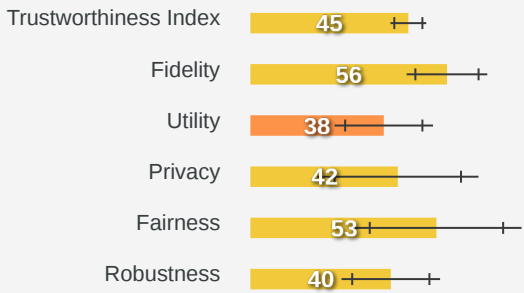

High Fidelity. High Diversity.

⚠ There is 17.07% real data leakage!

Downstream task performance has a F1 score of 0.574. The performance is 60.00% of the performance of a classifier trained on real data.

⚠ Bias Detected: Underprivileged community is disfavored by Equal Opportunity Difference of 0.60 w.r.t Privileged community.

Downstream classifier trained on the synthetic data has an adversarial F1 score of 0.581, with a delta difference from non-adversarial F1 score by -0.376

### Rank 3: DP-PGM (private - eps1)

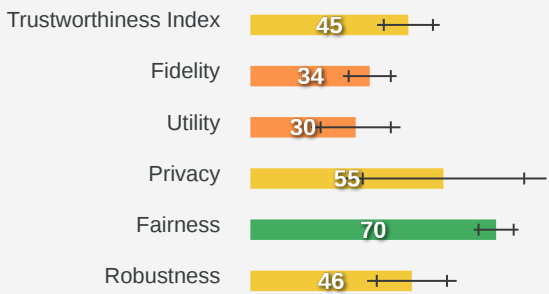

High Fidelity. Low Diversity.

There is no real data leakage!

Downstream task performance has a F1 score of 0.578. The performance is 60.43% of the performance of a classifier trained on real data.

⚠ Bias Detected: Underprivileged community is disfavored by Equal Opportunity Difference of 0.40 w.r.t Privileged community.

Downstream classifier trained on the synthetic data has an adversarial F1 score of 0.581, with a delta difference from non-adversarial F1 score by -0.376

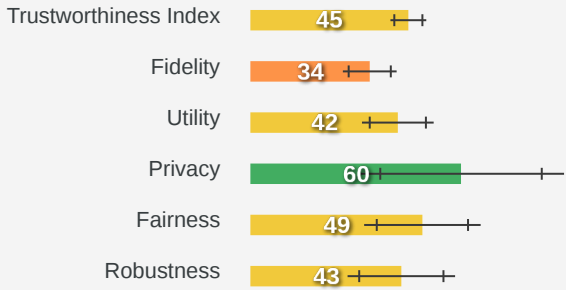

**Rank 4: DP-PGM (target, private - eps1)**

High Fidelity. Low Diversity.

There is no real data leakage!

Downstream task performance has a F1 score of 0.956. The performance is 100.00% of the performance of a classifier trained on real data.

⚠ Bias Detected: Underprivileged community is disfavored by Equal Opportunity Difference of 1.00 w.r.t Privileged community.

Downstream classifier trained on the synthetic data has an adversarial F1 score of 0.581, with a delta difference from non-adversarial F1 score by -0.376

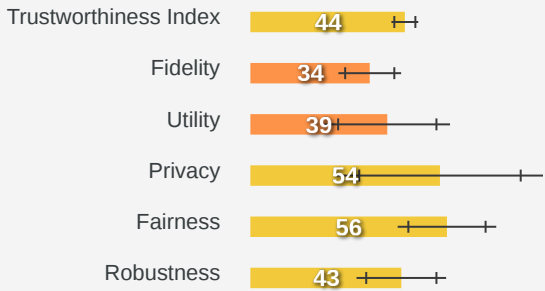

**Rank 5: DP-PGM (private - eps3)**

High Fidelity. Low Diversity.

There is no real data leakage!

Downstream task performance has a F1 score of 0.765. The performance is 80.00% of the performance of a classifier trained on real data.

⚠ Bias Detected: Underprivileged community is disfavored by Equal Opportunity Difference of 0.80 w.r.t Privileged community.

Downstream classifier trained on the synthetic data has an adversarial F1 score of 0.580, with a delta difference from non-adversarial F1 score by -0.376

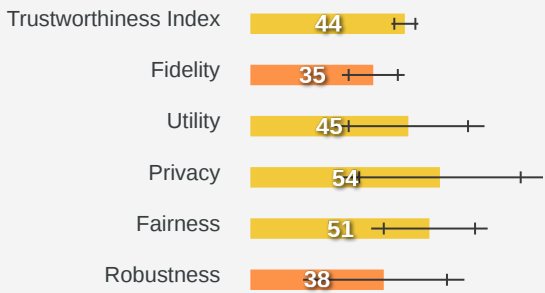

**Rank 6: DP-PGM (target, private - eps3)**

High Fidelity. Low Diversity.

There is no real data leakage!

Downstream task performance has a F1 score of 0.956. The performance is 100.00% of the performance of a classifier trained on real data.

⚠ Bias Detected: Underprivileged community is disfavored by Equal Opportunity Difference of 1.00 w.r.t Privileged community.

Downstream classifier trained on the synthetic data has an adversarial F1 score of 0.581, with a delta difference from non-adversarial F1 score by -0.376



\subsection{Pseudo-code }
\begin{figure}[htp!]
\centering
\includegraphics[scale=0.6]{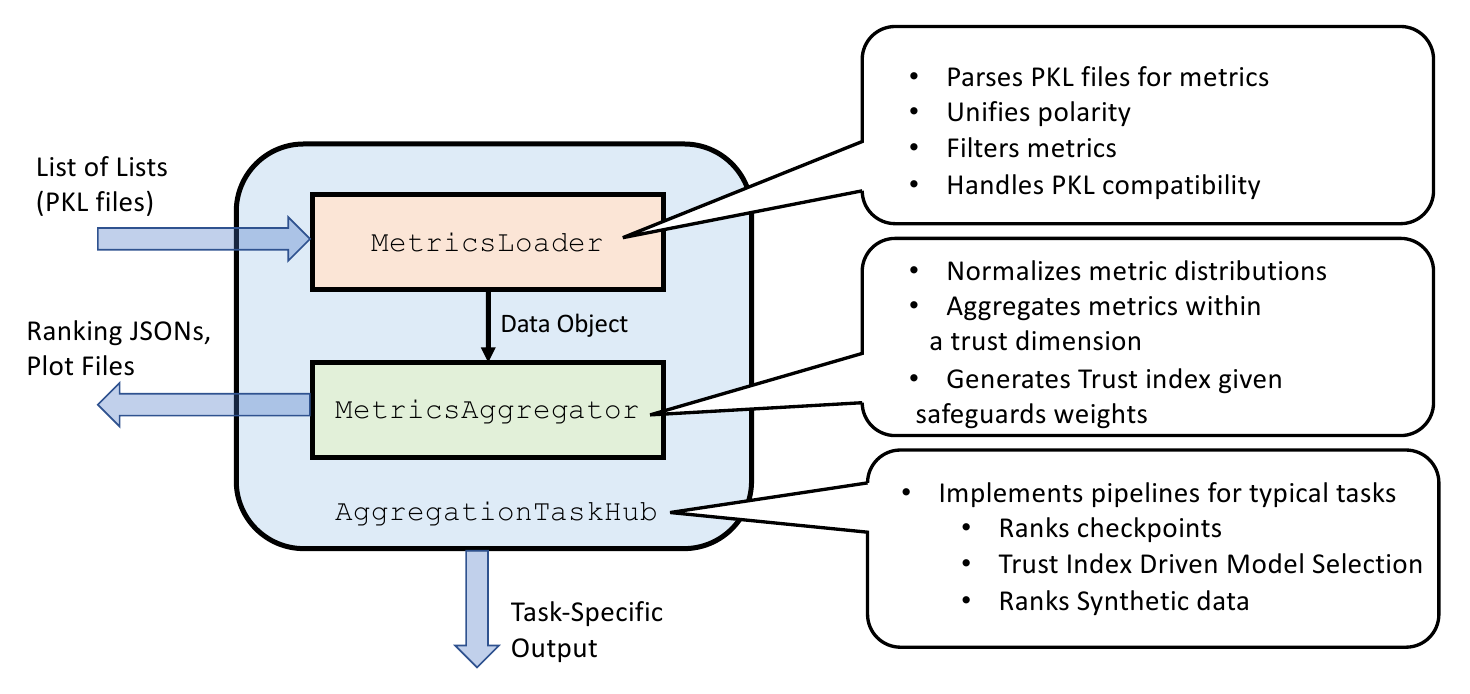} 
\caption{Aggregation Class Structure.}
\label{fig:aggClass}
\end{figure}

\begin{figure}[htp!]
\centering
\includegraphics[scale=0.6]{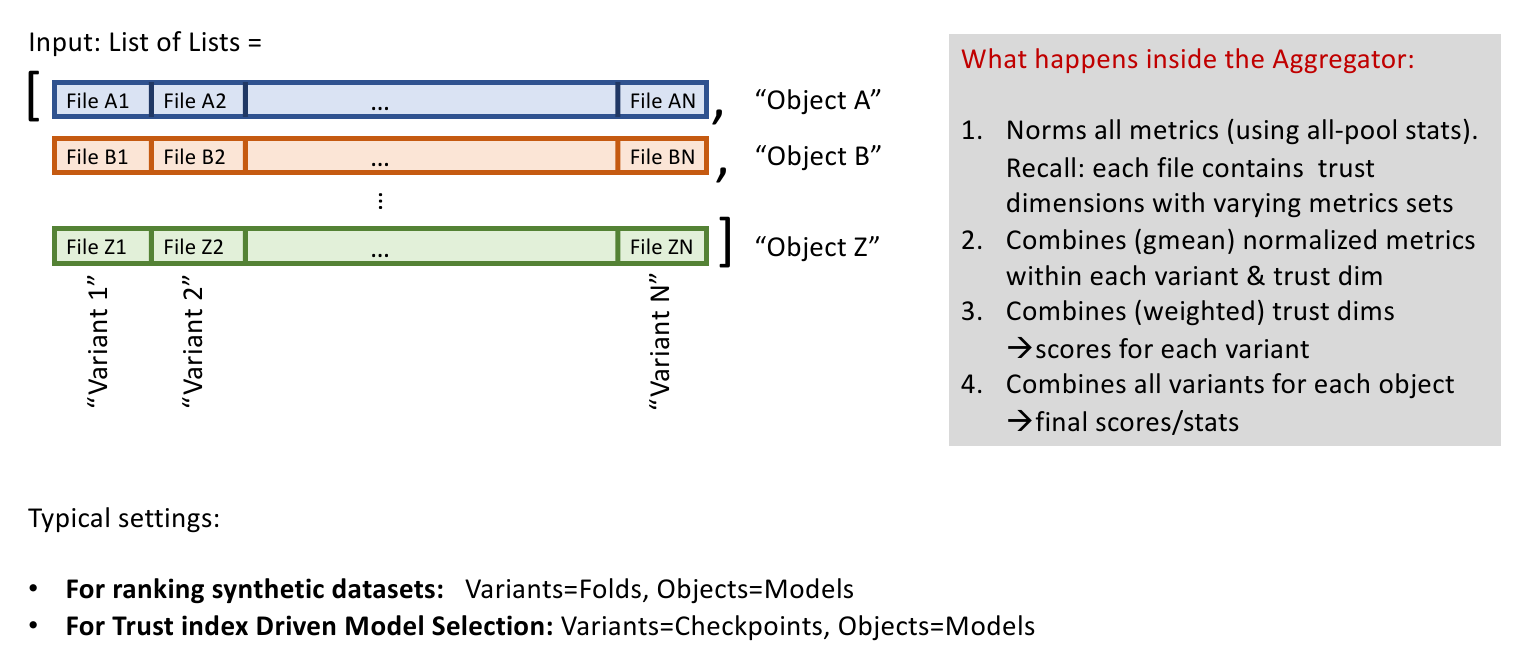} 
\caption{Aggregation inputs and tasks. }
\label{fig:aggInputs}
\end{figure}

\noindent Figure \ref{fig:aggClass} gives the structure of the aggregation class, used to preform metrics aggregation, trustworthiness index computation, trustworthiness index  driven  model selection and synthetic data rankings\footnote{Implementation can be found in file \texttt{metrics\_aggregation.py} in Supplementary Material}. The input to this class is a list of lists containing metrics computed for each trust dimension (see Figure \ref{fig:aggInputs}). The meaning of ``objects" and ``variants" varies depending on use case: 

\noindent (1) For ranking synthetic data from a generator trained on k-folds, each "object" corresponds to the generator's technique/type and each "variant" correspond to synthetic data coming from a generator trained on a corresponding fold. For utility, fairness and robustness, we compute metrics on test portion of the fold. 

\noindent (2) For TrustFormer model selection using trustworthiness index  cross validation, the list of list corresponds to models as "objects" and checkpoints as "variants". In this case we use in utility, fairness and robustness, metrics computed on the {\em validation} portion of the fold to prevent biased selection. 

\subsection{Additional Experimental Results} \label{app:additionalResults}
\subsection{Bank Marketing Dataset}
\begin{figure}[htp]
\centering
\includegraphics[width=0.7\linewidth]{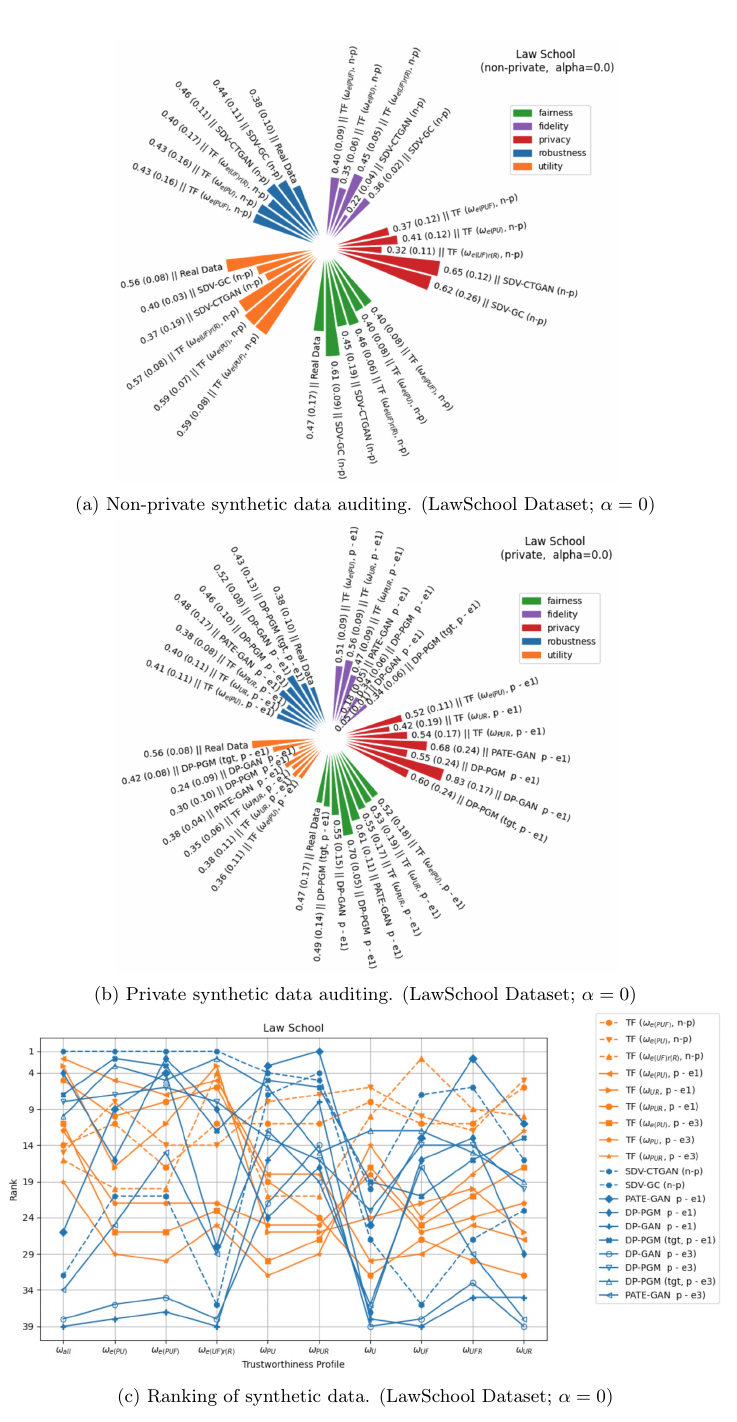} 
\caption{}
\label{Fig:LawSchool_0}
\end{figure}

\begin{figure}[htp]
\centering
\includegraphics[width=0.7\linewidth]{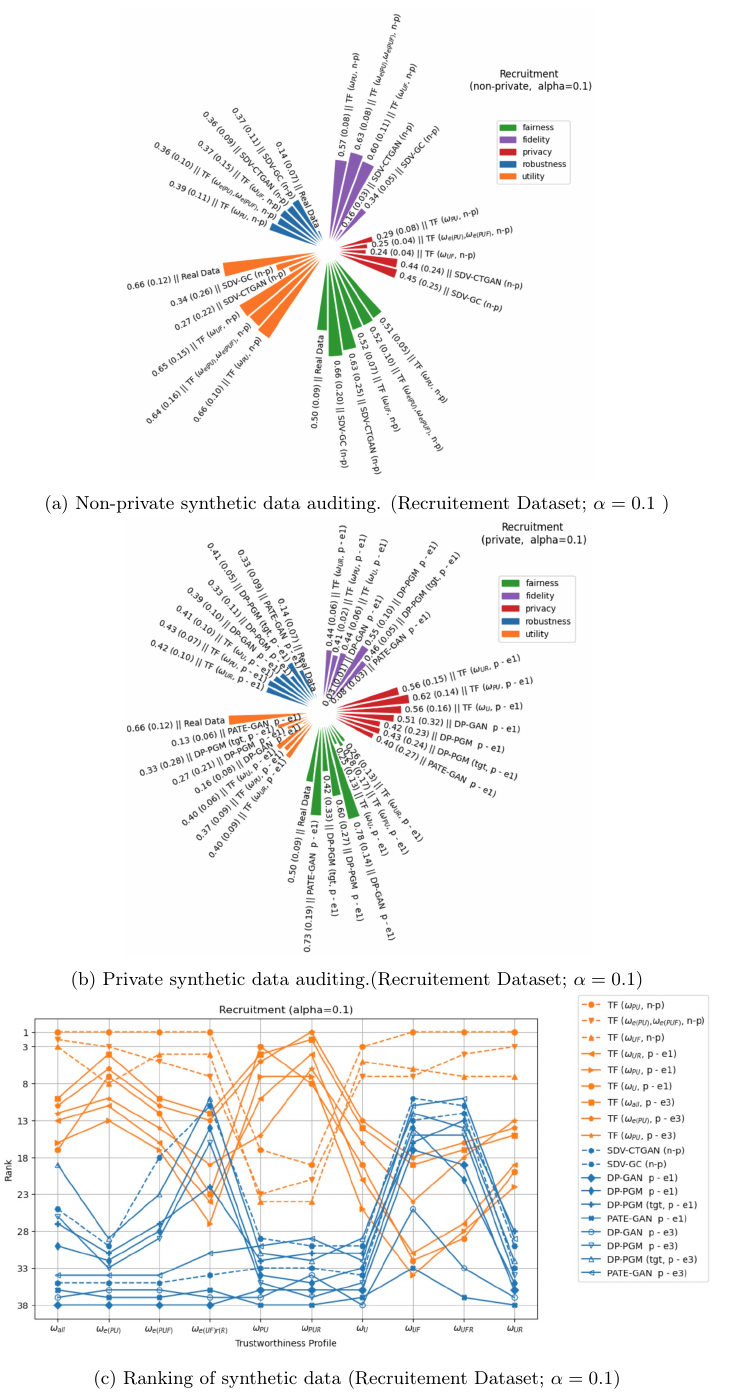} 
\caption{}
\label{Fig:Recruitement_alpha}
\end{figure}

\begin{figure*}[htp]
\centering
\includegraphics[width=0.7\linewidth]{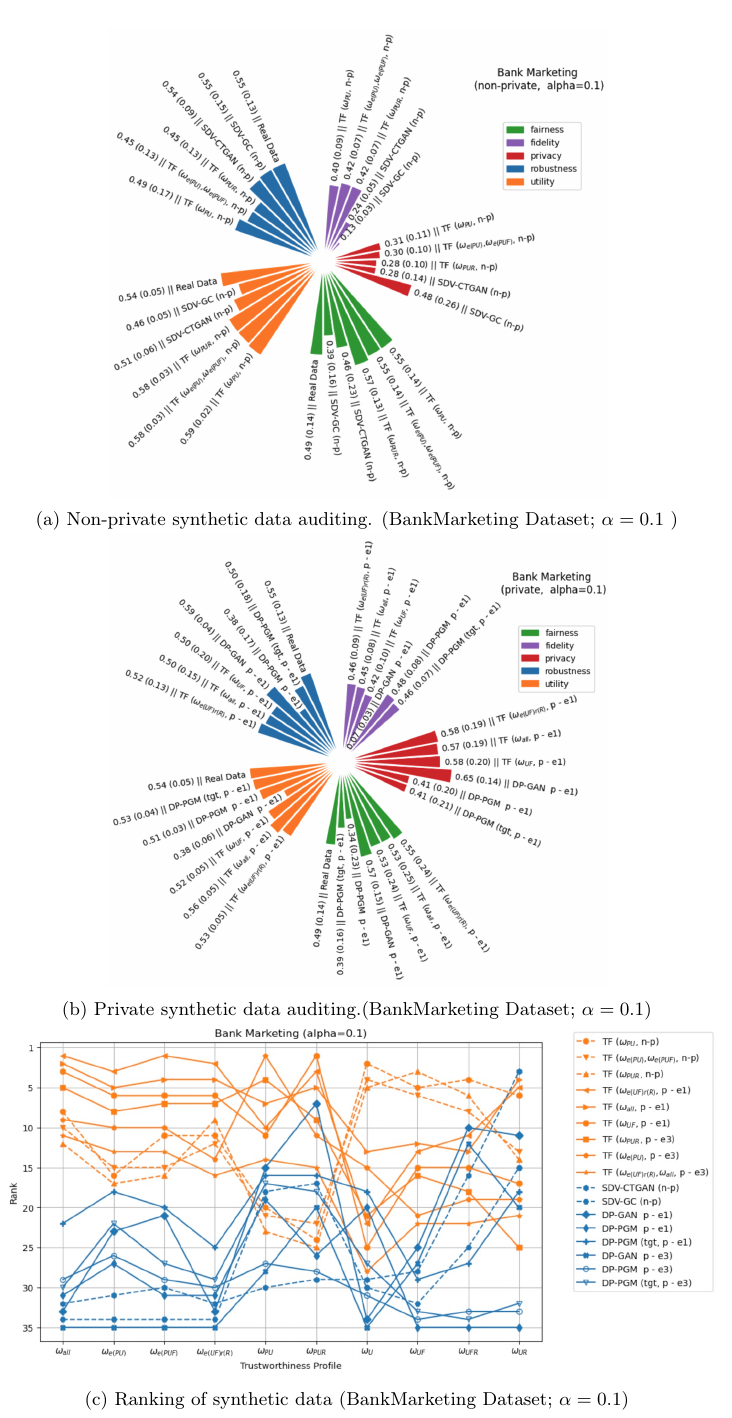} 
\caption{}
\label{Fig:BankMarketing_alpha}
\end{figure*}

\begin{table*}[htp!]
\resizebox{1\textwidth}{!}{
\begin{tabular}{lcccHHcHHcHH}
\toprule
 Model & Fidelity &Privacy &Utility &Utility & Utility & Fairness & Fairness & Fairness & Robustness & Robustness & Robustness\\
\midrule
\textbf{Non-private TrustFormer}&&& & & && &&& \\
TF ($\omega_{PU}$, n-p) & 0.42 (0.09) & \textbf{0.30 (0.11)} &\textcolor{blue}{\textbf{0.59 (0.01)}} & 0.69 (0.03) &0.53 (0.03) & 0.54 (0.14) &0.76 (0.18) & 0.43 (0.15) & 0.49 (0.17) &0.54 (0.22) & 0.40 (0.09) \\
TF ($\omega_{all}$, n-p) & 0.50 (0.07) & 0.24 (0.10) & 0.55 (0.07) & 0.65 (0.04) &0.49 (0.08) & 0.56 (0.14) &0.74 (0.22) & 0.46 (0.15) & 0.49 (0.15) &0.53 (0.22) & 0.41 (0.11) \\
TF ($\omega_{e(PU)},\omega_{e(PUF)}$, n-p) & 0.44 (0.07) & 0.29 (0.10) & 0.58 (0.03) & 0.67 (0.03) &0.53 (0.03) & 0.54 (0.14) &0.70 (0.23) & 0.46 (0.14) & 0.45 (0.13) &0.47 (0.17) & 0.40 (0.09) \\
TF ($\omega_{UR}$, n-p) & 0.45 (0.09) & 0.19 (0.09) & 0.58 (0.03) & 0.68 (0.05) &0.52 (0.05) & 0.56 (0.13) &0.73 (0.23) & 0.46 (0.14) & 0.56 (0.12) &0.65 (0.14) & 0.42 (0.20) \\
TF ($\omega_{UFR}$, n-p) & 0.48 (0.07) & 0.20 (0.12) & \textbf{0.59 (0.03)} & \textcolor{blue}{\textbf{0.70 (0.02)}} &0.52 (0.05) & \textbf{0.58 (0.12)} &0.78 (0.19) & 0.48 (0.14) & 0.49 (0.18) &0.54 (0.23) & 0.42 (0.20) \\
TF ($\omega_{PUR}$, n-p) & 0.44 (0.07) & 0.27 (0.09) & 0.58 (0.03) & 0.69 (0.02) &0.52 (0.05) & 0.56 (0.13) &0.73 (0.20) & 0.47 (0.15) & 0.45 (0.13) &0.47 (0.17) & 0.40 (0.09) \\
TF ($\omega_{U}$, n-p) & 0.47 (0.07) & 0.18 (0.10) & 0.58 (0.02) & 0.65 (0.06) &0.54 (0.04) & 0.55 (0.16) &0.77 (0.19) & 0.44 (0.19) & \textbf{0.59 (0.08)} &0.73 (0.10) & 0.39 (0.20) \\
TF ($\omega_{e(UF)r(R)}$, n-p) &\textbf{0.51 (0.03)} & 0.17 (0.11) & 0.58 (0.04) & 0.68 (0.04) &0.53 (0.06) & 0.48 (0.16) &0.71 (0.22) & 0.37 (0.17) & 0.51 (0.18) &0.61 (0.24) & 0.35 (0.23) \\
TF ($\omega_{UF}$, n-p) & 0.49 (0.05) & 0.18 (0.12) &\textcolor{blue}{\textbf{0.59 (0.01)}} & 0.65 (0.05) &0.55 (0.04) & 0.52 (0.17) &0.71 (0.22) & 0.42 (0.16) & 0.49 (0.17) &0.60 (0.23) & 0.33 (0.23) \\
\midrule 
\textbf{Private TrustFormer}&&& & & && &&& \\
TF ($\omega_{PU}$, p - $\epsilon=1$) & 0.38 (0.07) & \textbf{0.61 (0.22)} & 0.54 (0.03) & 0.60 (0.06) &0.51 (0.06) & 0.46 (0.15) &0.55 (0.32) & 0.41 (0.11) & 0.40 (0.18) &0.36 (0.20) & 0.50 (0.16) \\
TF ($\omega_{PUR}$, p - $\epsilon=1$) & 0.39 (0.08) & 0.60 (0.22) & \textbf{0.56 (0.04)} & \textbf{0.62 (0.06)} &0.51 (0.06) & 0.48 (0.14) &0.55 (0.32) & 0.43 (0.09) & 0.42 (0.17) &0.39 (0.19) & 0.47 (0.17) \\
TF ($\omega_{e(PU)},\omega_{e(PUF)}$, p - $\epsilon=1$) & 0.43 (0.10) & \textbf{0.61 (0.22)} & 0.53 (0.05) & 0.53 (0.09) &0.52 (0.03) & 0.56 (0.23) &0.58 (0.31) & 0.54 (0.20) & 0.44 (0.17) &0.42 (0.17) & 0.49 (0.19) \\
 TF ($\omega_{UF}$, p - $\epsilon=1$) & 0.43 (0.10) & 0.58 (0.20) & 0.53 (0.04) & 0.55 (0.08) &0.51 (0.03) & 0.53 (0.24) &0.59 (0.30) & 0.50 (0.23) & 0.50 (0.20) &0.48 (0.18) & 0.54 (0.24) \\
TF ($\omega_{all}$, p - $\epsilon=1$) & 0.45 (0.08) & 0.57 (0.19) & \textbf{0.56 (0.04)} & 0.60 (0.07) &0.53 (0.03) & 0.52 (0.25) &0.57 (0.31) & 0.50 (0.23) & 0.50 (0.15) &0.50 (0.23) & 0.49 (0.19) \\
TF ($\omega_{e(UF)r(R)}$, p - $\epsilon=1$) & 0.47 (0.09) & 0.57 (0.19) & 0.53 (0.05) & 0.55 (0.09) &0.52 (0.03) & \textbf{0.55 (0.24)} &0.60 (0.30) & 0.52 (0.22) & 0.52 (0.14) &0.54 (0.20) & 0.48 (0.20) \\
TF ($\omega_{UFR}$, p - $\epsilon=1$) & 0.42 (0.10) & 0.57 (0.24) & \textbf{0.56 (0.03)} & \textbf{0.62 (0.05)} &0.52 (0.05) & 0.50 (0.17) &0.55 (0.32) & 0.48 (0.09) & 0.42 (0.17) &0.38 (0.17) & 0.50 (0.18) \\
TF ($\omega_{U}$, p - $\epsilon=1$) & 0.47 (0.10) & 0.54 (0.18) & 0.52 (0.04) & 0.55 (0.06) &0.50 (0.07) & 0.48 (0.23) &0.48 (0.37) & 0.48 (0.16) & 0.39 (0.19) &0.37 (0.20) & 0.42 (0.20) \\
TF ($\omega_{UR}$, p - $\epsilon=1$) & 0.46 (0.08) & 0.50 (0.19) & 0.54 (0.04) & 0.58 (0.07) &0.51 (0.05) & 0.46 (0.20) &0.48 (0.37) & 0.45 (0.12) & 0.40 (0.19) &0.38 (0.17) & 0.43 (0.23) \\
TF ($\omega_{PU}$, p - $\epsilon=3$) & 0.42 (0.08) & {\textbf{0.61 (0.18)}} & 0.53 (0.05) & 0.54 (0.12) &0.52 (0.06) & \textbf{0.55 (0.25)} &0.47 (0.37) & 0.61 (0.18) & 0.45 (0.21) &0.48 (0.20) & 0.39 (0.22) \\
TF ($\omega_{PUR}$, p - $\epsilon=3$) & 0.45 (0.09) & 0.59 (0.19) & 0.53 (0.05) & 0.55 (0.11) &0.52 (0.05) &\textbf{ 0.55 (0.25)} &0.49 (0.36) & 0.59 (0.20) & 0.47 (0.21) &0.47 (0.20) & 0.45 (0.28) \\
TF ($\omega_{UR}$, p - $\epsilon=3$) & 0.50 (0.12) & 0.34 (0.30) & 0.53 (0.04) & 0.53 (0.08) &\textcolor{blue}{\textbf{0.54 (0.05)}} & 0.50 (0.22) &0.52 (0.35) & 0.48 (0.19) & \textbf{0.55 (0.15)} &0.55 (0.15) & 0.55 (0.23) \\
TF ($\omega_{e(PUF)}$, p - $\epsilon=3$) & 0.50 (0.04) & 0.54 (0.12) & 0.49 (0.10) & 0.53 (0.13) &0.47 (0.11) & 0.47 (0.30) &0.48 (0.37) & 0.47 (0.25) & 0.50 (0.18) &0.51 (0.18) & 0.49 (0.25) \\
TF ($\omega_{UFR}$, p - $\epsilon=3$) & 0.51 (0.05) & 0.37 (0.23) & 0.47 (0.08) & 0.46 (0.08) &0.47 (0.10) & 0.45 (0.28) &0.48 (0.37) & 0.44 (0.24) & 0.49 (0.15) &0.48 (0.13) & 0.51 (0.26) \\
TF ($\omega_{e(PU)}$, p - $\epsilon=3$) & 0.53 (0.05) & 0.49 (0.06) & 0.54 (0.04) & 0.57 (0.09) &0.52 (0.05) & 0.50 (0.28) &0.52 (0.35) & 0.48 (0.24) & 0.50 (0.18) &0.48 (0.20) & 0.52 (0.23) \\
TF ($\omega_{e(UF)r(R)},\omega_{all}$, p - $\epsilon=3$) & \textcolor{blue}{\textbf{0.54 (0.05)}} & 0.45 (0.08) & 0.52 (0.06) & 0.53 (0.10) &0.52 (0.06) & 0.50 (0.29) &0.52 (0.35) & 0.49 (0.26) & 0.49 (0.18) &0.47 (0.19) & 0.52 (0.23) \\
TF ($\omega_{U}$, p - $\epsilon=3$) & 0.54 (0.08) & 0.34 (0.19) & 0.55 (0.03) & 0.56 (0.08) &0.54 (0.04) & 0.45 (0.24) &0.51 (0.36) & 0.42 (0.22) & 0.49 (0.17) &0.53 (0.23) & 0.43 (0.23) \\
TF ($\omega_{UF}$, p - $\epsilon=3$) & 0.52 (0.06) & 0.33 (0.24) & 0.52 (0.02) & 0.52 (0.04) &0.53 (0.04) & 0.46 (0.12) &0.51 (0.34) & 0.42 (0.13) & 0.41 (0.20) &0.42 (0.17) & 0.39 (0.32) \\
\midrule 
\midrule 
\textbf{Non-private Baselines}&&& & & && &&& \\
CTGAN (n-p) \cite{SDV} & \textbf{0.26 (0.05)} & 0.27 (0.13) & \textbf{0.52 (0.06)} & 0.62 (0.08) &0.46 (0.06) & \textbf{0.46 (0.23)} &0.70 (0.23) & 0.35 (0.22) & 0.54 (0.08) &0.58 (0.11) & 0.47 (0.06) \\
Gaussian-Copula (n-p)\cite{SDV} & 0.14 (0.03) & \textbf{0.48 (0.26)} & 0.47 (0.05) & 0.49 (0.05) &0.45 (0.06) & 0.39 (0.16) &0.54 (0.34) & 0.32 (0.12) & \textbf{0.55 (0.15)} &0.52 (0.17) & 0.61 (0.18) \\
\midrule 
\midrule
\textbf{Private Baselines}&&& & & && &&& \\
DP-GAN (p - $\epsilon=1$) \cite{qian2023synthcity,chen2020gan}& 0.07 (0.03) & \textcolor{blue}{\textbf{0.65 (0.14)}} & 0.38 (0.06) & 0.43 (0.11) &0.35 (0.04) & 0.57 (0.15) &0.60 (0.30) & 0.55 (0.12) & 0.59 (0.05) &0.61 (0.12) & 0.55 (0.19) \\
DP-GAN (p - $\epsilon=3$) \cite{qian2023synthcity,chen2020gan}& 0.07 (0.03) & 0.44 (0.30) & 0.38 (0.07) & 0.45 (0.16) &0.35 (0.03) & \textcolor{blue}{\textbf{0.60 (0.17)}} &0.67 (0.33) & 0.55 (0.12) & \textcolor{blue}{\textbf{0.61 (0.06)}} &0.61 (0.12) & 0.61 (0.13) \\
DP-PGM (p - $\epsilon=1$)\cite{mckenna2021winning} & 0.48 (0.08) & 0.40 (0.20) & 0.50 (0.03) & 0.53 (0.05) &0.49 (0.02) & 0.34 (0.23) &0.41 (0.40) & 0.30 (0.24) & 0.38 (0.17) &0.39 (0.20) & 0.35 (0.13) \\
DP-PGM (p - $\epsilon=3$)\cite{mckenna2021winning} & 0.50 (0.09) & 0.37 (0.21) & 0.46 (0.05) & 0.48 (0.04) &0.45 (0.09) & 0.38 (0.19) &0.41 (0.40) & 0.35 (0.16) & 0.44 (0.16) &0.46 (0.18) & 0.40 (0.16) \\
DP-PGM (target, p - $\epsilon=1$) \cite{mckenna2021winning}& 0.46 (0.07) & 0.40 (0.21) & \textbf{0.53 (0.04)} & 0.56 (0.07) &0.51 (0.05) & 0.39 (0.16) &0.50 (0.37) & 0.33 (0.10) & 0.50 (0.18) &0.53 (0.23) & 0.44 (0.19) \\
DP-PGM (target, p - $\epsilon=3$)\cite{mckenna2021winning} & \textbf{0.50 (0.08)} & 0.40 (0.21) & 0.52 (0.05) & 0.59 (0.11) &0.47 (0.06) & 0.37 (0.25) &0.44 (0.39) & 0.34 (0.26) & 0.40 (0.19) &0.46 (0.23) & 0.31 (0.15) \\
\midrule 
\midrule 
\textbf{Real Data} & N/A & N/A & 0.54 (0.05) & 0.65 (0.06) &0.48 (0.07) & 0.49 (0.14) &0.63 (0.27) & 0.41 (0.11) & 0.55 (0.12) &0.62 (0.16) & 0.42 (0.08) \\
\bottomrule
\end{tabular}
}
\captionof{table}{Bank Marketing: trust dimension indices. In bold highest index within each group of synthetic data. In blue highest value across all methods including real data.  }
\label{table:bank_marketing_metrics}
\end{table*}

\begin{table*}[htp!]
\resizebox{1\textwidth}{!}{
\begin{tabular}{lHHHccHccHcc}
\toprule
 Model & Fidelity &Privacy &Utility &Utility & Utility & Fairness & Fairness & Fairness & Robustness & Robustness & Robustness\\
 & & & &(Debiased $\cmark$) & (Debiased $\xmark$) & & (Debiased$\cmark$) &(Debiased $\xmark$) &&(Debiased $\cmark$) & (Debiased $\xmark$)\\
\midrule
\textbf{Non-private TrustFormer}&&& & & && &&& \\
TF ($\omega_{PU}$, n-p) & 0.42 (0.09) & \textbf{0.30 (0.11)} &\textcolor{blue}{\textbf{0.59 (0.01)}} & 0.69 (0.03) &0.53 (0.03) & 0.54 (0.14) &0.76 (0.18) & 0.43 (0.15) & 0.49 (0.17) &0.54 (0.22) & 0.40 (0.09) \\
TF ($\omega_{all}$, n-p) & 0.50 (0.07) & 0.24 (0.10) & 0.55 (0.07) & 0.65 (0.04) &0.49 (0.08) & 0.56 (0.14) &0.74 (0.22) & 0.46 (0.15) & 0.49 (0.15) &0.53 (0.22) & 0.41 (0.11) \\
TF ($\omega_{e(PU)},\omega_{e(PUF)}$, n-p) & 0.44 (0.07) & 0.29 (0.10) & 0.58 (0.03) & 0.67 (0.03) &0.53 (0.03) & 0.54 (0.14) &0.70 (0.23) & 0.46 (0.14) & 0.45 (0.13) &0.47 (0.17) & 0.40 (0.09) \\
TF ($\omega_{UR}$, n-p) & 0.45 (0.09) & 0.19 (0.09) & 0.58 (0.03) & 0.68 (0.05) &0.52 (0.05) & 0.56 (0.13) &0.73 (0.23) & 0.46 (0.14) & 0.56 (0.12) &\textcolor{blue}{\textbf{0.65 (0.14)}} & \textbf{0.42 (0.20)} \\
TF ($\omega_{UFR}$, n-p) & 0.48 (0.07) & 0.20 (0.12) & \textbf{0.59 (0.03)} & \textcolor{blue}{\textbf{0.70 (0.02)}} &0.52 (0.05) & \textbf{0.58 (0.12)} &\textcolor{blue}{\textbf{0.78 (0.19)}} & \textbf{0.48 (0.14)} & 0.49 (0.18) &0.54 (0.23) & 0.42 (0.20) \\
TF ($\omega_{PUR}$, n-p) & 0.44 (0.07) & 0.27 (0.09) & 0.58 (0.03) & 0.69 (0.02) &0.52 (0.05) & 0.56 (0.13) &0.73 (0.20) & 0.47 (0.15) & 0.45 (0.13) &0.47 (0.17) & 0.40 (0.09) \\
TF ($\omega_{U}$, n-p) & 0.47 (0.07) & 0.18 (0.10) & 0.58 (0.02) & 0.65 (0.06) &0.54 (0.04) & 0.55 (0.16) &0.77 (0.19) & 0.44 (0.19) & \textbf{0.59 (0.08)} &0.73 (0.10) & 0.39 (0.20) \\
TF ($\omega_{e(UF)r(R)}$, n-p) &\textbf{0.51 (0.03)} & 0.17 (0.11) & 0.58 (0.04) & 0.68 (0.04) &0.53 (0.06) & 0.48 (0.16) &0.71 (0.22) & 0.37 (0.17) & 0.51 (0.18) &0.61 (0.24) & 0.35 (0.23) \\
TF ($\omega_{UF}$, n-p) & 0.49 (0.05) & 0.18 (0.12) &\textcolor{blue}{\textbf{0.59 (0.01)}} & 0.65 (0.05) &\textcolor{blue}{\textbf{0.55 (0.04)}} & 0.52 (0.17) &0.71 (0.22) & 0.42 (0.16) & 0.49 (0.17) &0.60 (0.23) & 0.33 (0.23) \\
\midrule 
\textbf{Private TrustFormer}&&& & & && &&& \\
TF ($\omega_{PU}$, p - $\epsilon=1$) & 0.38 (0.07) & \textbf{0.61 (0.22)} & 0.54 (0.03) & 0.60 (0.06) &0.51 (0.06) & 0.46 (0.15) &0.55 (0.32) & 0.41 (0.11) & 0.40 (0.18) &0.36 (0.20) & 0.50 (0.16) \\
TF ($\omega_{PUR}$, p - $\epsilon=1$) & 0.39 (0.08) & 0.60 (0.22) & \textbf{0.56 (0.04)} & \textbf{0.62 (0.06)} &0.51 (0.06) & 0.48 (0.14) &0.55 (0.32) & 0.43 (0.09) & 0.42 (0.17) &0.39 (0.19) & 0.47 (0.17) \\
TF ($\omega_{e(PU)},\omega_{e(PUF)}$, p - $\epsilon=1$) & 0.43 (0.10) & \textbf{0.61 (0.22)} & 0.53 (0.05) & 0.53 (0.09) &0.52 (0.03) & 0.56 (0.23) &0.58 (0.31) & 0.54 (0.20) & 0.44 (0.17) &0.42 (0.17) & 0.49 (0.19) \\
 TF ($\omega_{UF}$, p - $\epsilon=1$) & 0.43 (0.10) & 0.58 (0.20) & 0.53 (0.04) & 0.55 (0.08) &0.51 (0.03) & 0.53 (0.24) &0.59 (0.30) & 0.50 (0.23) & 0.50 (0.20) &0.48 (0.18) & 0.54 (0.24) \\
TF ($\omega_{all}$, p - $\epsilon=1$) & 0.45 (0.08) & 0.57 (0.19) & \textbf{0.56 (0.04)} & 0.60 (0.07) &0.53 (0.03) & 0.52 (0.25) &0.57 (0.31) & 0.50 (0.23) & 0.50 (0.15) &0.50 (0.23) & 0.49 (0.19) \\
TF ($\omega_{e(UF)r(R)}$, p - $\epsilon=1$) & 0.47 (0.09) & 0.57 (0.19) & 0.53 (0.05) & 0.55 (0.09) &0.52 (0.03) & \textbf{0.55 (0.24)} &\textbf{0.60 (0.30)} & 0.52 (0.22) & 0.52 (0.14) &0.54 (0.20) & 0.48 (0.20) \\
TF ($\omega_{UFR}$, p - $\epsilon=1$) & 0.42 (0.10) & 0.57 (0.24) & \textbf{0.56 (0.03)} & \textbf{0.62 (0.05)} &0.52 (0.05) & 0.50 (0.17) &0.55 (0.32) & 0.48 (0.09) & 0.42 (0.17) &0.38 (0.17) & 0.50 (0.18) \\
TF ($\omega_{U}$, p - $\epsilon=1$) & 0.47 (0.10) & 0.54 (0.18) & 0.52 (0.04) & 0.55 (0.06) &0.50 (0.07) & 0.48 (0.23) &0.48 (0.37) & 0.48 (0.16) & 0.39 (0.19) &0.37 (0.20) & 0.42 (0.20) \\
TF ($\omega_{UR}$, p - $\epsilon=1$) & 0.46 (0.08) & 0.50 (0.19) & 0.54 (0.04) & 0.58 (0.07) &\textbf{0.51 (0.05)} & 0.46 (0.20) &0.48 (0.37) & 0.45 (0.12) & 0.40 (0.19) &0.38 (0.17) & 0.43 (0.23) \\
TF ($\omega_{PU}$, p - $\epsilon=3$) & 0.42 (0.08) & {\textbf{0.61 (0.18)}} & 0.53 (0.05) & 0.54 (0.12) &0.52 (0.06) & \textbf{0.55 (0.25)} &0.47 (0.37) & \textcolor{blue}{\textbf{0.61 (0.18)}}& 0.45 (0.21) &0.48 (0.20) & 0.39 (0.22) \\
TF ($\omega_{PUR}$, p - $\epsilon=3$) & 0.45 (0.09) & 0.59 (0.19) & 0.53 (0.05) & 0.55 (0.11) &0.52 (0.05) &\textbf{ 0.55 (0.25)} &0.49 (0.36) & 0.59 (0.20) & 0.47 (0.21) &0.47 (0.20) & 0.45 (0.28) \\
TF ($\omega_{UR}$, p - $\epsilon=3$) & 0.50 (0.12) & 0.34 (0.30) & 0.53 (0.04) & 0.53 (0.08) &{\textbf{0.54 (0.05)}} & 0.50 (0.22) &0.52 (0.35) & 0.48 (0.19) & \textbf{0.55 (0.15)} &\textbf{0.55 (0.15)} & 0.55 (0.23) \\
TF ($\omega_{e(PUF)}$, p - $\epsilon=3$) & 0.50 (0.04) & 0.54 (0.12) & 0.49 (0.10) & 0.53 (0.13) &0.47 (0.11) & 0.47 (0.30) &0.48 (0.37) & 0.47 (0.25) & 0.50 (0.18) &0.51 (0.18) & 0.49 (0.25) \\
TF ($\omega_{UFR}$, p - $\epsilon=3$) & 0.51 (0.05) & 0.37 (0.23) & 0.47 (0.08) & 0.46 (0.08) &0.47 (0.10) & 0.45 (0.28) &0.48 (0.37) & 0.44 (0.24) & 0.49 (0.15) &0.48 (0.13) & 0.51 (0.26) \\
TF ($\omega_{e(PU)}$, p - $\epsilon=3$) & 0.53 (0.05) & 0.49 (0.06) & \textbf{0.54 (0.04)} & 0.57 (0.09) &0.52 (0.05) & 0.50 (0.28) &0.52 (0.35) & 0.48 (0.24) & 0.50 (0.18) &0.48 (0.20) & \textbf{0.52 (0.23)} \\
TF ($\omega_{e(UF)r(R)},\omega_{all}$, p - $\epsilon=3$) & \textcolor{blue}{\textbf{0.54 (0.05)}} & 0.45 (0.08) & 0.52 (0.06) & 0.53 (0.10) &0.52 (0.06) & 0.50 (0.29) &0.52 (0.35) & 0.49 (0.26) & 0.49 (0.18) &0.47 (0.19) & 0.52 (0.23) \\
TF ($\omega_{U}$, p - $\epsilon=3$) & 0.54 (0.08) & 0.34 (0.19) & 0.55 (0.03) & 0.56 (0.08) &0.54 (0.04) & 0.45 (0.24) &0.51 (0.36) & 0.42 (0.22) & 0.49 (0.17) &0.53 (0.23) & 0.43 (0.23) \\
TF ($\omega_{UF}$, p - $\epsilon=3$) & 0.52 (0.06) & 0.33 (0.24) & 0.52 (0.02) & 0.52 (0.04) &0.53 (0.04) & 0.46 (0.12) &0.51 (0.34) & 0.42 (0.13) & 0.41 (0.20) &0.42 (0.17) & 0.39 (0.32) \\
\midrule 
\midrule 
\textbf{Non-private Baselines}&&& & & && &&& \\
CTGAN (n-p) \cite{SDV} & \textbf{0.26 (0.05)} & 0.27 (0.13) & \textbf{0.52 (0.06)} & \textbf{0.62 (0.08)} &\textbf{0.46 (0.06)} & \textbf{0.46 (0.23)} &\textbf{0.70 (0.23)} & \textbf{0.35 (0.22)} & 0.54 (0.08) &\textbf{0.58 (0.11)} & 0.47 (0.06) \\
Gaussian-Copula (n-p)\cite{SDV} & 0.14 (0.03) & \textbf{0.48 (0.26)} & 0.47 (0.05) & 0.49 (0.05) &0.45 (0.06) & 0.39 (0.16) &0.54 (0.34) & 0.32 (0.12) & \textbf{0.55 (0.15)} &0.52 (0.17) & \textcolor{blue}{\textbf{0.61 (0.18)}} \\
\midrule 
\midrule
\textbf{Private Baselines}&&& & & && &&& \\
DP-GAN (p - $\epsilon=1$) \cite{qian2023synthcity,chen2020gan}& 0.07 (0.03) & \textcolor{blue}{\textbf{0.65 (0.14)}} & 0.38 (0.06) & 0.43 (0.11) &0.35 (0.04) & 0.57 (0.15) &0.60 (0.30) & \textbf{0.55 (0.12)} & 0.59 (0.05) &\textbf{0.61 (0.12)} & 0.55 (0.19) \\
DP-GAN (p - $\epsilon=3$) \cite{qian2023synthcity,chen2020gan}& 0.07 (0.03) & 0.44 (0.30) & 0.38 (0.07) & 0.45 (0.16) &0.35 (0.03) & \textcolor{blue}{\textbf{0.60 (0.17)}} &\textbf{0.67 (0.33)} & 0.55 (0.12) & \textcolor{blue}{\textbf{0.61 (0.06)}} &0.61 (0.12) & \textcolor{blue}{\textbf{0.61 (0.13)}} \\
DP-PGM (p - $\epsilon=1$)\cite{mckenna2021winning} & 0.48 (0.08) & 0.40 (0.20) & 0.50 (0.03) & 0.53 (0.05) &0.49 (0.02) & 0.34 (0.23) &0.41 (0.40) & 0.30 (0.24) & 0.38 (0.17) &0.39 (0.20) & 0.35 (0.13) \\
DP-PGM (p - $\epsilon=3$)\cite{mckenna2021winning} & 0.50 (0.09) & 0.37 (0.21) & 0.46 (0.05) & 0.48 (0.04) &0.45 (0.09) & 0.38 (0.19) &0.41 (0.40) & 0.35 (0.16) & 0.44 (0.16) &0.46 (0.18) & 0.40 (0.16) \\
DP-PGM (target, p - $\epsilon=1$) \cite{mckenna2021winning}& 0.46 (0.07) & 0.40 (0.21) & \textbf{0.53 (0.04)} & 0.56 (0.07) &\textbf{0.51 (0.05)} & 0.39 (0.16) &0.50 (0.37) & 0.33 (0.10) & 0.50 (0.18) &0.53 (0.23) & 0.44 (0.19) \\
DP-PGM (target, p - $\epsilon=3$)\cite{mckenna2021winning} & \textbf{0.50 (0.08)} & 0.40 (0.21) & 0.52 (0.05) & \textbf{0.59 (0.11)} &0.47 (0.06) & 0.37 (0.25) &0.44 (0.39) & 0.34 (0.26) & 0.40 (0.19) &0.46 (0.23) & 0.31 (0.15) \\
\midrule 
\midrule 
\textbf{Real Data} & N/A & N/A & 0.54 (0.05) & 0.65 (0.06) &0.48 (0.07) & 0.49 (0.14) &0.63 (0.27) & 0.41 (0.11) & 0.55 (0.12) &0.62 (0.16) & 0.42 (0.08) \\
\bottomrule
\end{tabular}
}
\captionof{table}{Bank Marketing downstream task evaluation. Study on the effect of debiasing in utility training on trust indices of utility, fairness and robustness.}
\label{table:bank_marketing_metrics_biased_debiased}
\end{table*}

\begin{table*}[htp!]
\resizebox{1\textwidth}{!} {
\begin{tabular}{lrrrrrrrrrr}
\toprule
Model &$\omega_{all}$ &$\omega_{e(PU)}$ &$\omega_{e(PUF)}$ &$\omega_{e(UF)r(R)}$ &$\omega_{PU}$ &$\omega_{PUR}$ &$\omega_{U}$ &$\omega_{UF}$ &$\omega_{UFR}$ &$\omega_{UR}$ \\
\midrule
\textbf{Non-private TrustFormer}&&& & & && &&& \\
TrustFormer ($\omega_{PU}$, n-p) & 18 & 18 & 17 & 16 & 22 & 21 &\rankone{1}& 4 & 4 & 6 \\
 TrustFormer ($\omega_{all}$, n-p) & 20 & 25 & 23 & 19 & 30 & 30 & 12 & 7 & 7 & 12 \\
 TrustFormer ($\omega_{e(PU)}$,$\omega_{e(PUF)}$, n-p) & 21 & 22 & 18 & 17 & 25 & 24 & 5 & 6 & 11 & 15 \\
 TrustFormer ($\omega_{UR}$, n-p) & 26 & 28 & 26 & 23 & 32 & 31 & 8 &\rankthree{3}&\ranktwo{2}&\ranktwo{2}\\
 TrustFormer ($\omega_{UFR}$, n-p) & 24 & 27 & 25 & 21 & 31 & 33 &\rankthree{3}&\rankone{1}&\rankthree{3}& 5 \\
 TrustFormer ($\omega_{PUR}$, n-p) & 22 & 23 & 19 & 15 & 28 & 28 & 7 &\ranktwo{2}& 8 & 16 \\
 TrustFormer ($\omega_{U}$, n-p) & 27 & 29 & 28 & 25 & 34 & 32 & 6 & 5 &\rankone{1}&\rankone{1}\\
 TrustFormer ($\omega_{e(UF)r(R)}$, n-p) & 31 & 34 & 32 & 29 & 35 & 35 & 4 & 15 & 12 &\rankthree{3}\\
 TrustFormer ($\omega_{UF}$, n-p) & 29 & 30 & 31 & 26 & 33 & 34 &\ranktwo{2}& 8 & 6 & 7 \\
\midrule 
\textbf{Private TrustFormer}&&& & & && &&& \\
TrustFormer ($\omega_{PU}$, p - $\epsilon=$1) & 16 & 11 & 12 & 13 &\ranktwo{2}& 10 & 15 & 21 & 28 & 29 \\
 TrustFormer ($\omega_{PUR}$, p - $\epsilon=$1) & 12 & 9 & 9 & 10 &\rankone{1}& 8 & 11 & 19 & 25 & 25 \\
 TrustFormer ($\omega_{e(PU)}$,$\omega_{e(PUF)}$, p - $\epsilon=$1) & 6 & 4 & 4 & 4 & 4 & 7 & 22 & 9 & 18 & 24 \\
 TrustFormer ($\omega_{UF}$, p - $\epsilon=$1) & 5 & 6 & 6 & 7 & 9 &\rankthree{3}& 25 & 16 & 13 & 17 \\
 TrustFormer ($\omega_{all}$, p - $\epsilon=$1) &\ranktwo{2}&\ranktwo{2}&\ranktwo{2}&\rankthree{3}& 5 &\ranktwo{2}& 10 & 10 & 9 & 8 \\
 TrustFormer ($\omega_{e(UF)r(R)}$, p - $\epsilon=$1) &\rankone{1}&\rankone{1}&\rankone{1}&\rankone{1}& 8 &\rankone{1}& 18 & 11 & 5 & 10 \\
 TrustFormer ($\omega_{UFR}$, p - $\epsilon=$1) & 10 & 8 & 8 & 8 & 6 & 11 & 9 & 14 & 23 & 23 \\
 TrustFormer ($\omega_{U}$, p - $\epsilon=$1) & 13 & 13 & 13 & 11 & 10 & 15 & 26 & 23 & 32 & 33 \\
 TrustFormer ($\omega_{UR}$, p - $\epsilon=$1) & 14 & 14 & 14 & 12 & 12 & 16 & 16 & 22 & 31 & 31 \\
 TrustFormer ($\omega_{PU}$, p - $\epsilon=$3) & 7 & 5 & 5 & 5 &\rankthree{3}& 5 & 20 & 13 & 17 & 22 \\
 TrustFormer ($\omega_{PUR}$, p - $\epsilon=$3) &\rankthree{3}&\rankthree{3}&\rankthree{3}&\ranktwo{2}& 7 & 4 & 19 & 12 & 15 & 20 \\
 TrustFormer ($\omega_{UR}$, p - $\epsilon=$3) & 11 & 15 & 15 & 18 & 21 & 19 & 17 & 18 & 10 & 4 \\
 TrustFormer ($\omega_{e(PUF)}$, p - $\epsilon=$3) & 9 & 10 & 10 & 14 & 13 & 12 & 30 & 27 & 24 & 21 \\
 TrustFormer ($\omega_{UFR}$, p - $\epsilon=$3) & 17 & 19 & 21 & 24 & 24 & 22 & 31 & 30 & 27 & 27 \\
 TrustFormer ($\omega_{e(PU)}$, p - $\epsilon=$3) & 4 & 7 & 7 & 6 & 11 & 9 & 14 & 17 & 16 & 13 \\
TrustFormer ($\omega_{e(UF)r(R)}$,$\omega_{all}$, p - $\epsilon=$3) & 8 & 12 & 11 & 9 & 15 & 14 & 24 & 20 & 20 & 18 \\
 TrustFormer ($\omega_{U}$, p - $\epsilon=$3) & 15 & 16 & 16 & 20 & 20 & 20 & 13 & 24 & 22 & 11 \\
 TrustFormer ($\omega_{UF}$, p - $\epsilon=$3) & 23 & 21 & 22 & 22 & 23 & 29 & 23 & 25 & 29 & 30 \\

\midrule 
\midrule 
\textbf{Non-Private Baselines}&&& & & && &&& \\
CTGAN (n-p)\cite{SDV} & 32 & 33 & 33 & 32 & 29 & 25 & 28 & 26 & 19 & 9 \\
Gaussian-Copula (n-p)\cite{SDV} & 33 & 31 & 34 & 33 & 16 & 13 & 32 & 33 & 30 & 19 \\
\midrule 
\textbf{Private Baselines}&&& & & && &&& \\
DP-GAN (p - $\epsilon=$1)\cite{chen2020gan,qian2023synthcity} & 34 & 32 & 30 & 34 & 14 & 6 & 35 & 29 & 21 & 28 \\
 DP-GAN (p - $\epsilon=$3)\cite{chen2020gan,qian2023synthcity} & 35 & 35 & 35 & 35 & 27 & 18 & 34 & 28 & 14 & 26 \\
 DP-PGM (p - $\epsilon=$1)\cite{mckenna2021winning} & 30 & 24 & 29 & 30 & 19 & 26 & 29 & 34 & 35 & 35 \\
 DP-PGM (p - $\epsilon=$3) \cite{mckenna2021winning} & 28 & 26 & 27 & 31 & 26 & 27 & 33 & 35 & 34 & 34 \\
 DP-PGM (target, p - $\epsilon=$1)\cite{mckenna2021winning}  & 19 & 17 & 20 & 27 & 17 & 17 & 21 & 31 & 26 & 14 \\
 DP-PGM (target, p - $\epsilon=$3)\cite{mckenna2021winning}  & 25 & 20 & 24 & 28 & 18 & 23 & 27 & 32 & 33 & 32 \\

\bottomrule
\end{tabular}
}
\captionof{table}{Bank Marketing dataset ranking. We see that overall TrustFormers that use trustworthiness index driven selection corresponding to the desired trade-offs outperform other synthetic data, across these desired trade-offs defined in Table \ref{Table:weights}. Note that the ranking here is on the mean trustworthiness index, where downstream tasks are evaluated on the test data in each real data fold. }
\label{table:bank_marketing_ranking}
\end{table*}

\begin{table*}[htp!]
\resizebox{1\textwidth}{!} {
\begin{tabular}{lrrrrrrrrrr}
\toprule
Model &$\omega_{all}$ &$\omega_{e(PU)}$ &$\omega_{e(PUF)}$ &$\omega_{e(UF)r(R)}$ &$\omega_{PU}$ &$\omega_{PUR}$ &$\omega_{U}$ &$\omega_{UF}$ &$\omega_{UFR}$ &$\omega_{UR}$ \\
\midrule
\textbf{Non-private TrustFormer}&&& & & && &&& \\
TrustFormer ($\omega_{PU}$, n-p) & 8 & 16 & 11 & 11 & 20 & 24 &\ranktwo{2}& 5 & 4 & 6 \\
 TrustFormer ($\omega_{all}$, n-p) & 14 & 19 & 17 & 13 & 29 & 21 & 19 &\ranktwo{2}& 5 & 12 \\
 TrustFormer ($\omega_{e(PU)}$,$\omega_{e(PUF)}$, n-p) & 10 & 15 & 15 & 12 & 21 & 22 & 4 & 6 & 8 & 13 \\
 TrustFormer ($\omega_{UR}$, n-p) & 20 & 29 & 25 & 21 & 32 & 31 & 7 & 4 &\ranktwo{2}&\ranktwo{2}\\
 TrustFormer ($\omega_{UFR}$, n-p) & 17 & 28 & 24 & 18 & 31 & 32 & 6 &\rankone{1}&\rankthree{3}& 10 \\
 TrustFormer ($\omega_{PUR}$, n-p) & 12 & 17 & 16 & 9 & 23 & 25 & 5 &\rankthree{3}& 6 & 14 \\
 TrustFormer ($\omega_{U}$, n-p) & 25 & 30 & 28 & 24 & 34 & 34 &\rankthree{3}& 7 &\rankone{1}&\rankone{1}\\
 TrustFormer ($\omega_{e(UF)r(R)}$, n-p) & 26 & 32 & 33 & 27 & 35 & 33 & 8 & 9 & 9 & 9 \\
 TrustFormer ($\omega_{UF}$, n-p) & 28 & 33 & 32 & 26 & 33 & 35 &\rankone{1}& 8 & 7 & 8 \\

\midrule 
\textbf{Private TrustFormer}&&& & & && &&& \\
TrustFormer ($\omega_{PU}$, p - $\epsilon=$1) & 15 & 7 & 8 & 8 & 6 & 8 & 14 & 19 & 28 & 28 \\
 TrustFormer ($\omega_{PUR}$, p - $\epsilon=$1) & 6 &\ranktwo{2}&\ranktwo{2}& 5 & 5 & 4 & 11 & 14 & 24 & 24 \\
 TrustFormer ($\omega_{e(PU)}$,$\omega_{e(PUF)}$, p - $\epsilon=$1) & 4 & 4 & 5 &\rankthree{3}& 8 &\ranktwo{2}& 23 & 10 & 17 & 23 \\
 TrustFormer ($\omega_{UF}$, p - $\epsilon=$1) &\rankthree{3}& 6 & 6 & 6 & 11 &\rankone{1}& 25 & 15 & 15 & 17 \\
 TrustFormer ($\omega_{all}$, p - $\epsilon=$1) &\ranktwo{2}& 5 & 4 & 4 & 7 & 5 & 13 & 12 & 13 & 4 \\
 TrustFormer ($\omega_{e(UF)r(R)}$, p - $\epsilon=$1) &\rankone{1}&\rankthree{3}&\rankone{1}&\ranktwo{2}& 10 &\rankthree{3}& 22 & 13 & 11 & 5 \\
 TrustFormer ($\omega_{UFR}$, p - $\epsilon=$1) & 7 &\rankone{1}&\rankthree{3}&\rankone{1}& 9 & 6 & 9 & 11 & 20 & 22 \\
 TrustFormer ($\omega_{U}$, p - $\epsilon=$1) & 21 & 11 & 12 & 15 & 12 & 12 & 26 & 24 & 32 & 34 \\
 TrustFormer ($\omega_{UR}$, p - $\epsilon=$1) & 23 & 12 & 14 & 17 & 13 & 13 & 16 & 23 & 30 & 30 \\
 TrustFormer ($\omega_{PU}$, p - $\epsilon=$3) & 13 & 9 & 9 & 10 &\rankthree{3}& 10 & 24 & 17 & 21 & 27 \\
 TrustFormer ($\omega_{PUR}$, p - $\epsilon=$3) & 5 & 8 & 7 & 7 & 4 & 9 & 21 & 16 & 18 & 25 \\
 TrustFormer ($\omega_{UR}$, p - $\epsilon=$3) & 16 & 20 & 19 & 20 & 25 & 19 & 17 & 20 & 14 & 7 \\
 TrustFormer ($\omega_{e(PUF)}$, p - $\epsilon=$3) & 18 & 14 & 18 & 23 &\ranktwo{2}& 14 & 32 & 30 & 29 & 26 \\
 TrustFormer ($\omega_{UFR}$, p - $\epsilon=$3) & 27 & 25 & 26 & 28 & 24 & 27 & 33 & 31 & 31 & 29 \\
 TrustFormer ($\omega_{e(PU)}$, p - $\epsilon=$3) & 9 & 10 & 10 & 14 &\rankone{1}& 11 & 15 & 21 & 19 & 19 \\
TrustFormer ($\omega_{e(UF)r(R)}$,$\omega_{all}$, p - $\epsilon=$3) & 11 & 13 & 13 & 16 & 14 & 15 & 28 & 22 & 22 & 21 \\
 TrustFormer ($\omega_{U}$, p - $\epsilon=$3) & 24 & 21 & 23 & 22 & 22 & 23 & 10 & 26 & 23 & 16 \\
 TrustFormer ($\omega_{UF}$, p - $\epsilon=$3) & 19 & 24 & 22 & 19 & 26 & 30 & 12 & 18 & 26 & 31 \\
 
\midrule 
\midrule 
\textbf{Non-Private Baselines}&&& & & && &&& \\
CTGAN (n-p)\cite{SDV} & 32 & 31 & 30 & 32 & 30 & 29 & 29 & 28 & 16 &\rankthree{3}\\
Gaussian-Copula (n-p)\cite{SDV } & 34 & 34 & 34 & 34 & 18 & 17 & 30 & 32 & 25 & 15 \\
 
\midrule 
\textbf{Private Baselines}&&& & & && &&& \\
 DP-GAN (p - $\epsilon=$1)\cite{chen2020gan,qian2023synthcity} & 33 & 23 & 21 & 33 & 15 & 7 & 34 & 25 & 10 & 11 \\
 DP-GAN (p - $\epsilon=$3)\cite{chen2020gan,qian2023synthcity} & 35 & 35 & 35 & 35 & 28 & 20 & 35 & 27 & 12 & 20 \\
 DP-PGM (p - $\epsilon=$1) \cite{mckenna2021winning} & 31 & 27 & 31 & 31 & 19 & 26 & 20 & 35 & 35 & 35 \\
 DP-PGM (p - $\epsilon=$3)\cite{mckenna2021winning}  & 29 & 26 & 29 & 30 & 27 & 28 & 31 & 34 & 33 & 33 \\
 DP-PGM (target, p - $\epsilon=$1)\cite{mckenna2021winning}  & 22 & 18 & 20 & 25 & 16 & 16 & 18 & 29 & 27 & 18 \\
 DP-PGM (target, p - $\epsilon=$3) \cite{mckenna2021winning} & 30 & 22 & 27 & 29 & 17 & 18 & 27 & 33 & 34 & 32 \\
 
\bottomrule
\end{tabular}
}
\captionof{table}{Bank Marketing dataset. Ranking  with uncertainty of all considered synthetic data using $R^{\alpha}_{\tau}$ defined in Equation \eqref{eq:mean_var} for $\alpha=0.1$. We see that TrustFormer synthetic data is still aligned with the prescribed safeguards while having low volatility in its trustworthiness index.   }
\label{table:bank_marketing_ranking_uncertainty}
\end{table*}

\subsection{Recruitment Dataset}

\begin{table*}[htp!]
\resizebox{1\textwidth}{!}{
\begin{tabular}{lcccHHcHHcHH}
\toprule
 Model & Fidelity &Privacy &Utility &Utility & Utility & Fairness & Fairness & Fairness & Robustness & Robustness & Robustness\\

\midrule
\textbf{Non-private TrustFormer}&&& & & && &&& \\
 TF ($\omega_{PU}$, n-p) & 0.57 (0.08) & \textbf{0.29 (0.08)} & 0.66 (0.10) & 0.54 (0.07) &0.75 (0.15) & 0.51 (0.05) &0.37 (0.08) & 0.64 (0.07) &\textbf{0.39 (0.11)} &0.40 (0.11) & 0.37 (0.14) \\
TF ($\omega_{UR}$, n-p) & 0.54 (0.08) & 0.24 (0.04) & 0.67 (0.10) & 0.56 (0.07) &0.75 (0.15) & 0.51 (0.07) &0.41 (0.11) & 0.59 (0.06) & 0.34 (0.14) &0.35 (0.15) & 0.33 (0.17) \\
TF ($\omega_{PUR}$, n-p) & 0.54 (0.08) & 0.25 (0.05) & \textcolor{blue}{\textbf{0.69 (0.14)}} & 0.60 (0.12) &0.76 (0.16) & 0.51 (0.08) &0.39 (0.09) & 0.61 (0.13) & 0.35 (0.13) &0.36 (0.12) & 0.33 (0.17) \\
TF ($\omega_{e(PU)},\omega_{e(PUF)}$, n-p) & 0.63 (0.08) & 0.25 (0.04) & 0.64 (0.16) & 0.50 (0.17) &0.75 (0.16) & \textbf{0.52 (0.10)} &0.39 (0.09) & 0.64 (0.13) & 0.36 (0.10) &0.37 (0.12) & 0.35 (0.10) \\
 TF ($\omega_{UF}$, n-p) & 0.60 (0.11) & 0.24 (0.04) & 0.65 (0.15) & 0.52 (0.17) &0.76 (0.17) & \textbf{0.52 (0.07)} &0.36 (0.07) & 0.66 (0.12) & 0.37 (0.15) &0.39 (0.13) & 0.33 (0.19) \\
 TF ($\omega_{e(UF)r(R)}$, n-p) & \textcolor{blue}{\textbf{0.65 (0.09)}} & 0.23 (0.04) & 0.61 (0.18) & 0.45 (0.19) &0.75 (0.17) & 0.50 (0.09) &0.35 (0.09) & 0.63 (0.13) & 0.36 (0.13) &0.38 (0.13) & 0.32 (0.16) \\
 TF ($\omega_{UFR}$, n-p) & 0.59 (0.10) & 0.23 (0.05) & 0.66 (0.15) & 0.54 (0.16) &0.75 (0.16) & 0.52 (0.07) &0.40 (0.07) & 0.62 (0.13) & 0.38 (0.14) &0.41 (0.12) & 0.34 (0.18) \\
 TF ($\omega_{all}$, n-p) & 0.64 (0.09) & 0.22 (0.05) & 0.62 (0.18) & 0.47 (0.19) &0.74 (0.17) & 0.50 (0.09) &0.39 (0.09) & 0.59 (0.14) & 0.37 (0.12) &0.40 (0.11) & 0.32 (0.16) \\
 TF ($\omega_{U}$, n-p) & 0.64 (0.09) & 0.24 (0.04) & 0.65 (0.14) & 0.52 (0.15) &0.75 (0.16) & 0.52 (0.07) &0.37 (0.08) & 0.65 (0.10) & \textbf{0.39 (0.13)} &0.41 (0.11) & 0.35 (0.18) \\
\midrule 
\textbf{Private TrustFormer}&&& & & && &&& \\
TF ($\omega_{PU}$, p - $\epsilon=$1) & 0.41 (0.02) & \textcolor{blue}{\textbf{0.62 (0.14)}} & 0.37 (0.09) & 0.43 (0.10) &0.33 (0.10) & 0.28 (0.17) &0.34 (0.12) & 0.25 (0.21) & 0.43 (0.07) &0.41 (0.08) & 0.47 (0.15) \\
TF ($\omega_{UF}$, p - $\epsilon=$1) & 0.36 (0.06) & 0.60 (0.16) & 0.33 (0.07) & 0.43 (0.08) &0.28 (0.07) &\textbf{0.36 (0.19)} &0.43 (0.16) & 0.32 (0.21) & 0.33 (0.09) &0.31 (0.14) & 0.39 (0.06) \\
 TF ($\omega_{all}$, p - $\epsilon=$1) & 0.43 (0.07) & 0.60 (0.15) & 0.35 (0.07) & 0.42 (0.09) &0.31 (0.07) & 0.29 (0.12) &0.40 (0.09) & 0.24 (0.16) & 0.39 (0.09) &0.37 (0.12) & 0.45 (0.17) \\
 TF ($\omega_{UFR}$, p - $\epsilon=$1) & 0.41 (0.09) & 0.60 (0.15) & 0.38 (0.05) & 0.44 (0.07) &0.34 (0.04) & 0.31 (0.11) &0.43 (0.08) & 0.25 (0.15) & 0.40 (0.09) &0.37 (0.13) & 0.44 (0.17) \\
 TF ($\omega_{e(PUF)}$, p - $\epsilon=$1) & 0.39 (0.03) & 0.61 (0.14) & 0.38 (0.07) & 0.46 (0.09) &0.34 (0.06) & 0.32 (0.16) &0.41 (0.11) & 0.28 (0.19) & 0.39 (0.11) &0.34 (0.11) & 0.52 (0.15) \\
TF ($\omega_{e(PU)},\omega_{PUR}$, p - $\epsilon=$1) & 0.42 (0.02) & 0.61 (0.12) & 0.36 (0.08) & 0.43 (0.10) &0.31 (0.07) & 0.27 (0.18) &0.37 (0.10) & 0.22 (0.22) & 0.40 (0.10) &0.35 (0.11) & 0.52 (0.12) \\
TF ($\omega_{U}$, p - $\epsilon=$1) & 0.44 (0.06) & 0.56 (0.16) & 0.40 (0.06) & 0.46 (0.09) &0.36 (0.06) & 0.25 (0.13) &0.35 (0.13) & 0.20 (0.13) & 0.41 (0.10) &0.35 (0.10) & 0.59 (0.12) \\
 TF ($\omega_{UR}$, p - $\epsilon=$1) & 0.44 (0.06) & 0.56 (0.15) & 0.40 (0.09) & 0.44 (0.10) &0.37 (0.11) & 0.26 (0.13) &0.37 (0.13) & 0.21 (0.13) & 0.42 (0.10) &0.36 (0.12) & 0.58 (0.10) \\
TF ($\omega_{e(UF)r(R)}$, p - $\epsilon=$1) & 0.39 (0.03) & 0.59 (0.11) & 0.37 (0.06) & 0.43 (0.08) &0.33 (0.06) & 0.31 (0.16) &0.36 (0.11) & 0.29 (0.19) & 0.38 (0.10) &0.34 (0.10) & 0.50 (0.12) \\
TF ($\omega_{e(PUF)}$, p - $\epsilon=$3) & 0.47 (0.07) & 0.53 (0.12) & 0.40 (0.13) & 0.43 (0.15) &0.38 (0.12) & 0.30 (0.12) &0.35 (0.10) & 0.27 (0.14) & 0.42 (0.06) &0.49 (0.05) & 0.31 (0.16) \\
TF ($\omega_{all}$, p - $\epsilon=$3) & 0.50 (0.07) & 0.52 (0.13) & \textbf{0.45 (0.11)} & 0.48 (0.10) &0.43 (0.13) & 0.29 (0.10) &0.35 (0.11) & 0.25 (0.13) & 0.45 (0.05) &0.48 (0.05) & 0.39 (0.16) \\
TF ($\omega_{PU}$, p - $\epsilon=$3) & 0.50 (0.09) & 0.48 (0.13) & 0.43 (0.10) & 0.46 (0.06) &0.41 (0.12) & 0.26 (0.09) &0.30 (0.08) & 0.24 (0.11) &\textcolor{blue}{\textbf{ 0.51 (0.07)}} &0.49 (0.06) & 0.55 (0.19) \\
TF ($\omega_{e(PU)}$, p - $\epsilon=$3) & 0.49 (0.08) & 0.51 (0.11) & 0.45 (0.11) & 0.48 (0.09) &0.43 (0.12) & 0.29 (0.10) &0.36 (0.10) & 0.25 (0.13) & 0.46 (0.03) &0.46 (0.07) & 0.46 (0.14) \\
TF ($\omega_{PUR}$, p - $\epsilon=$3) & 0.45 (0.13) & 0.49 (0.13) & 0.41 (0.07) & 0.42 (0.03) &0.41 (0.10) & 0.20 (0.10) &0.27 (0.14) & 0.16 (0.09) & 0.45 (0.08) &0.45 (0.10) & 0.44 (0.14) \\
TF ($\omega_{e(UF)r(R)}$, p - $\epsilon=$3) & \textbf{0.52 (0.07)} & 0.48 (0.09) & 0.45 (0.11) & 0.50 (0.11) &0.43 (0.12) & 0.24 (0.13) &0.25 (0.14) & 0.23 (0.14) & 0.47 (0.04) &0.48 (0.06) & 0.46 (0.17) \\
TF ($\omega_{UR}$, p - $\epsilon=$3) & 0.46 (0.12) & 0.47 (0.11) & 0.44 (0.06) & 0.41 (0.05) &0.46 (0.10) & 0.21 (0.14) &0.27 (0.14) & 0.17 (0.15) & 0.46 (0.09) &0.47 (0.11) & 0.44 (0.10) \\
TF ($\omega_{UFR}$, p - $\epsilon=$3) & 0.49 (0.09) & 0.45 (0.10) & 0.43 (0.11) & 0.43 (0.11) &0.43 (0.13) & 0.25 (0.11) &0.30 (0.15) & 0.22 (0.11) & 0.45 (0.10) &0.46 (0.10) & 0.45 (0.21) \\
TF ($\omega_{U}$, p - $\epsilon=$3) & 0.48 (0.07) & 0.41 (0.11) & 0.45 (0.11) & 0.43 (0.13) &0.47 (0.12) & 0.22 (0.10) &0.24 (0.12) & 0.20 (0.11) & 0.49 (0.11) &0.46 (0.10) & 0.56 (0.13) \\
TF ($\omega_{UF}$, p - $\epsilon=$3) & 0.49 (0.09) & 0.44 (0.10) & \textbf{0.45 (0.10)} & 0.43 (0.12) &0.46 (0.11) & 0.25 (0.11) &0.32 (0.14) & 0.21 (0.12) & 0.40 (0.15) &0.41 (0.12) & 0.39 (0.25) \\
\midrule 
\midrule 
\textbf{Non-private Baselines}&&& & & && &&& \\
 CTGAN (n-p) \cite{SDV} & 0.16 (0.03) & 0.44 (0.24) & 0.27 (0.22) & 0.27 (0.17) &0.27 (0.29) & 0.63 (0.25) &0.47 (0.36) & 0.75 (0.15) & 0.36 (0.09) &0.37 (0.08) & 0.34 (0.21) \\
 Gaussian-Copula (n-p)\cite{SDV} & \textbf{0.34 (0.05)} & \textbf{0.45 (0.25)} & \textbf{0.34 (0.26)} & 0.34 (0.14) &0.34 (0.36) & \textbf{0.66 (0.20)} &0.56 (0.28) & 0.75 (0.13) &\textbf{0.37 (0.11)} &0.38 (0.10) & 0.36 (0.15) \\
 \midrule 
 \textbf{Private Baselines}&&& & & && &&& \\
DP-GAN (p - $\epsilon=$1)\cite{qian2023synthcity,chen2020gan} & 0.03 (0.01) & 0.51 (0.32) & 0.16 (0.08) & 0.18 (0.13) &0.14 (0.07) & 0.78 (0.14) &0.75 (0.18) & 0.80 (0.13) & 0.39 (0.10) &0.41 (0.11) & 0.36 (0.21) \\
 DP-GAN (p - $\epsilon=$3)\cite{qian2023synthcity,chen2020gan} & 0.04 (0.01) &\textcolor{blue}{\textbf{ 0.62 (0.33)}} & 0.12 (0.04) & 0.20 (0.07) &0.08 (0.04) & \textcolor{blue}{\textbf{0.79 (0.17)}} &0.78 (0.18) & 0.80 (0.16) & 0.37 (0.17) &0.44 (0.16) & 0.26 (0.18) \\
DP-PGM (p - $\epsilon=$1)\cite{mckenna2021winning} & 0.55 (0.10) & 0.42 (0.23) & 0.27 (0.21) & 0.29 (0.14) &0.26 (0.26) & 0.60 (0.27) &0.44 (0.37) & 0.74 (0.20) & 0.33 (0.11) &0.35 (0.13) & 0.30 (0.19) \\
DP-PGM (p - $\epsilon=$3) \cite{mckenna2021winning}& \textbf{0.61 (0.10)} & 0.37 (0.25) & 0.27 (0.26) & 0.28 (0.13) &0.27 (0.37) & 0.62 (0.28) &0.41 (0.41) & 0.83 (0.14) & 0.31 (0.19) &0.34 (0.21) & 0.25 (0.19) \\
 DP-PGM (target, p - $\epsilon=$1)\cite{mckenna2021winning} & 0.46 (0.05) & 0.43 (0.24) & 0.33 (0.28) & 0.28 (0.20) &0.36 (0.35) & 0.42 (0.33) &0.38 (0.41) & 0.46 (0.29) & 0.41 (0.05) &0.42 (0.09) & 0.40 (0.19) \\
 DP-PGM (target, p - $\epsilon=$3)\cite{mckenna2021winning} & 0.57 (0.05) & 0.41 (0.23) & \textbf{0.35 (0.27)} & 0.28 (0.17) &0.40 (0.37) & 0.52 (0.27) &0.48 (0.33) & 0.55 (0.22) & 0.33 (0.10) &0.38 (0.12) & 0.26 (0.11) \\
PATE-GAN (p - $\epsilon=$1)\cite{qian2023synthcity,yoon2018pategan} & 0.08 (0.03) & 0.40 (0.27) & 0.13 (0.06) & 0.20 (0.08) &0.10 (0.06) & 0.73 (0.19) &0.58 (0.30) & 0.86 (0.08) & 0.33 (0.09) &0.33 (0.05) & 0.33 (0.22) \\
PATE-GAN (p - $\epsilon=$3)\cite{qian2023synthcity,yoon2018pategan} & 0.15 (0.06) & 0.50 (0.29) & 0.29 (0.18) & 0.25 (0.13) &0.32 (0.23) & 0.61 (0.24) &0.53 (0.32) & 0.67 (0.18) & \textbf{0.42 (0.09)} &0.43 (0.09) & 0.41 (0.12) \\
\midrule 
\midrule
 \textbf{Real Data} & N/A & N/A & 0.66 (0.12) & 0.55 (0.09) &0.75 (0.17) & 0.50 (0.09) &0.39 (0.05) & 0.59 (0.15) & 0.14 (0.07) &0.08 (0.07) & 0.48 (0.19) \\
\bottomrule
\end{tabular}
}
\captionof{table}{Recruitment Dataset: trust dimension indices. In bold highest index within each group of synthetic data. In blue highest value across all methods including real data. }
\label{table:recruitment_metrics}
\end{table*}

\begin{table*}[htp!]
\resizebox{1\textwidth}{!}{
\begin{tabular}{lHHHccHccHcc}
\toprule
 Model & Fidelity &Privacy &Utility &Utility & Utility & Fairness & Fairness & Fairness & Robustness & Robustness & Robustness\\
 & & & &(Debiased $\cmark$) & (Debiased $\xmark$) & & (Debiased $\cmark$) &(Debiased $\xmark$) && (Debiased $\cmark$) & (Debiased $\xmark$)\\

\midrule
\textbf{Non-private TrustFormer}&&& & & && &&& \\
 TF ($\omega_{PU}$, n-p) & 0.57 (0.08) & \textbf{0.29 (0.08)} & 0.66 (0.10) & 0.54 (0.07) &0.75 (0.15) & 0.51 (0.05) &0.37 (0.08) & 0.64 (0.07) &\textbf{0.39 (0.11)} &0.40 (0.11) & \textbf{0.37 (0.14)} \\
TF ($\omega_{UR}$, n-p) & 0.54 (0.08) & 0.24 (0.04) & 0.67 (0.10) & 0.56 (0.07) &0.75 (0.15) & 0.51 (0.07) &\textbf{0.41 (0.11)} & 0.59 (0.06) & 0.34 (0.14) &0.35 (0.15) & 0.33 (0.17) \\
TF ($\omega_{PUR}$, n-p) & 0.54 (0.08) & 0.25 (0.05) & \textcolor{blue}{\textbf{0.69 (0.14)}} & \textcolor{blue}{\textbf{0.60 (0.12)}} &\textcolor{blue}{\textbf{0.76 (0.16)}} & 0.51 (0.08) &0.39 (0.09) & 0.61 (0.13) & 0.35 (0.13) &0.36 (0.12) & 0.33 (0.17) \\
TF ($\omega_{e(PU)},\omega_{e(PUF)}$, n-p) & 0.63 (0.08) & 0.25 (0.04) & 0.64 (0.16) & 0.50 (0.17) &0.75 (0.16) & \textbf{0.52 (0.10)} &0.39 (0.09) & 0.64 (0.13) & 0.36 (0.10) &0.37 (0.12) & 0.35 (0.10) \\
 TF ($\omega_{UF}$, n-p) & 0.60 (0.11) & 0.24 (0.04) & 0.65 (0.15) & 0.52 (0.17) &\textcolor{blue}{\textbf{0.76 (0.17)}} & \textbf{0.52 (0.07)} &0.36 (0.07) & \textbf{0.66 (0.12)} & 0.37 (0.15) &0.39 (0.13) & 0.33 (0.19) \\
 TF ($\omega_{e(UF)r(R)}$, n-p) & \textcolor{blue}{\textbf{0.65 (0.09)}} & 0.23 (0.04) & 0.61 (0.18) & 0.45 (0.19) &0.75 (0.17) & 0.50 (0.09) &0.35 (0.09) & 0.63 (0.13) & 0.36 (0.13) &0.38 (0.13) & 0.32 (0.16) \\
 TF ($\omega_{UFR}$, n-p) & 0.59 (0.10) & 0.23 (0.05) & 0.66 (0.15) & 0.54 (0.16) &0.75 (0.16) & 0.52 (0.07) &0.40 (0.07) & 0.62 (0.13) & 0.38 (0.14) &\textbf{0.41 (0.12)} & 0.34 (0.18) \\
 TF ($\omega_{all}$, n-p) & 0.64 (0.09) & 0.22 (0.05) & 0.62 (0.18) & 0.47 (0.19) &0.74 (0.17) & 0.50 (0.09) &0.39 (0.09) & 0.59 (0.14) & 0.37 (0.12) &0.40 (0.11) & 0.32 (0.16) \\
 TF ($\omega_{U}$, n-p) & 0.64 (0.09) & 0.24 (0.04) & 0.65 (0.14) & 0.52 (0.15) &0.75 (0.16) & 0.52 (0.07) &0.37 (0.08) & 0.65 (0.10) & \textbf{0.39 (0.13)} &\textbf{0.41 (0.11)} & 0.35 (0.18) \\

\midrule 
\textbf{Private TrustFormer}&&& & & && &&& \\
TF ($\omega_{PU}$, p - $\epsilon=$1) & 0.41 (0.02) & \textcolor{blue}{\textbf{0.62 (0.14)}} & 0.37 (0.09) & 0.43 (0.10) &0.33 (0.10) & 0.28 (0.17) &0.34 (0.12) & 0.25 (0.21) & 0.43 (0.07) &0.41 (0.08) & 0.47 (0.15) \\
TF ($\omega_{UF}$, p - $\epsilon=$1) & 0.36 (0.06) & 0.60 (0.16) & 0.33 (0.07) & 0.43 (0.08) &0.28 (0.07) &\textbf{0.36 (0.19)} &\textbf{0.43 (0.16)} & \textbf{0.32 (0.21)} & 0.33 (0.09) &0.31 (0.14) & 0.39 (0.06) \\
 TF ($\omega_{all}$, p - $\epsilon=$1) & 0.43 (0.07) & 0.60 (0.15) & 0.35 (0.07) & 0.42 (0.09) &0.31 (0.07) & 0.29 (0.12) &0.40 (0.09) & 0.24 (0.16) & 0.39 (0.09) &0.37 (0.12) & 0.45 (0.17) \\
 TF ($\omega_{UFR}$, p - $\epsilon=$1) & 0.41 (0.09) & 0.60 (0.15) & 0.38 (0.05) & 0.44 (0.07) &0.34 (0.04) & 0.31 (0.11) &\textbf{0.43 (0.08)} & 0.25 (0.15) & 0.40 (0.09) &0.37 (0.13) & 0.44 (0.17) \\
 TF ($\omega_{e(PUF)}$, p - $\epsilon=$1) & 0.39 (0.03) & 0.61 (0.14) & 0.38 (0.07) & 0.46 (0.09) &0.34 (0.06) & 0.32 (0.16) &0.41 (0.11) & 0.28 (0.19) & 0.39 (0.11) &0.34 (0.11) & 0.52 (0.15) \\
TF ($\omega_{e(PU)},\omega_{PUR}$, p - $\epsilon=$1) & 0.42 (0.02) & 0.61 (0.12) & 0.36 (0.08) & 0.43 (0.10) &0.31 (0.07) & 0.27 (0.18) &0.37 (0.10) & 0.22 (0.22) & 0.40 (0.10) &0.35 (0.11) & 0.52 (0.12) \\
TF ($\omega_{U}$, p - $\epsilon=$1) & 0.44 (0.06) & 0.56 (0.16) & 0.40 (0.06) & 0.46 (0.09) &0.36 (0.06) & 0.25 (0.13) &0.35 (0.13) & 0.20 (0.13) & 0.41 (0.10) &0.35 (0.10) & \textcolor{blue}{\textbf{0.59 (0.12)}} \\
 TF ($\omega_{UR}$, p - $\epsilon=$1) & 0.44 (0.06) & 0.56 (0.15) & 0.40 (0.09) & 0.44 (0.10) &0.37 (0.11) & 0.26 (0.13) &0.37 (0.13) & 0.21 (0.13) & 0.42 (0.10) &0.36 (0.12) & 0.58 (0.10) \\
TF ($\omega_{e(UF)r(R)}$, p - $\epsilon=$1) & 0.39 (0.03) & 0.59 (0.11) & 0.37 (0.06) & 0.43 (0.08) &0.33 (0.06) & 0.31 (0.16) &0.36 (0.11) & 0.29 (0.19) & 0.38 (0.10) &0.34 (0.10) & 0.50 (0.12) \\
TF ($\omega_{e(PUF)}$, p - $\epsilon=$3) & 0.47 (0.07) & 0.53 (0.12) & 0.40 (0.13) & 0.43 (0.15) &0.38 (0.12) & 0.30 (0.12) &0.35 (0.10) & 0.27 (0.14) & 0.42 (0.06) &\textcolor{blue}{\textbf{0.49 (0.05)}} & 0.31 (0.16) \\
TF ($\omega_{all}$, p - $\epsilon=$3) & 0.50 (0.07) & 0.52 (0.13) & \textbf{0.45 (0.11)} & 0.48 (0.10) &0.43 (0.13) & 0.29 (0.10) &0.35 (0.11) & 0.25 (0.13) & 0.45 (0.05) &0.48 (0.05) & 0.39 (0.16) \\
TF ($\omega_{PU}$, p - $\epsilon=$3) & 0.50 (0.09) & 0.48 (0.13) & 0.43 (0.10) & 0.46 (0.06) &0.41 (0.12) & 0.26 (0.09) &0.30 (0.08) & 0.24 (0.11) &\textcolor{blue}{\textbf{ 0.51 (0.07)}} &0.49 (0.06) & 0.55 (0.19) \\
TF ($\omega_{e(PU)}$, p - $\epsilon=$3) & 0.49 (0.08) & 0.51 (0.11) & 0.45 (0.11) & 0.48 (0.09) &0.43 (0.12) & 0.29 (0.10) &0.36 (0.10) & 0.25 (0.13) & 0.46 (0.03) &0.46 (0.07) & 0.46 (0.14) \\
TF ($\omega_{PUR}$, p - $\epsilon=$3) & 0.45 (0.13) & 0.49 (0.13) & 0.41 (0.07) & 0.42 (0.03) &0.41 (0.10) & 0.20 (0.10) &0.27 (0.14) & 0.16 (0.09) & 0.45 (0.08) &0.45 (0.10) & 0.44 (0.14) \\
TF ($\omega_{e(UF)r(R)}$, p - $\epsilon=$3) & \textbf{0.52 (0.07)} & 0.48 (0.09) & 0.45 (0.11) & \textbf{0.50 (0.11)} &0.43 (0.12) & 0.24 (0.13) &0.25 (0.14) & 0.23 (0.14) & 0.47 (0.04) &0.48 (0.06) & 0.46 (0.17) \\
TF ($\omega_{UR}$, p - $\epsilon=$3) & 0.46 (0.12) & 0.47 (0.11) & 0.44 (0.06) & 0.41 (0.05) &0.46 (0.10) & 0.21 (0.14) &0.27 (0.14) & 0.17 (0.15) & 0.46 (0.09) &0.47 (0.11) & 0.44 (0.10) \\
TF ($\omega_{UFR}$, p - $\epsilon=$3) & 0.49 (0.09) & 0.45 (0.10) & 0.43 (0.11) & 0.43 (0.11) &0.43 (0.13) & 0.25 (0.11) &0.30 (0.15) & 0.22 (0.11) & 0.45 (0.10) &0.46 (0.10) & 0.45 (0.21) \\
TF ($\omega_{U}$, p - $\epsilon=$3) & 0.48 (0.07) & 0.41 (0.11) & 0.45 (0.11) & 0.43 (0.13) &\textbf{0.47 (0.12)} & 0.22 (0.10) &0.24 (0.12) & 0.20 (0.11) & 0.49 (0.11) &0.46 (0.10) & 0.56 (0.13) \\
TF ($\omega_{UF}$, p - $\epsilon=$3) & 0.49 (0.09) & 0.44 (0.10) & \textbf{0.45 (0.10)} & 0.43 (0.12) &0.46 (0.11) & 0.25 (0.11) &0.32 (0.14) & 0.21 (0.12) & 0.40 (0.15) &0.41 (0.12) & 0.39 (0.25) \\
\midrule 
\midrule 
\textbf{Non-private Baselines}&&& & & && &&& \\
 CTGAN (n-p) \cite{SDV} & 0.16 (0.03) & 0.44 (0.24) & 0.27 (0.22) & 0.27 (0.17) &0.27 (0.29) & 0.63 (0.25) &0.47 (0.36) & \textbf{0.75 (0.15)} & 0.36 (0.09) &0.37 (0.08) & 0.34 (0.21) \\
 Gaussian-Copula (n-p)\cite{SDV} & \textbf{0.34 (0.05)} & \textbf{0.45 (0.25)} & \textbf{0.34 (0.26)} & \textbf{0.34 (0.14)} &\textbf{0.34 (0.36)} & \textbf{0.66 (0.20)} &\textbf{0.56 (0.28)} & \textbf{0.75 (0.13)} &\textbf{0.37 (0.11)} &\textbf{0.38 (0.10)} & \textbf{0.36 (0.15)} \\
 \midrule 
 \textbf{Private Baselines}&&& & & && &&& \\
DP-GAN (p - $\epsilon=$1)\cite{qian2023synthcity,chen2020gan} & 0.03 (0.01) & 0.51 (0.32) & 0.16 (0.08) & 0.18 (0.13) &0.14 (0.07) & 0.78 (0.14) &0.75 (0.18) & 0.80 (0.13) & 0.39 (0.10) &0.41 (0.11) & 0.36 (0.21) \\
 DP-GAN (p - $\epsilon=$3)\cite{qian2023synthcity,chen2020gan} & 0.04 (0.01) &\textcolor{blue}{\textbf{ 0.62 (0.33)}} & 0.12 (0.04) & 0.20 (0.07) &0.08 (0.04) & \textcolor{blue}{\textbf{0.79 (0.17)}} &\textcolor{blue}{\textbf{0.78 (0.18)}} & 0.80 (0.16) & 0.37 (0.17) &\textbf{0.44 (0.16)} & 0.26 (0.18) \\
DP-PGM (p - $\epsilon=$1)\cite{mckenna2021winning} & 0.55 (0.10) & 0.42 (0.23) & 0.27 (0.21) & \textbf{0.29 (0.14)} &0.26 (0.26) & 0.60 (0.27) &0.44 (0.37) & 0.74 (0.20) & 0.33 (0.11) &0.35 (0.13) & 0.30 (0.19) \\
DP-PGM (p - $\epsilon=$3) \cite{mckenna2021winning}& \textbf{0.61 (0.10)} & 0.37 (0.25) & 0.27 (0.26) & 0.28 (0.13) &0.27 (0.37) & 0.62 (0.28) &0.41 (0.41) & 0.83 (0.14) & 0.31 (0.19) &0.34 (0.21) & 0.25 (0.19) \\
 DP-PGM (target, p - $\epsilon=$1)\cite{mckenna2021winning} & 0.46 (0.05) & 0.43 (0.24) & 0.33 (0.28) & 0.28 (0.20) &0.36 (0.35) & 0.42 (0.33) &0.38 (0.41) & 0.46 (0.29) & 0.41 (0.05) &0.42 (0.09) & 0.40 (0.19) \\
 DP-PGM (target, p - $\epsilon=$3)\cite{mckenna2021winning} & 0.57 (0.05) & 0.41 (0.23) & \textbf{0.35 (0.27)} & 0.28 (0.17) &\textbf{0.40 (0.37)} & 0.52 (0.27) &0.48 (0.33) & 0.55 (0.22) & 0.33 (0.10) &0.38 (0.12) & 0.26 (0.11) \\
PATE-GAN (p - $\epsilon=$1)\cite{qian2023synthcity,yoon2018pategan} & 0.08 (0.03) & 0.40 (0.27) & 0.13 (0.06) & 0.20 (0.08) &0.10 (0.06) & 0.73 (0.19) &0.58 (0.30) & \textcolor{blue}{\textbf{0.86 (0.08)}} & 0.33 (0.09) &0.33 (0.05) & 0.33 (0.22) \\
PATE-GAN (p - $\epsilon=$3)\cite{qian2023synthcity,yoon2018pategan} & 0.15 (0.06) & 0.50 (0.29) & 0.29 (0.18) & 0.25 (0.13) &0.32 (0.23) & 0.61 (0.24) &0.53 (0.32) & 0.67 (0.18) & \textbf{0.42 (0.09)} &0.43 (0.09) & \textbf{0.41 (0.12)} \\
\midrule 
\midrule
 \textbf{Real Data} & N/A & N/A & 0.66 (0.12) & 0.55 (0.09) &0.75 (0.17) & 0.50 (0.09) &0.39 (0.05) & 0.59 (0.15) & 0.14 (0.07) &0.08 (0.07) & 0.48 (0.19) \\
\bottomrule
\end{tabular}
}
\captionof{table}{Recruitment downstream task evaluation. Study on the effect of debiasing in utility training on trust indices of utility, fairness and robustness.}
\label{table:recruitment_metric_biased_debiased}
\end{table*}

\begin{table*}[htp!]
\resizebox{1\textwidth}{!} {
\begin{tabular}{lrrrrrrrrrr}
\toprule
 Model &$\omega_{all}$ &$\omega_{e(PU)}$ &$\omega_{e(PUF)}$ &$\omega_{e(UF)r(R)}$ &$\omega_{PU}$ &$\omega_{PUR}$ &$\omega_{U}$ &$\omega_{UF}$ &$\omega_{UFR}$ &$\omega_{UR}$ \\
\midrule
\textbf{Non-Private TrustFormer}&&& & & && &&& \\
TF ($\omega_{PU}$, n-p) & \rankone{1} & \rankone{1} & \rankone{1} & \rankone{1} & 19 &19 & \rankthree{3} &4 & \rankone{1} & \rankone{1} \\
 TF ($\omega_{UR}$, n-p) &10 & 16 & 8 & 7 & 23 &29 & \ranktwo{2} & \rankthree{3} & 7 &8 \\
TF ($\omega_{PUR}$, n-p) & 6 &6 & 4 & \rankthree{3} & 21 &21 & \rankone{1} & \rankone{1} & 5 &4 \\
TF ($\omega_{e(PU)},\omega_{e(PUF)}$, n-p) & \rankthree{3} &5 & \rankthree{3} & \ranktwo{2} & 22 &24 & 7 &7 & 6 &6 \\
TF ($\omega_{UF}$, n-p) & 4 &8 & 5 & 5 & 24 &26 & 5 &5 & 4 &5 \\
TF ($\omega_{e(UF)r(R)}$, n-p) & 7 & 18 & 9 & 8 & 31 &30 & 9 &9 & 9 & 10 \\
TF ($\omega_{UFR}$, n-p) & 5 & 10 & 6 & 6 & 27 &27 & 4 & \ranktwo{2} & \rankthree{3} & \rankthree{3} \\
TF ($\omega_{all}$, n-p) & 8 & 21 &10 & 9 & 32 &31 & 8 &8 & 8 &7 \\
TF ($\omega_{U}$, n-p) & \ranktwo{2} &4 & \ranktwo{2} & 4 & 25 &23 & 6 &6 & \ranktwo{2} & \ranktwo{2} \\
\midrule
\textbf{Private TrustFormer}&&& & & && &&& \\
TF ($\omega_{PU}$, p - $\epsilon=$1) &22 & 13 &19 &26 &4 & 5 &25 & 30 &29 & 22 \\
TF ($\omega_{UF}$, p - $\epsilon=$1) &32 & 29 &26 &22 & 16 &20 &30 & 22 &34 & 32 \\
TF ($\omega_{all}$, p - $\epsilon=$1) &25 & 19 &23 &24 & 11 &16 &27 & 31 &33 & 27 \\
TF ($\omega_{UFR}$, p - $\epsilon=$1) &21 & 14 &17 &19 &5 &13 &23 & 23 &26 & 24 \\
TF ($\omega_{e(PUF)}$, p - $\epsilon=$1) &20 & 12 &14 &18 & \ranktwo{2} &10 &22 & 19 &23 & 23 \\
TF ($\omega_{e(PU)},\omega_{PUR}$, p - $\epsilon=$1) &28 & 23 &28 &31 & 10 &15 &26 & 34 &35 & 25 \\
 TF ($\omega_{U}$, p - $\epsilon=$1) &26 & 17 &27 &29 &6 & 8 &20 & 32 &31 & 21 \\
TF ($\omega_{UR}$, p - $\epsilon=$1) &23 & 15 &25 &25 &7 & 7 &21 & 29 &27 & 19 \\
TF ($\omega_{e(UF)r(R)}$, p - $\epsilon=$1) &27 & 20 &21 &20 &9 &17 &24 & 24 &28 & 26 \\
TF ($\omega_{e(PUF)}$, p - $\epsilon=$3) &16 & 11 &16 &17 & 12 &14 &19 & 21 &21 & 20 \\
TF ($\omega_{all}$, p - $\epsilon=$3) & 9 & \ranktwo{2} &12 &14 & \rankone{1} & \rankthree{3} &13 & 18 &15 & 15 \\
TF ($\omega_{PU}$, p - $\epsilon=$3) &13 &7 &18 &21 & 14 & \rankone{1} &17 & 25 &16 & 11 \\
TF ($\omega_{e(PU)}$, p - $\epsilon=$3) &11 & \rankthree{3} &13 &15 & \rankthree{3} & \ranktwo{2} &12 & 17 &13 & 13 \\
TF ($\omega_{PUR}$, p - $\epsilon=$3) &33 & 30 &34 &35 & 15 & 9 &18 & 38 &36 & 17 \\
TF ($\omega_{e(UF)r(R)}$, p - $\epsilon=$3) &15 &9 &24 &23 &8 & 4 &10 & 27 &20 & 12 \\
 TF ($\omega_{UR}$, p - $\epsilon=$3) &31 & 27 &32 &34 & 13 & 6 &15 & 37 &32 & 14 \\
 TF ($\omega_{UFR}$, p - $\epsilon=$3) &24 & 22 &29 &28 & 17 &12 &16 & 28 &22 & 16 \\
TF ($\omega_{U}$, p - $\epsilon=$3) &30 & 28 &33 &33 & 20 &11 &11 & 33 &25 &9 \\
TF ($\omega_{UF}$, p - $\epsilon=$3) &29 & 26 &31 &27 & 18 &18 &14 & 26 &30 & 18 \\
\midrule
\midrule
\textbf{Non-Private Baselines}&&& & & && &&& \\
CTGAN (n-p)\cite{SDV} &35 & 35 &35 &32 & 33 &33 &34 & 13 &14 & 33 \\
Gaussian-Copula (n-p)\cite{SDV} &14 & 25 & 7 &10 & 26 &28 &29 & 10 &10 & 29 \\
\midrule
\textbf{Private Baselines}&&& & & && &&& \\
 DP-GAN (p - $\epsilon=$1)\cite{qian2023synthcity,chen2020gan} &38 & 38 &38 &38 & 36 &35 &36 & 20 &24 & 36 \\
DP-GAN (p - $\epsilon=$3)\cite{qian2023synthcity,chen2020gan} &37 & 36 &36 &37 & 37 &37 &38 & 36 &37 & 38 \\
 DP-PGM (p - $\epsilon=$1)\cite{mckenna2021winning} &17 & 32 &15 &13 & 34 &34 &35 & 15 &18 & 34 \\
  DP-PGM (p - $\epsilon=$3)\cite{mckenna2021winning} &18 & 33 &20 &12 & 35 &36 &33 & 14 &19 & 35 \\
 DP-PGM (target, p - $\epsilon=$1) \cite{mckenna2021winning}&19 & 31 &22 &16 & 29 &25 &31 & 16 &17 & 28 \\
 DP-PGM (target, p - $\epsilon=$3)\cite{mckenna2021winning} &12 & 24 &11 &11 & 30 &32 &28 & 11 &12 & 31 \\
 PATE-GAN (p - $\epsilon=$1)\cite{qian2023synthcity,yoon2018pategan} &36 & 37 &37 &36 & 38 &38 &37 & 35 &38 & 37 \\
 PATE-GAN (p - $\epsilon=$3) \cite{qian2023synthcity,yoon2018pategan} &34 & 34 &30 &30 & 28 &22 &32 & 12 &11 & 30 \\
\bottomrule
\end{tabular}
}
\captionof{table}{Recruitment dataset ranking. We see that overall TrustFormers that use trustworthiness index driven cross-validation corresponding to the desired trade-offs outperform other synthetic data,  across these desired trade-offs defined in Table \ref{Table:weights}. Note that the ranking here is on the mean trustworthiness index, where downstream tasks are evaluated on the test data in each real data fold.  }
\label{table:recruitment_ranking}
\end{table*}

\begin{table*}[htp!]
\resizebox{1\textwidth}{!} {
\begin{tabular}{lrrrrrrrrrr}
\toprule
 Model &$\omega_{all}$ &$\omega_{e(PU)}$ &$\omega_{e(PUF)}$ &$\omega_{e(UF)r(R)}$ &$\omega_{PU}$ &$\omega_{PUR}$ &$\omega_{U}$ &$\omega_{UF}$ &$\omega_{UFR}$ &$\omega_{UR}$ \\
\midrule
\textbf{Non-Private TrustFormer}&&& & & && &&& \\
 TF ($\omega_{PU}$, n-p) &\rankone{1}&\rankone{1}&\rankone{1}&\rankone{1}& 17 & 19 &\rankthree{3}&\rankone{1}&\rankone{1}&\rankone{1}\\
 TF ($\omega_{UR}$, n-p) & 5 &\ranktwo{2}&\ranktwo{2}&\rankthree{3}& 22 & 26 &\ranktwo{2}&\ranktwo{2}& 5 & 10 \\
 TF ($\omega_{PUR}$, n-p) & 6 & 5 &\rankthree{3}&\ranktwo{2}& 20 & 23 &\rankone{1}&\rankthree{3}&\rankthree{3}& 6 \\
 TF ($\omega_{e(PU)}$, $\omega_{e(PUF)}$, n-p) &\ranktwo{2}&\rankthree{3}& 5 & 7 & 23 & 21 & 7 & 7 & 4 &\rankthree{3}\\
 TF ($\omega_{UF}$, n-p) &\rankthree{3}& 8 & 4 & 4 & 24 & 24 & 5 & 6 & 7 & 7 \\
 TF ($\omega_{e(UF)r(R)}$, n-p) & 8 & 21 & 8 & 8 & 27 & 27 & 9 & 9 & 9 & 11 \\
 TF ($\omega_{UFR}$, n-p) & 7 & 12 & 7 & 6 & 26 & 25 & 4 & 4 &\ranktwo{2}&\ranktwo{2}\\
 TF ($\omega_{all}$, n-p) & 9 & 24 & 9 & 9 & 28 & 28 & 8 & 8 & 8 & 5 \\
 TF ($\omega_{U}$, n-p) & 4 & 9 & 6 & 5 & 25 & 22 & 6 & 5 & 6 & 4 \\

\midrule
\textbf{Private TrustFormer}&&& & & && &&& \\
 TF ($\omega_{PU}$, p - $\epsilon=$1) & 16 & 13 & 17 & 27 & 7 & 7 & 25 & 34 & 28 & 22 \\
 TF ($\omega_{UF}$, p - $\epsilon=$1) & 29 & 20 & 21 & 21 & 8 & 16 & 28 & 23 & 36 & 31 \\
 TF ($\omega_{all}$, p - $\epsilon=$1) & 20 & 23 & 22 & 25 & 13 & 18 & 27 & 30 & 34 & 27 \\
 TF ($\omega_{UFR}$, p - $\epsilon=$1) & 18 & 15 & 13 & 15 & 6 & 15 & 20 & 20 & 24 & 23 \\
 TF ($\omega_{e(PUF)}$, p - $\epsilon=$1) & 21 & 18 & 15 & 17 &\ranktwo{2}& 14 & 22 & 22 & 26 & 24 \\
 TF ($\omega_{e(PU)}$, $\omega_{PUR}$, p - $\epsilon=$1) & 24 & 19 & 25 & 30 & 9 & 10 & 26 & 36 & 38 & 25 \\
 TF ($\omega_{U}$, p - $\epsilon=$1) & 17 & 7 & 12 & 23 &\rankthree{3}& 8 & 19 & 32 & 29 & 20 \\
 TF ($\omega_{UR}$, p - $\epsilon=$1) & 13 & 11 & 16 & 24 & 10 & 4 & 21 & 31 & 27 & 19 \\
TF ($\omega_{e(UF)r(R)}$, p - $\epsilon=$1) & 22 & 16 & 19 & 20 &\rankone{1}& 12 & 24 & 26 & 31 & 26 \\
 TF ($\omega_{e(PUF)}$, p - $\epsilon=$3) & 15 & 22 & 20 & 18 & 16 & 17 & 23 & 21 & 20 & 21 \\
 TF ($\omega_{all}$, p - $\epsilon=$3) & 10 & 4 & 10 & 12 & 4 &\ranktwo{2}& 14 & 19 & 17 & 15 \\
 TF ($\omega_{PU}$, p - $\epsilon=$3) & 12 & 10 & 14 & 19 & 15 & 6 & 16 & 24 & 18 & 13 \\
 TF ($\omega_{e(PU)}$, p - $\epsilon=$3) & 11 & 6 & 11 & 13 & 5 &\rankone{1}& 13 & 18 & 16 & 14 \\
 TF ($\omega_{PUR}$, p - $\epsilon=$3) & 31 & 17 & 30 & 33 & 14 & 9 & 18 & 38 & 35 & 17 \\
TF ($\omega_{e(UF)r(R)}$, p - $\epsilon=$3) & 14 & 14 & 24 & 26 & 11 &\rankthree{3}& 11 & 29 & 22 & 12 \\
 TF ($\omega_{UR}$, p - $\epsilon=$3) & 32 & 25 & 32 & 35 & 12 & 5 & 10 & 37 & 32 & 8 \\
 TF ($\omega_{UFR}$, p - $\epsilon=$3) & 23 & 26 & 26 & 29 & 18 & 13 & 17 & 28 & 23 & 16 \\
 TF ($\omega_{U}$, p - $\epsilon=$3) & 28 & 27 & 33 & 32 & 21 & 11 & 12 & 35 & 25 & 9 \\
 TF ($\omega_{UF}$, p - $\epsilon=$3) & 33 & 28 & 31 & 28 & 19 & 20 & 15 & 27 & 30 & 18 \\

\midrule
\midrule
\textbf{Non-Private Baselines}&&& & & && &&& \\
 SDV-CTGAN (n-p) & 35 & 35 & 35 & 34 & 33 & 33 & 34 & 13 & 12 & 33 \\
 SDV-Gaussian-Copula (n-p) & 25 & 30 & 18 & 11 & 29 & 30 & 30 & 10 & 11 & 30 \\

\midrule
\textbf{Private Baselines}&&& & & && &&& \\
 DP-GAN (p - $\epsilon=$1) & 38 & 38 & 38 & 38 & 36 & 36 & 36 & 17 & 19 & 36 \\
 DP-GAN (p - $\epsilon=$3) & 37 & 36 & 36 & 37 & 37 & 34 & 38 & 25 & 33 & 37 \\
 DP-PGM (p - $\epsilon=$1) & 30 & 32 & 28 & 14 & 34 & 35 & 33 & 14 & 21 & 35 \\
 DP-PGM (target, p - $\epsilon=$1) & 27 & 31 & 27 & 22 & 32 & 31 & 31 & 16 & 13 & 28 \\
 DP-PGM (p - $\epsilon=$3) & 26 & 33 & 29 & 16 & 35 & 37 & 35 & 15 & 15 & 34 \\
 DP-PGM (target, p - $\epsilon=$3) & 19 & 29 & 23 & 10 & 31 & 32 & 29 & 12 & 14 & 32 \\
 PATE-GAN (p - $\epsilon=$1) & 36 & 37 & 37 & 36 & 38 & 38 & 37 & 33 & 37 & 38 \\
 PATE-GAN (p - $\epsilon=$3) & 34 & 34 & 34 & 31 & 30 & 29 & 32 & 11 & 10 & 29 \\
\bottomrule
\end{tabular}}
\captionof{table}{Recruitment dataset.  Ranking  under uncertainty of all considered synthetic data using $R^{\alpha}_{\tau}$ defined in Equation \eqref{eq:mean_var} for $\alpha=0.1$. We see that TrustFormer synthetic data is still aligned with the prescribed safeguards while having low volatility in its trustworthiness index.  }
\label{table:recruitment_ranking_uncertainty}
\end{table*}



\subsection{Law School}
\begin{table*}[htp!]
\resizebox{0.95\textwidth}{!}{
\centering
\begin{tabular}{lcccHHcHHcHH}
\toprule
 Model & Fidelity &Privacy &Utility &Utility & Utility & Fairness & Fairness & Fairness & Robustness & Robustness & Robustness\\
\midrule
\textbf{Non-private TrustFormer}&&& & & && &&& &\\
 TF($\omega_{U}$, n-p) & 0.37 (0.07) & 0.34 (0.13) & \textcolor{blue}{\textbf{0.60 (0.07)}} & 0.56 (0.07) &\textcolor{blue}{\textbf{0.63 (0.07)}} & 0.40 (0.02) &0.43 (0.09) & 0.39 (0.04) & 0.44 (0.16) &\textbf{0.43 (0.15)} & 0.48 (0.22) \\
 TF($\omega_{all}$, n-p) & \textbf{0.47 (0.06)} & 0.30 (0.10) & 0.57 (0.08) & 0.57 (0.05) &0.57 (0.12) & 0.44 (0.03) &0.44 (0.07) & 0.43 (0.02) & 0.41 (0.17) &0.41 (0.14) & 0.41 (0.23) \\
 TF($\omega_{e(UF)r(R)}$, n-p) & 0.45 (0.05) & 0.32 (0.11) & 0.57 (0.08) & 0.55 (0.07) &0.58 (0.11) & \textbf{0.46 (0.06)} &0.46 (0.11) & \textbf{0.45 (0.03)} & 0.40 (0.17) &0.40 (0.14) & 0.41 (0.23) \\
 TF($\omega_{PUR}$, n-p) & 0.30 (0.07) & 0.44 (0.07) & 0.60 (0.08) & 0.57 (0.10) &0.62 (0.08) & 0.41 (0.07) &0.45 (0.10) & 0.39 (0.06) & 0.42 (0.13) &0.38 (0.13) & 0.53 (0.22) \\
TF($\omega_{UR}$, n-p) & 0.34 (0.04) & 0.39 (0.13) & 0.60 (0.08) & 0.56 (0.08) &0.62 (0.08) & 0.42 (0.06) &0.45 (0.10) & 0.40 (0.05) & 0.43 (0.15) &0.40 (0.16) & 0.51 (0.18) \\
 TF($\omega_{e(PUF)}$, n-p) & 0.40 (0.09) & 0.37 (0.12) & 0.59 (0.08) & \textcolor{blue}{\textbf{0.58 (0.06)}} &0.59 (0.10) & 0.40 (0.08) &0.45 (0.09) & 0.38 (0.09) & 0.43 (0.16) &0.41 (0.16) & 0.48 (0.20) \\
TF($\omega_{e(PU)}$, n-p) & 0.35 (0.06) & 0.41 (0.12) & 0.59 (0.07) & 0.55 (0.08) &0.62 (0.08) & 0.40 (0.08) &0.45 (0.09) & 0.37 (0.09) & 0.43 (0.16) &0.39 (0.16) & 0.51 (0.21) \\
TF($\omega_{UFR}$, n-p) & 0.32 (0.06) & 0.40 (0.13) & 0.60 (0.08) & 0.56 (0.08) &0.63 (0.09) & 0.42 (0.06) &0.45 (0.10) & 0.40 (0.05) & 0.42 (0.14) &0.39 (0.16) & 0.49 (0.18) \\
TF($\omega_{PU}$, n-p) & 0.29 (0.07) & \textbf{0.45 (0.07)} & 0.60 (0.09) & 0.57 (0.11) &0.62 (0.08) & 0.41 (0.04) &0.44 (0.09) & 0.38 (0.05) & 0.44 (0.10) &0.42 (0.09) & 0.48 (0.23) \\
TF($\omega_{UF}$, n-p) & 0.34 (0.08) & 0.34 (0.11) & 0.59 (0.09) & 0.55 (0.12) &0.62 (0.08) & 0.44 (0.06) &\textbf{0.46 (0.10)} & 0.43 (0.05) & \textbf{0.46 (0.12)} &0.41 (0.16) & \textcolor{blue}{\textbf{0.57 (0.11)}} \\
\midrule 
\textbf{Private TrustFormer}&&& & & && &&& &\\
 TF($\omega_{PU}$, p - $\varepsilon=$1) & 0.47 (0.12) & \textbf{0.57 (0.14)} & 0.34 (0.07) & 0.36 (0.04) &0.33 (0.09) & 0.51 (0.20) &0.62 (0.12) & 0.45 (0.24) & 0.36 (0.13) &0.40 (0.09) & 0.29 (0.28) \\
 TF($\omega_{e(PUF)}$, p - $\varepsilon=$1) & 0.49 (0.11) & 0.54 (0.10) & 0.34 (0.07) & 0.38 (0.05) &0.32 (0.10) & 0.50 (0.19) &0.62 (0.12) & 0.43 (0.23) & 0.40 (0.10) &0.40 (0.09) & 0.40 (0.22) \\
 TF($\omega_{PUR}$, p - $\varepsilon=$1) & 0.47 (0.09) & 0.54 (0.17) & 0.35 (0.06) & 0.36 (0.06) &0.34 (0.13) & \textbf{0.55 (0.17)} &\textcolor{blue}{\textbf{0.64 (0.10)}} & 0.50 (0.20) & 0.38 (0.08) &0.36 (0.05) & 0.41 (0.18) \\
 TF($\omega_{e(PU)}$, p - $\varepsilon=$1) & 0.51 (0.09) & 0.52 (0.11) & 0.36 (0.11) & 0.43 (0.15) &0.32 (0.11) & 0.52 (0.18) &0.58 (0.17) & 0.48 (0.19) & 0.41 (0.11) &0.42 (0.10) & 0.40 (0.22) \\
 TF($\omega_{UR}$, p - $\varepsilon=$1) & 0.56 (0.09) & 0.42 (0.19) & 0.38 (0.11) & 0.44 (0.14) &0.35 (0.15) & 0.53 (0.19) &0.57 (0.19) & 0.50 (0.19) & 0.40 (0.11) &0.38 (0.10) & \textbf{0.43 (0.17)} \\
TF($\omega_{UFR}$, p - $\varepsilon=$1) & 0.56 (0.09) & 0.38 (0.20) & 0.37 (0.09) & 0.38 (0.04) &0.37 (0.17) & 0.51 (0.20) &0.64 (0.13) & 0.44 (0.24) & 0.38 (0.10) &0.39 (0.10) & 0.36 (0.20) \\
TF($\omega_{all}$, p - $\varepsilon=$1) & 0.58 (0.08) & 0.39 (0.17) & 0.36 (0.13) & 0.38 (0.11) &0.35 (0.17) & 0.48 (0.22) &0.57 (0.17) & 0.42 (0.24) & 0.36 (0.12) &0.39 (0.11) & 0.31 (0.28) \\
 TF($\omega_{e(UF)r(R)}$, p - $\varepsilon=$1) & 0.57 (0.12) & 0.41 (0.09) & 0.33 (0.14) & 0.38 (0.11) &0.29 (0.16) & 0.45 (0.23) &0.57 (0.17) & 0.39 (0.27) & 0.36 (0.12) &0.40 (0.11) & 0.28 (0.27) \\
TF($\omega_{UF}$, p - $\varepsilon=$1) & 0.53 (0.13) & 0.43 (0.18) & 0.35 (0.13) & 0.37 (0.12) &0.34 (0.13) & 0.48 (0.26) &0.60 (0.19) & 0.41 (0.29) & 0.33 (0.14) &0.40 (0.11) & 0.22 (0.30) \\
TF($\omega_{U}$, p - $\varepsilon=$1) & 0.58 (0.12) & 0.33 (0.24) & 0.37 (0.17) & 0.38 (0.10) &0.36 (0.23) & 0.55 (0.19) &0.60 (0.16) & \textbf{0.52 (0.21)} & 0.35 (0.15) &0.40 (0.09) & 0.26 (0.28) \\
TF($\omega_{PU}$, p - $\varepsilon=$3) & 0.52 (0.04) & 0.39 (0.23) & 0.43 (0.12) & 0.41 (0.09) &0.44 (0.18) & 0.46 (0.13) &0.57 (0.12) & 0.40 (0.14) & 0.40 (0.17) &0.44 (0.10) & 0.33 (0.30) \\
TF($\omega_{PUR}$, p - $\varepsilon=$3) & 0.58 (0.06) & 0.30 (0.24) & 0.43 (0.12) & 0.42 (0.06) &0.44 (0.22) & 0.47 (0.12) &0.55 (0.10) & 0.42 (0.13) & \textbf{0.42 (0.16)} &0.44 (0.13) & 0.37 (0.25) \\
 TF($\omega_{e(PU)}$, p - $\varepsilon=$3) & 0.61 (0.08) & 0.32 (0.24) & 0.43 (0.15) & \textbf{0.46 (0.05)} &0.40 (0.25) & 0.46 (0.13) &0.50 (0.11) & 0.44 (0.13) & 0.41 (0.17) &0.46 (0.14) & 0.32 (0.25) \\
TF($\omega_{e(PUF)},\omega_{all}$, p - $\varepsilon=$3) & 0.65 (0.05) & 0.26 (0.19) & 0.40 (0.17) & 0.42 (0.09) &0.38 (0.26) & 0.47 (0.13) &0.51 (0.14) & 0.44 (0.13) & 0.43 (0.16) &0.45 (0.14) & 0.39 (0.22) \\
 TF($\omega_{e(UF)r(R)}$, p - $\varepsilon=$3) & 0.66 (0.05) & 0.25 (0.15) & 0.40 (0.17) & 0.42 (0.09) &0.38 (0.26) & 0.47 (0.13) &0.52 (0.14) & 0.43 (0.13) & 0.40 (0.18) &0.41 (0.17) & 0.38 (0.22) \\
TF($\omega_{UF}$, p - $\varepsilon=$3) & \textcolor{blue}{\textbf{0.67 (0.03)}} & 0.15 (0.08) & \textbf{0.47 (0.11)} & 0.41 (0.10) &0.50 (0.16) & 0.46 (0.11) &0.56 (0.17) & 0.40 (0.09) & 0.38 (0.16) &0.43 (0.11) & 0.30 (0.29) \\
TF($\omega_{UR}$, p - $\varepsilon=$3) & 0.64 (0.06) & 0.20 (0.18) & 0.44 (0.11) & 0.41 (0.08) &0.46 (0.17) & 0.46 (0.09) &0.55 (0.16) & 0.41 (0.08) & 0.38 (0.19) &0.44 (0.13) & 0.29 (0.30) \\
 TF($\omega_{UFR}$, p - $\varepsilon=$3) & 0.66 (0.04) & 0.17 (0.20) & 0.43 (0.09) & 0.40 (0.08) & \textbf{ 0.46 (0.15)} & 0.47 (0.09) &0.56 (0.16) & 0.42 (0.08) & 0.37 (0.16) &0.42 (0.09) & 0.30 (0.31) \\
TF($\omega_{U}$, p - $\varepsilon=$3) & 0.64 (0.04) & 0.12 (0.07) & 0.43 (0.12) & 0.44 (0.07) &0.43 (0.18) & 0.48 (0.12) &0.53 (0.14) & 0.46 (0.11) & 0.41 (0.16) & \textbf{ 0.47 (0.07)} & 0.30 (0.31) \\

\midrule 
\midrule 
\textbf{Non-Private Baselines}&&& & & && &&& &\\
CTGAN (n-p) \cite{SDV} & 0.22 (0.04) & \textbf{0.65 (0.12)} & 0.37 (0.19) & \textbf{0.57 (0.11)} &0.28 (0.24) & 0.45 (0.19) &0.43 (0.08) & 0.46 (0.27) & \textbf{0.46 (0.11)} &\textbf{0.44 (0.11)} & 0.49 (0.21) \\
 Gaussian-Copula (n-p)\cite{SDV}& \textbf{0.36 (0.02)} & 0.62 (0.26) & \textbf{0.40 (0.03)} & 0.45 (0.02) &\textbf{0.38 (0.06)} & \textbf{0.61 (0.09)} &\textbf{0.53 (0.09)} &\textbf{ 0.67 (0.10)} & 0.44 (0.11) &0.39 (0.08) & \textbf{0.55 (0.25)} \\
 \midrule
 \textbf{Private Baselines}&&& & & && &&& &\\
DP-GAN (p - $\varepsilon=$1)\cite{xie2018differentially,qian2023synthcity}& 0.05 (0.01) & \textcolor{blue}{\textbf{0.83 (0.17)}} & 0.24 (0.09) & 0.39 (0.16) &0.17 (0.19) & 0.55 (0.15) &0.59 (0.10) & 0.52 (0.24) & 0.52 (0.08) &0.54 (0.12) & 0.48 (0.34) \\
DP-GAN (p - $\varepsilon=$3)\cite{xie2018differentially,qian2023synthcity}& 0.06 (0.01) & \textcolor{blue}{\textbf{0.83 (0.17)}} & 0.21 (0.16) & 0.44 (0.18) &0.13 (0.14) & 0.64 (0.08) &0.59 (0.16) & 0.67 (0.11) & \textcolor{blue}{\textbf{0.51 (0.25)}} & \textcolor{blue}{\textbf{ 0.58 (0.22)}} & 0.40 (0.35) \\
DP-PGM (p - $\varepsilon=$1)\cite{mckenna2021winning} & 0.34 (0.06) & 0.55 (0.24) & 0.30 (0.10) & 0.44 (0.09) &0.24 (0.11) & \textcolor{blue}{\textbf{0.70 (0.05)}} &\textbf{0.59 (0.03)} & \textcolor{blue}{\textbf{0.78 (0.09)}} & 0.46 (0.10) &0.46 (0.08) & 0.45 (0.23) \\
DP-PGM (p - $\varepsilon=$3)\cite{mckenna2021winning} & 0.34 (0.07) & 0.54 (0.23) & 0.39 (0.14) & 0.48 (0.13) &0.35 (0.15) & 0.56 (0.11) &0.49 (0.11) & 0.61 (0.16) & 0.43 (0.10) &0.40 (0.09) &\textbf{ 0.50 (0.17)} \\
DP-PGM (target, p - $\varepsilon=$1)\cite{mckenna2021winning} & 0.34 (0.06) & 0.60 (0.24) & 0.42 (0.08) & 0.41 (0.16) & \textbf{ 0.42 (0.11)} & 0.49 (0.14) &0.52 (0.18) & 0.46 (0.20) & 0.43 (0.13) &0.40 (0.10) & 0.49 (0.27) \\
DP-PGM (target, p - $\varepsilon=$3)\cite{mckenna2021winning} & \textbf{0.35 (0.07)} & 0.54 (0.24) & \textbf{0.45 (0.17)} & \textcolor{blue}{\textbf{0.60 (0.07)}} &0.37 (0.22) & 0.51 (0.13) &0.45 (0.10) & 0.56 (0.16) & 0.38 (0.18) &0.38 (0.15) & 0.36 (0.28) \\
PATE-GAN (p - $\varepsilon=$1) \cite{yoon2018pategan,qian2023synthcity} & 0.18 (0.05) & 0.68 (0.24) & 0.38 (0.04) & 0.49 (0.16) &0.32 (0.09) & 0.61 (0.11) &0.52 (0.13) & 0.68 (0.14) & 0.48 (0.17) &0.50 (0.11) & 0.43 (0.31) \\
PATE-GAN (p - $\varepsilon=$3)\cite{yoon2018pategan,qian2023synthcity} & 0.21 (0.03) & 0.67 (0.27) & 0.33 (0.07) & 0.45 (0.04) &0.26 (0.08) & 0.65 (0.07) &0.57 (0.06) & \textcolor{blue}{\textbf{0.71 (0.11)}} & 0.35 (0.06) &0.33 (0.12) & 0.37 (0.31) \\
\midrule 
\midrule
\textbf{Real Data} & N/A & N/A & 0.56 (0.08) & 0.53 (0.08) &0.58 (0.12) & 0.47 (0.17) &0.44 (0.08) & 0.49 (0.24) & 0.38 (0.10) &0.38 (0.06) & 0.36 (0.20) \\
\bottomrule
\end{tabular}
}
\captionof{table}{Law school dataset: trust dimension indices. In bold highest index within each group of synthetic data. In blue highest value across all methods including real data. }

\label{table:law_school_metrics}

\end{table*}

\begin{table*}[htp]
\resizebox{0.95\textwidth}{!}{
\centering
\begin{tabular}{lHHHccHccHcc}
\toprule
 Model & Fidelity &Privacy &Utility &Utility & Utility & Fairness & Fairness & Fairness & Robustness & Robustness & Robustness\\
 & & & &(Debiased $\cmark$) & (Debiased $\xmark$) & & (Debiased$\cmark$) &(Debiased $\xmark$) &&(Debiased $\cmark$) & (Debiased $\xmark$)\\
\midrule
\textbf{Non-private TrustFormer}&&& & & && &&& &\\
 TF($\omega_{U}$, n-p) & 0.37 (0.07) & 0.34 (0.13) &\textcolor{blue}{\textbf{0.60 (0.07)}} & 0.56 (0.07) &\textcolor{blue}{\textbf{0.63 (0.07)}} & 0.40 (0.02) &0.43 (0.09) & 0.39 (0.04) & 0.44 (0.16) &\textbf{0.43 (0.15)} & 0.48 (0.22) \\
 TF($\omega_{all}$, n-p) & \textbf{0.47 (0.06)} & 0.30 (0.10) & 0.57 (0.08) & 0.57 (0.05) &0.57 (0.12) & 0.44 (0.03) &0.44 (0.07) & 0.43 (0.02) & 0.41 (0.17) &0.41 (0.14) & 0.41 (0.23) \\
 TF($\omega_{e(UF)r(R)}$, n-p) & 0.45 (0.05) & 0.32 (0.11) & 0.57 (0.08) & 0.55 (0.07) &0.58 (0.11) & \textbf{0.46 (0.06)} &0.46 (0.11) & \textbf{0.45 (0.03)} & 0.40 (0.17) &0.40 (0.14) & 0.41 (0.23) \\
 TF($\omega_{PUR}$, n-p) & 0.30 (0.07) & 0.44 (0.07) & 0.60 (0.08) & 0.57 (0.10) &0.62 (0.08) & 0.41 (0.07) &0.45 (0.10) & 0.39 (0.06) & 0.42 (0.13) &0.38 (0.13) & 0.53 (0.22) \\
TF($\omega_{UR}$, n-p) & 0.34 (0.04) & 0.39 (0.13) & 0.60 (0.08) & 0.56 (0.08) &0.62 (0.08) & 0.42 (0.06) &0.45 (0.10) & 0.40 (0.05) & 0.43 (0.15) &0.40 (0.16) & 0.51 (0.18) \\
 TF($\omega_{e(PUF)}$, n-p) & 0.40 (0.09) & 0.37 (0.12) & 0.59 (0.08) & \textcolor{blue}{\textbf{0.58 (0.06)}} &0.59 (0.10) & 0.40 (0.08) &0.45 (0.09) & 0.38 (0.09) & 0.43 (0.16) &0.41 (0.16) & 0.48 (0.20) \\
TF($\omega_{e(PU)}$, n-p) & 0.35 (0.06) & 0.41 (0.12) & 0.59 (0.07) & 0.55 (0.08) &0.62 (0.08) & 0.40 (0.08) &0.45 (0.09) & 0.37 (0.09) & 0.43 (0.16) &0.39 (0.16) & 0.51 (0.21) \\
TF($\omega_{UFR}$, n-p) & 0.32 (0.06) & 0.40 (0.13) & 0.60 (0.08) & 0.56 (0.08) &0.63 (0.09) & 0.42 (0.06) &0.45 (0.10) & 0.40 (0.05) & 0.42 (0.14) &0.39 (0.16) & 0.49 (0.18) \\
TF($\omega_{PU}$, n-p) & 0.29 (0.07) & \textbf{0.45 (0.07)} & 0.60 (0.09) & 0.57 (0.11) &0.62 (0.08) & 0.41 (0.04) &0.44 (0.09) & 0.38 (0.05) & 0.44 (0.10) &0.42 (0.09) & 0.48 (0.23) \\
TF($\omega_{UF}$, n-p) & 0.34 (0.08) & 0.34 (0.11) & 0.59 (0.09) & 0.55 (0.12) &0.62 (0.08) & 0.44 (0.06) &\textbf{0.46 (0.10)} & 0.43 (0.05) & \textbf{0.46 (0.12)} &0.41 (0.16) & \textcolor{blue}{\textbf{0.57 (0.11)}} \\
\midrule 
\textbf{Private TrustFormer}&&& & & && &&& &\\
 TF($\omega_{PU}$, p - $\varepsilon=$1) & 0.47 (0.12) & \textbf{0.57 (0.14)} & 0.34 (0.07) & 0.36 (0.04) &0.33 (0.09) & 0.51 (0.20) &0.62 (0.12) & 0.45 (0.24) & 0.36 (0.13) &0.40 (0.09) & 0.29 (0.28) \\
 TF($\omega_{e(PUF)}$, p - $\varepsilon=$1) & 0.49 (0.11) & 0.54 (0.10) & 0.34 (0.07) & 0.38 (0.05) &0.32 (0.10) & 0.50 (0.19) &0.62 (0.12) & 0.43 (0.23) & 0.40 (0.10) &0.40 (0.09) & 0.40 (0.22) \\
 TF($\omega_{PUR}$, p - $\varepsilon=$1) & 0.47 (0.09) & 0.54 (0.17) & 0.35 (0.06) & 0.36 (0.06) &0.34 (0.13) & \textbf{0.55 (0.17)} &\textcolor{blue}{\textbf{0.64 (0.10)}} & 0.50 (0.20) & 0.38 (0.08) &0.36 (0.05) & 0.41 (0.18) \\
 TF($\omega_{e(PU)}$, p - $\varepsilon=$1) & 0.51 (0.09) & 0.52 (0.11) & 0.36 (0.11) & 0.43 (0.15) &0.32 (0.11) & 0.52 (0.18) &0.58 (0.17) & 0.48 (0.19) & 0.41 (0.11) &0.42 (0.10) & 0.40 (0.22) \\
 TF($\omega_{UR}$, p - $\varepsilon=$1) & 0.56 (0.09) & 0.42 (0.19) & 0.38 (0.11) & 0.44 (0.14) &0.35 (0.15) & 0.53 (0.19) &0.57 (0.19) & 0.50 (0.19) & 0.40 (0.11) &0.38 (0.10) & \textbf{0.43 (0.17)} \\
TF($\omega_{UFR}$, p - $\varepsilon=$1) & 0.56 (0.09) & 0.38 (0.20) & 0.37 (0.09) & 0.38 (0.04) &0.37 (0.17) & 0.51 (0.20) &0.64 (0.13) & 0.44 (0.24) & 0.38 (0.10) &0.39 (0.10) & 0.36 (0.20) \\
TF($\omega_{all}$, p - $\varepsilon=$1) & 0.58 (0.08) & 0.39 (0.17) & 0.36 (0.13) & 0.38 (0.11) &0.35 (0.17) & 0.48 (0.22) &0.57 (0.17) & 0.42 (0.24) & 0.36 (0.12) &0.39 (0.11) & 0.31 (0.28) \\
 TF($\omega_{e(UF)r(R)}$, p - $\varepsilon=$1) & 0.57 (0.12) & 0.41 (0.09) & 0.33 (0.14) & 0.38 (0.11) &0.29 (0.16) & 0.45 (0.23) &0.57 (0.17) & 0.39 (0.27) & 0.36 (0.12) &0.40 (0.11) & 0.28 (0.27) \\
TF($\omega_{UF}$, p - $\varepsilon=$1) & 0.53 (0.13) & 0.43 (0.18) & 0.35 (0.13) & 0.37 (0.12) &0.34 (0.13) & 0.48 (0.26) &0.60 (0.19) & 0.41 (0.29) & 0.33 (0.14) &0.40 (0.11) & 0.22 (0.30) \\
TF($\omega_{U}$, p - $\varepsilon=$1) & 0.58 (0.12) & 0.33 (0.24) & 0.37 (0.17) & 0.38 (0.10) &0.36 (0.23) & 0.55 (0.19) &0.60 (0.16) & \textbf{0.52 (0.21)} & 0.35 (0.15) &0.40 (0.09) & 0.26 (0.28) \\
TF($\omega_{PU}$, p - $\varepsilon=$3) & 0.52 (0.04) & 0.39 (0.23) & 0.43 (0.12) & 0.41 (0.09) &0.44 (0.18) & 0.46 (0.13) &0.57 (0.12) & 0.40 (0.14) & 0.40 (0.17) &0.44 (0.10) & 0.33 (0.30) \\
TF($\omega_{PUR}$, p - $\varepsilon=$3) & 0.58 (0.06) & 0.30 (0.24) & 0.43 (0.12) & 0.42 (0.06) &0.44 (0.22) & 0.47 (0.12) &0.55 (0.10) & 0.42 (0.13) & \textbf{0.42 (0.16)} &0.44 (0.13) & 0.37 (0.25) \\
 TF($\omega_{e(PU)}$, p - $\varepsilon=$3) & 0.61 (0.08) & 0.32 (0.24) & 0.43 (0.15) & \textbf{0.46 (0.05)} &0.40 (0.25) & 0.46 (0.13) &0.50 (0.11) & 0.44 (0.13) & 0.41 (0.17) &0.46 (0.14) & 0.32 (0.25) \\
TF($\omega_{e(PUF)},\omega_{all}$, p - $\varepsilon=$3) & 0.65 (0.05) & 0.26 (0.19) & 0.40 (0.17) & 0.42 (0.09) &0.38 (0.26) & 0.47 (0.13) &0.51 (0.14) & 0.44 (0.13) & 0.43 (0.16) &0.45 (0.14) & 0.39 (0.22) \\
 TF($\omega_{e(UF)r(R)}$, p - $\varepsilon=$3) & 0.66 (0.05) & 0.25 (0.15) & 0.40 (0.17) & 0.42 (0.09) &0.38 (0.26) & 0.47 (0.13) &0.52 (0.14) & 0.43 (0.13) & 0.40 (0.18) &0.41 (0.17) & 0.38 (0.22) \\
TF($\omega_{UF}$, p - $\varepsilon=$3) & \textcolor{blue}{\textbf{0.67 (0.03)}} & 0.15 (0.08) & 0.47 (0.11) & 0.41 (0.10) &0.50 (0.16) & 0.46 (0.11) &0.56 (0.17) & 0.40 (0.09) & 0.38 (0.16) &0.43 (0.11) & 0.30 (0.29) \\
TF($\omega_{UR}$, p - $\varepsilon=$3) & 0.64 (0.06) & 0.20 (0.18) & \textbf{0.44 (0.11)} & 0.41 (0.08) &0.46 (0.17) & 0.46 (0.09) &0.55 (0.16) & 0.41 (0.08) & 0.38 (0.19) &0.44 (0.13) & 0.29 (0.30) \\
 TF($\omega_{UFR}$, p - $\varepsilon=$3) & 0.66 (0.04) & 0.17 (0.20) & 0.43 (0.09) & 0.40 (0.08) & \textbf{ 0.46 (0.15)} & 0.47 (0.09) &0.56 (0.16) & 0.42 (0.08) & 0.37 (0.16) &0.42 (0.09) & 0.30 (0.31) \\
TF($\omega_{U}$, p - $\varepsilon=$3) & 0.64 (0.04) & 0.12 (0.07) & 0.43 (0.12) & 0.44 (0.07) &0.43 (0.18) & 0.48 (0.12) &0.53 (0.14) & 0.46 (0.11) & 0.41 (0.16) & \textbf{ 0.47 (0.07)} & 0.30 (0.31) \\

\midrule 
\midrule 
\textbf{Non-Private Baselines}&&& & & && &&& &\\
CTGAN (n-p) \cite{SDV} & 0.22 (0.04) & \textbf{0.65 (0.12)} & 0.37 (0.19) & \textbf{0.57 (0.11)} &0.28 (0.24) & 0.45 (0.19) &0.43 (0.08) & 0.46 (0.27) & \textbf{0.46 (0.11)} &\textbf{0.44 (0.11)} & 0.49 (0.21) \\
 Gaussian-Copula (n-p)\cite{SDV}& \textbf{0.36 (0.02)} & 0.62 (0.26) & \textbf{0.40 (0.03)} & 0.45 (0.02) &\textbf{0.38 (0.06)} & \textbf{0.61 (0.09)} &\textbf{0.53 (0.09)} &\textbf{ 0.67 (0.10)} & 0.44 (0.11) &0.39 (0.08) & \textbf{0.55 (0.25)} \\
 \midrule
 \textbf{Private Baselines}&&& & & && &&& &\\
DP-GAN (p - $\varepsilon=$1)\cite{xie2018differentially,qian2023synthcity}& 0.05 (0.01) & \textcolor{blue}{\textbf{0.83 (0.17)}} & 0.24 (0.09) & 0.39 (0.16) &0.17 (0.19) & 0.55 (0.15) &0.59 (0.10) & 0.52 (0.24) & 0.52 (0.08) &0.54 (0.12) & 0.48 (0.34) \\
DP-GAN (p - $\varepsilon=$3)\cite{xie2018differentially,qian2023synthcity}& 0.06 (0.01) & \textcolor{blue}{\textbf{0.83 (0.17)}} & 0.21 (0.16) & 0.44 (0.18) &0.13 (0.14) & 0.64 (0.08) &0.59 (0.16) & 0.67 (0.11) & \textcolor{blue}{\textbf{0.51 (0.25)}} & \textcolor{blue}{\textbf{ 0.58 (0.22)}} & 0.40 (0.35) \\
DP-PGM (p - $\varepsilon=$1)\cite{mckenna2021winning} & 0.34 (0.06) & 0.55 (0.24) & 0.30 (0.10) & 0.44 (0.09) &0.24 (0.11) & \textcolor{blue}{\textbf{0.70 (0.05)}} &\textbf{0.59 (0.03)} & 0.78 (0.09) & 0.46 (0.10) &0.46 (0.08) & 0.45 (0.23) \\
DP-PGM (p - $\varepsilon=$3)\cite{mckenna2021winning} & 0.34 (0.07) & 0.54 (0.23) & 0.39 (0.14) & 0.48 (0.13) &0.35 (0.15) & 0.56 (0.11) &0.49 (0.11) & 0.61 (0.16) & 0.43 (0.10) &0.40 (0.09) &\textbf{ 0.50 (0.17)} \\
DP-PGM (target, p - $\varepsilon=$1)\cite{mckenna2021winning} & 0.34 (0.06) & 0.60 (0.24) & 0.42 (0.08) & 0.41 (0.16) & \textbf{ 0.42 (0.11)} & 0.49 (0.14) &0.52 (0.18) & 0.46 (0.20) & 0.43 (0.13) &0.40 (0.10) & 0.49 (0.27) \\
DP-PGM (target, p - $\varepsilon=$3)\cite{mckenna2021winning} & \textbf{0.35 (0.07)} & 0.54 (0.24) & \textbf{0.45 (0.17)} & \textcolor{blue}{\textbf{0.60 (0.07)}} &0.37 (0.22) & 0.51 (0.13) &0.45 (0.10) & 0.56 (0.16) & 0.38 (0.18) &0.38 (0.15) & 0.36 (0.28) \\
PATE-GAN (p - $\varepsilon=$1) \cite{yoon2018pategan,qian2023synthcity} & 0.18 (0.05) & 0.68 (0.24) & 0.38 (0.04) & 0.49 (0.16) &0.32 (0.09) & 0.61 (0.11) &0.52 (0.13) & 0.68 (0.14) & 0.48 (0.17) &0.50 (0.11) & 0.43 (0.31) \\
PATE-GAN (p - $\varepsilon=$3)\cite{yoon2018pategan,qian2023synthcity} & 0.21 (0.03) & 0.67 (0.27) & 0.33 (0.07) & 0.45 (0.04) &0.26 (0.08) & 0.65 (0.07) &0.57 (0.06) & 0.71 (0.11) & 0.35 (0.06) &0.33 (0.12) & 0.37 (0.31) \\
\midrule 
\midrule
\textbf{Real Data} & N/A & N/A & 0.56 (0.08) & 0.53 (0.08) &0.58 (0.12) & 0.47 (0.17) &0.44 (0.08) & 0.49 (0.24) & 0.38 (0.10) &0.38 (0.06) & 0.36 (0.20) \\
\bottomrule
\end{tabular}
}
\captionof{table}{Law School downstream task evaluation. Study on the effect of debiasing in utility training on trust indices of utility, fairness and robustness.}

\label{table:law_school_metrics_baised_debiased}

\end{table*}


\begin{table*}[htp!]
\resizebox{1\textwidth}{!} {
\begin{tabular}{lcccccccccc}
\toprule
Model &$\omega_{all}$ &$\omega_{e(PU)}$ &$\omega_{e(PUF)}$ &$\omega_{e(UF)r(R)}$ &$\omega_{PU}$ &$\omega_{PUR}$ &$\omega_{U}$ &$\omega_{UF}$ &$\omega_{UFR}$ &$\omega_{UR}$ \\
\midrule
\textbf{Non-private TrustFormer}&&& & & && &&& \\
 TF ($\omega_{U}$, n-p) &25 & 19 &23 &24 & 14 &13 & \rankone{1 }&8 & 4 &\ranktwo{2}\\
 TF ($\omega_{all}$, n-p) &17 & 23 &25 & 7 & 23 &22 & 9 &4 &10 &9 \\
 TF ($\omega_{e(UF)r(R)}$, n-p) &16 & 20 &20 & 4 & 21 &21 &10 &\ranktwo{2}& 9 & 10 \\
 TF ($\omega_{PUR}$, n-p) &23 &6 &13 &13 &\ranktwo{2}& \rankthree{3} & \ranktwo{2} &6 & 8 &7 \\
TF ($\omega_{UR}$, n-p) &21 & 12 &16 &10 & 10 & 9 & 5 &\rankthree{3} & \rankthree{3} &4 \\
 TF ($\omega_{e(PUF)}$, n-p) &14 & 11 &17 &11 & 11 &11 & 8 & 11 &11 &6 \\
TF ($\omega_{e(PU)}$, n-p) &15 &8 &14 &14 &8 & 7 & 6 & 10 &12 &5 \\
 TF ($\omega_{UFR}$, n-p) &28 & 15 &18 &19 &9 &10 & 4 &5 & 7 &8 \\
TF ($\omega_{PU}$, n-p) &20 &4 &12 &18 &\rankone{1} & \ranktwo{2} & \rankthree{3} &9 & 5 &\rankthree{3} \\
TF ($\omega_{UF}$, n-p) &22 & 18 &19 &17 & 15 &12 & 7 &\rankone{1} & \rankone{1} &\rankone{1} \\

\midrule 
\textbf{Private TrustFormer}&&& & & && &&& \\
 TF ($\omega_{PU}$, p - $\epsilon=1$) & 9 & 13 & 9 &15 & 17 &23 &33 & 33 &37 & 34 \\
TF ($\omega_{e(PUF)}$, p - $\epsilon=1$) & 6 & 14 &10 &21 & 20 &20 &34 & 34 &34 & 30 \\
TF ($\omega_{PUR}$, p - $\epsilon=1$) & 5 & 10 & 8 & 6 & 19 &24 &32 & 27 &30 & 32 \\
 TF ($\omega_{e(PU)}$, p - $\epsilon=1$) & \ranktwo{2} &5 & 7 & 5 & 18 &18 &30 & 29 &25 & 27 \\
 TF ($\omega_{UR}$, p - $\epsilon=1$) & \rankthree{3} & 17 &11 & \rankthree{3} & 26 &26 &24 & 22 &20 & 26 \\
TF ($\omega_{UFR}$, p - $\epsilon=1$) &13 & 24 &24 &20 & 28 &28 &26 & 28 &31 & 28 \\
TF ($\omega_{all}$, p - $\epsilon=1$) &18 & 27 &27 &26 & 29 &30 &29 & 32 &36 & 31 \\
TF ($\omega_{e(UF)r(R)}$, p - $\epsilon=1$) &30 & 31 &31 &33 & 31 &32 &35 & 37 &39 & 36 \\
 TF ($\omega_{UF}$, p - $\epsilon=1$) &29 & 28 &28 &27 & 27 &31 &31 & 35 &38 & 37 \\
TF ($\omega_{U}$, p - $\epsilon=1$) &24 & 30 &29 &16 & 33 &34 &28 & 23 &32 & 33 \\
 TF ($\omega_{PU}$, p - $\epsilon=3$) &12 & 22 &22 &22 & 25 &25 &18 & 26 &24 & 22 \\
TF ($\omega_{PUR}$, p - $\epsilon=3$) &19 & 29 &30 &25 & 32 &29 &14 & 24 &18 & 12 \\
 TF ($\omega_{e(PU)}$, p - $\epsilon=3$) &11 & 26 &26 &23 & 30 &27 &17 & 25 &21 & 17 \\
 TF ($\omega_{e(PUF)}, \omega_{all}$, p - $\epsilon=3$) &27 & 32 &32 &30 & 34 &33 &21 & 30 &22 & 18 \\
TF ($\omega_{e(UF)r(R)}$, p - $\epsilon=3$) &31 & 33 &33 &32 & 35 &35 &22 & 31 &28 & 25 \\
 TF ($\omega_{UF}$, p - $\epsilon=3$) &36 & 37 &38 &35 & 38 &38 &11 & 15 &19 & 14 \\
 TF ($\omega_{UR}$, p - $\epsilon=3$) &33 & 34 &34 &31 & 36 &36 &13 & 20 &23 & 21 \\
TF ($\omega_{UFR}$, p - $\epsilon=3$) &35 & 35 &36 &34 & 37 &37 &15 & 19 &26 & 24 \\
TF ($\omega_{U}$, p - $\epsilon=3$) &37 & 39 &39 &37 & 39 &39 &16 & 18 &17 & 15 \\

\midrule 
\midrule 
\textbf{Non-private Baselines}&&& & & && &&& \\
CTGAN (n-p) \cite{SDV} &32 & 21 &21 &36 &7 & 4 &27 & 36 &27 & 23 \\
Gaussian-Copula (n-p)\cite{SDV} & \rankone{1} &\rankone{1} & \rankone{1} & \rankone{1} &4 & 5 &20 &7 & 6 & 16 \\

\midrule 
\textbf{Private Baselines}&&& & & && &&& \\
DP-GAN (p - $\epsilon=1$)\cite{qian2023synthcity,chen2020gan} &39 & 38 &37 &39 & 16 & 8 &38 & 39 &35 & 35 \\
DP-GAN (p - $\epsilon=3$)\cite{qian2023synthcity,chen2020gan} &38 & 36 &35 &38 & 22 &14 &39 & 38 &33 & 39 \\
DP-PGM (p - $\epsilon=1$)\cite{mckenna2021winning} & 4 & 16 & \ranktwo{2} & 9 & 24 &17 &37 & 16 &13 & 29 \\
DP-PGM (p - $\epsilon=3$)\cite{mckenna2021winning} & 8 &7 & 6 & 8 & 13 &16 &23 & 14 &14 & 20 \\
 DP-PGM (target, p - $\epsilon=1$)\cite{mckenna2021winning} & 7 &\ranktwo{2}& \rankthree{3} &12 &5 & 6 &19 & 21 &16 & 13 \\
 DP-PGM (target, p - $\epsilon=3$)\cite{mckenna2021winning} &10 &\rankthree{3} & 5 & \ranktwo{2} &6 &15 &12 & 12 &15 & 19 \\
PATE-GAN (p - $\epsilon=1$)\cite{qian2023synthcity,yoon2018pategan} &26 &9 & 4 &28 &\rankthree{3} & \rankone{1} &25 & 13 & \ranktwo{2} & 11 \\
 PATE-GAN (p - $\epsilon=3$)\cite{qian2023synthcity,yoon2018pategan} &34 & 25 &15 &29 & 12 &19 &36 & 17 &29 & 38 \\
\bottomrule
\end{tabular}
}
\captionof{table}{Law school dataset synthetic data ranking .We see that overall TrustFormers that use trustworthiness index driven cross-validation corresponding to the desired trade-offs outperform other synthetic data,  on 6 out of 10 weighting in Table \ref{Table:weights}. Note that the ranking here is on the mean trustworthiness index , where downstream tasks are evaluated on the test data in each real data fold. }
\label{table:law_school_ranking}
\end{table*}

\begin{table*}[htp!]
\resizebox{1\textwidth}{!} {
 \begin{tabular}{lcccccccccc}
\toprule
Model &$\omega_{all}$ &$\omega_{e(PU)}$ &$\omega_{e(PUF)}$ &$\omega_{e(UF)r(R)}$ &$\omega_{PU}$ &$\omega_{PUR}$ &$\omega_{U}$ &$\omega_{UF}$ &$\omega_{UFR}$ &$\omega_{UR}$ \\
\midrule
\textbf{Non-private TrustFormer}&&& & & && &&& \\
  TF ($\omega_{U}$, n-p) & 23 & 11 & 14 &\rankone{1}& 12 & 7 &\rankone{1}&\rankthree{3}& 9 &\rankthree{3}\\
 TF ($\omega_{all}$, n-p) & 9 & 9 & 7 & 4 & 22 & 18 & 9 &\ranktwo{2}& 16 & 9 \\
 TF ($\omega_{e(UF)r(R)}$, n-p) & 10 & 10 & 11 &\rankthree{3}& 20 & 17 & 10 &\rankone{1}& 14 & 10 \\
 TF ($\omega_{PUR}$, n-p) & 4 &\rankone{1}&\ranktwo{2}& 11 &\ranktwo{2}&\rankthree{3}& 4 & 8 & 4 & 5 \\
 TF ($\omega_{UR}$, n-p) &\ranktwo{2}& 6 & 10 & 9 & 6 & 5 &\rankthree{3}& 5 & 5 & 4 \\
 TF ($\omega_{e(PUF)}$, n-p) & 6 & 4 & 5 & 10 & 10 & 8 & 7 & 11 & 10 & 8 \\
 TF ($\omega_{e(PU)}$, n-p) &\rankthree{3}&\rankthree{3}& 4 & 12 &\rankthree{3}&\ranktwo{2}&\ranktwo{2}& 10 & 11 & 7 \\
 TF ($\omega_{UFR}$, n-p) & 8 & 7 & 15 & 13 & 5 & 4 & 5 & 7 & 6 & 6 \\
 TF ($\omega_{PU}$, n-p) & 7 &\ranktwo{2}&\rankone{1}& 7 &\rankone{1}&\rankone{1}& 6 & 6 &\rankthree{3}&\ranktwo{2}\\
 TF ($\omega_{UF}$, n-p) &\rankone{1}& 12 & 12 & 8 & 16 & 11 & 8 & 4 &\rankone{1}&\rankone{1}\\

 \midrule 
\textbf{Private TrustFormer}&&& & & && &&& \\
  TF ($\omega_{PU}$, p - $\epsilon=$1) & 18 & 14 & 18 & 24 & 14 & 21 & 29 & 32 & 21 & 34 \\
 TF ($\omega_{e(PUF)}$, p - $\epsilon=$1) & 16 & 15 & 19 & 25 & 19 & 19 & 31 & 33 & 13 & 28 \\
 TF ($\omega_{PUR}$, p - $\epsilon=$1) & 14 & 16 & 16 & 16 & 18 & 13 & 27 & 26 & 20 & 31 \\
 TF ($\omega_{e(PU)}$, p - $\epsilon=$1) & 5 & 5 & 9 & 14 & 17 & 20 & 28 & 30 & 12 & 26 \\
 TF ($\omega_{UR}$, p - $\epsilon=$1) & 13 & 20 & 20 & 15 & 24 & 26 & 23 & 21 & 8 & 25 \\
 TF ($\omega_{UFR}$, p - $\epsilon=$1) & 28 & 26 & 28 & 28 & 30 & 24 & 25 & 25 & 17 & 30 \\
 TF ($\omega_{all}$, p - $\epsilon=$1) & 30 & 25 & 29 & 30 & 28 & 30 & 30 & 34 & 34 & 32 \\
 TF ($\omega_{e(UF)r(R)}$, p - $\epsilon=$1) & 29 & 21 & 27 & 35 & 26 & 34 & 36 & 36 & 37 & 37 \\
 TF ($\omega_{UF}$, p - $\epsilon=$1) & 27 & 22 & 24 & 31 & 25 & 32 & 33 & 35 & 36 & 38 \\
 TF ($\omega_{U}$, p - $\epsilon=$1) & 24 & 27 & 25 & 19 & 33 & 33 & 32 & 23 & 32 & 36 \\
 TF ($\omega_{PU}$, p - $\epsilon=$3) & 25 & 23 & 23 & 21 & 23 & 23 & 19 & 13 & 26 & 19 \\
 TF ($\omega_{PUR}$, p - $\epsilon=$3) & 20 & 29 & 30 & 22 & 32 & 25 & 17 & 19 & 23 & 15 \\
 TF ($\omega_{e(PU)}$, p - $\epsilon=$3) & 17 & 24 & 26 & 18 & 31 & 22 & 21 & 22 & 27 & 18 \\
TF ($\omega_{e(PUF)}, \omega_{all}$, p - $\epsilon=$3) & 21 & 32 & 32 & 26 & 34 & 31 & 24 & 28 & 28 & 21 \\
 TF ($\omega_{e(UF)r(R)}$, p - $\epsilon=$3) & 26 & 31 & 31 & 23 & 35 & 35 & 26 & 29 & 33 & 29 \\
 TF ($\omega_{UF}$, p - $\epsilon=$3) & 34 & 36 & 36 & 32 & 37 & 38 & 11 & 12 & 24 & 14 \\
 TF ($\omega_{UR}$, p - $\epsilon=$3) & 32 & 34 & 34 & 29 & 36 & 36 & 13 & 17 & 31 & 20 \\
 TF ($\omega_{UFR}$, p - $\epsilon=$3) & 35 & 37 & 37 & 36 & 38 & 37 & 14 & 18 & 30 & 23 \\
 TF ($\omega_{U}$, p - $\epsilon=$3) & 37 & 39 & 39 & 34 & 39 & 39 & 18 & 20 & 22 & 17 \\

 \midrule 
\midrule 
\textbf{Non-private Baselines}&&& & & && &&& \\
 SDV-CTGAN (n-p) & 36 & 33 & 33 & 37 & 15 & 16 & 34 & 37 & 38 & 22 \\
 SDV-Gaussian-Copula (n-p) & 11 & 13 & 6 & 6 & 8 & 10 & 12 & 9 &\ranktwo{2}& 12 \\

\midrule 
\textbf{Private Baselines}&&& & & && &&& \\
 PATE-GAN (p - $\epsilon=$1) & 19 & 18 & 13 & 20 & 4 & 6 & 20 & 16 & 7 & 11 \\
 DP-PGM (p - $\epsilon=$1) & 31 & 28 & 21 & 27 & 29 & 29 & 37 & 31 & 19 & 27 \\
 DP-GAN (p - $\epsilon=$3) & 39 & 38 & 38 & 39 & 27 & 28 & 39 & 39 & 39 & 39 \\
 DP-GAN (p - $\epsilon=$1) & 38 & 35 & 35 & 38 & 11 & 9 & 38 & 38 & 35 & 33 \\
 DP-PGM (p - $\epsilon=$3) & 15 & 17 & 8 & 5 & 13 & 12 & 22 & 15 & 15 & 16 \\
 DP-PGM (target, p - $\epsilon=$1) & 22 & 19 & 17 & 17 & 9 & 14 & 16 & 27 & 18 & 13 \\
 DP-PGM (target, p - $\epsilon=$3) & 12 & 8 &\rankthree{3}&\ranktwo{2}& 7 & 15 & 15 & 14 & 25 & 24 \\
 PATE-GAN (p - $\epsilon=$3) & 33 & 30 & 22 & 33 & 21 & 27 & 35 & 24 & 29 & 35 \\

\bottomrule
\end{tabular}
}
\captionof{table}{Law School dataset. Ranking  under uncertainty of all considered synthetic data using $R^{\alpha}_{\tau}$ defined in Equation \eqref{eq:mean_var} for $\alpha=0.1$. We see that TrustFormer synthetic data is still aligned with the prescribed safeguards while having low volatility in its trustworthiness index. TrustFormer synthetic data is back to the top of the leader-board when we take into account uncertainty.}
\label{table:lawschool_ranking_uncertainty}
\end{table*}

\end{document}